\documentclass[runningheads]{llncs}
\usepackage{graphicx}
\usepackage[ruled,linesnumbered]{algorithm2e}
\usepackage{svg}
\usepackage{amsmath}
\usepackage{epstopdf}
\usepackage{tikz}  
\usepackage{multirow}
\usepackage{booktabs}
\usetikzlibrary{arrows.meta,positioning,calc}
\newcommand{\B}{\fontseries{b}\selectfont}
\newcommand{\xhdr}[1]{\vspace{1.7mm}\noindent{{\bf #1}}}
\begin{document}
\title{Fea2Fea: Exploring Structural Feature Correlations via Graph Neural Networks}
\vspace{-2cm}
\author{Jiaqing Xie\inst{1} \and Rex Ying\inst{2}}
\vspace{-2cm}
\institute{University of Edinburgh
\email{s2001696@.ed.ac.uk}
\and
Stanford University
\email{rexying@stanford.edu}}
\maketitle              % typeset the header of the contribution
\vspace{-1cm}
\begin{abstract}
Structural features are important features in a geometrical graph. Although there are some correlation analysis of features based on covariance, there is no relevant research on structural feature correlation analysis with graph neural networks. In this paper, we introuduce graph feature to feature 
(\textbf{Fea2Fea}) prediction pipelines in a low dimensional space to 
explore some preliminary results on structural feature correlation, which is based on graph neural network. The results show that there exists high 
correlation between some of the structural features. An irredundant feature combination with initial node features, which is filtered by graph neural network has improved its classification accuracy in some graph-based tasks. We compare differences between concatenation methods on connecting embeddings between features and show that \textit{the simplest is the best}. We generalize on the synthetic geometric graphs and certify the results on prediction difficulty between structural features. 
\vspace{-0.2cm}
\keywords{Graph neural networks \and Feature engineering}
\end{abstract}
\vspace{-1cm}
\section{Introduction}
\vspace{-0.4cm}
Designing graph neural networks (GNN) with various message passing architectures has been a trend recently. Many powerful graph convolution methods such as GraphSAGE \cite{hamilton2017inductive}, GCN \cite{kipf2016semi}, GIN \cite{xu2018powerful} and GAT \cite{velivckovic2017graph} have been proposed, with a wide range of applications including social networks \cite{Qiu_2018,Wang_2019,zhang2018link}, molecules  \cite{jin2019learning,Kearnes_2016,knyazev2018spectral,Xu_2019}, natural language processing \cite{palm2018recurrent,rahimi-etal-2018-semi,sorokin-gurevych-2018-modeling} and physics simulations \cite{sanchezgonzalez2020learning,seo*2020physicsaware}. Meanwhile, some issues about how to add extra node features reasonably are rising with the development of GNN, which are mainly discussed in this paper.

Some works emphasized the importance of adding structural feature information: pagerank \cite{klicpera2019predict,ilprints422}, node degree \cite{ying2019hierarchical}, clustering coefficient \cite{hamilton2017inductive} or adding different graph features \cite{you2020design}, including one hot vector, constant scalar, clustering coefficient and pagerank to perform node or graph classifications. Shortest path length is also covered when considering the importance of distant nodes \cite{dwivedi2020benchmarking,yin2020revisiting}.  However, previous works ignored the importance of structural features' correlations of each node. In addition, they do not take the advantage of graph feature's correlation information to add features to the original node feature but arbitrarily add these node features, which might include repetitive or highly related information, leading to data redundancy.

Supervised tasks for selecting features via correlation based algorithm have been proposed \cite{doi:10.1260/1748-3018.6.3.385,10.1007/978-981-13-7403-6_4}. One work shows whether there exists a strong correlation between node features and node labels \cite{duong2019node} and another work uses node correlation information to perform convolution pooling on graph classification tasks \cite{10.1145/3414274.3414490}. However, most correlation metrics are not based on deep or graph neural network models but based on simple analytical solutions or analysis based on covariance matrices, which might ignore the role of graph neural network on generating correlation between features with enriched message passing information from neighbours. A graph neural network based model to achieve node structural feature correlations has been not investigated.

\xhdr{Present work} 
In this paper, we propose a framework for processing graph \textit{
feature} to \textit{feature} prediction(feature mainly refers to a graph's structural feature in this paper), called \textbf{Fea2Fea}, which includes two important components: single feature to single feature prediction(Fea2Fea-single) and multiple features to single single feature prediction (Fea2Fea-multiple): 1) Fea2Fea-single returns the correlation matrix implemented by graph neural networks 2) Fea2Fea-multiple takes the use of the correlation matrix achieved in Fea2Fea-single to summarize some possible combinations of features with graph neural network based filter. Finally it is transferred to real world applications, such as combination of initial features with embedded filtered structural features to perform node or graph classification on some benchmark graph datasets, which is proved to be worked on some graph classfication task based datasets such as {\sc{Proteins}} and {\sc{Nci1}}. Details of the framework and model architectures are described in part 3.

Based on the experiments implemented according to the framework, our works have reached several important \textbf{findings}: 1) Graph neural networks are superior to deep neural networks without graph embedding layers in mutual prediction of graph features  2) The expressive power of the graph neural netowrk in achieving the correlation of graph features, such as the ease with which other structural features  to predict \textit{node degree} and general difficulty to predict \textit{pagerank} 3) A feature combination  of predicting another single feature via graph neural networks with different concatenation methods provides enriched embedding options but overall the simplest concatenation is the best 4) For some benchmark data sets, the initial node identity feature combined with additional structural embedded information leads to higher classification accuracy.
The main \textbf{advantages} of our works are that they 
1) illustrate feature correlations with low dimensionality, where the number of input dimensions is controlled to a maximum of 5 in Fea2Fea-multiple pipeline
2) filter additional redundant node features via graph neural network based models instead of covariance explained models
3) require a less rigorous constructed models to perform comparisons which means that we introduce the importance of adding non-correlated features other than defeating the state of the arts graph embedding methods.

\vspace{-0.5cm}
\section{Related Works}
\vspace{-0.52cm}
\xhdr{GNN expressiveness}
Some works that have emphasized the importance of adding structural features to input features are mentioned in the \textit{Introduction} part, as well as the work \cite{lerique2019joint} that uses graph structure combined with node features to perform predictions. There are also some works mentioning the importance of adding specific features to make graph neural network more expressive. GIN \cite{xu2018powerful} has shown that identity features are not powerful as it does not pass WL tests possibly which is to check symmetric graphs in the datasets. A recent work called identity-awared GNN \cite{you2021identityaware} emphasizes the self versus neighbour features, which proves that adding neighbourhood information as feature augmentation will distinguish symmetric graphs and pass WL tests more easily which is more expressive than GIN. Another work \cite{sato2020survey} also discusses the expressiveness of GNN. 

\xhdr{Feature correlation}
Many useful analytical methods for measuring feature correlation have been proposed. Besides the covariance based methods that are mentioned in part 1, there are some other useful methods, such as a cross-entopy model with the conception of symmetrical uncertainty metrics \cite{Yu2003FeatureSF} and computing a tree-decomposition based correlation graph to select unredundant features \cite{Ouali2019AFS}. However, the lack of research on graph neural network based model on achieving feature correlation/selection is the potential motive of this paper. We'd like to present some feature correlation results with GNN based filters.

\xhdr{Other graph embedding methods} Recent years, researches have been conducted on learning graph embeddings. Many useful graph convolution methods have been proposed besides GIN, GAT, GCN and GraphSAGE which are the four main graph embedding methods that are implemented in our paper. For example, the graph convolution with ARMA filters \cite{Bianchi_2021}, graph neural network with attention \cite{thekumparampil2018attentionbased},  with gated recurrent units \cite{li2017gated} and with chebyshev spectral graph convolutional operator \cite{defferrard2017convolutional}.

\vspace{-0.5cm}

\section{Methods}
\vspace{-0.3cm}
\subsection{preliminary graph feature extraction}
\vspace{-0.3cm}
\xhdr{Graph annotation} Given a graph $\mathcal G$($\mathcal V, \mathcal E$) where $\mathcal V$ is the vertex set of the graph and $\mathcal E$ is the edge set of the graph . We want to achieve a full-scale structural feature 
matrix $\textbf{x}_{\mathcal G} \in  \bbbr^{|\mathcal{V}| \times \mathcal{D}}$ from $\mathcal{G}$ that requires an adjacency matrix $\mathcal A_{g} \in \bbbr^{|V|*|V|}$ of $\mathcal G$,
 where $ V$ is the number of vertexes in the graph and $\mathcal D$ is
 the total input dimensions of structural features.
In this paper we mainly choose five of all structural features to show exploratory results, which are \textit{constant 
feature}(\textbf{Cons}), \textit{node degree}(\textbf{Deg}), \textit{clustering coefficient}(\textbf{Clu}), \textit{average path length}(\textbf{AvgLen}) and \textit{Pagerank}(\textbf{PR})
. In this case $\mathcal D$ is equal to 5.

\xhdr{Feature extraction} Constant feature of node u $\mathop{Cons}\limits_{u \in \mathcal{V}}(u)$ is given by c, 
where c $\in \bbbr^{+}$. Each node's constant feature is set to 1 in this paper for standardization. 
Degree of node u $\mathop{Deg}\limits_{u \in \mathcal{V}}(u)$ is equal to the number of node u's 
neighbours. 
Clustering coefficient of node u $\mathop{Clu}\limits_{u \in \mathcal{V}}(u)$ is given by $\frac{2e_{jk}}
{k_{i} * (k_{i} - 
1)}$, where $j,k \in \mathcal{V}$ and $e_{jk}$ represents the total possible edges between node u's 
neighbours and
$k_{i}$ is the number of node u's neighbours. Pagerank of a node u $\mathop{PR}\limits_{u \in \mathcal{V}}
(u)$ is given by:
 $\frac{1-q}{|\mathcal V|} + q \times \sum_{v \in {\mathcal 
N} 
  (u)}\frac{\mathop{PR}(v)}{\mathcal L(v)}$,
where $\mathcal N(\cdot)$ represents the node u's neighbours and $\mathcal L(\cdot)$ is the number of outbound links 
from node u
to its neighbours. In the traditional analysis of pagerank algorithm, the object graph is directed. 
However we can treat
undirected graph as bidirectional graph, where outbound link is equal to the number of edges from node u to 
its neighbours.
q is the residual probability and equal to 0.85 as default which can be regarded as a hyper-parameter. Average path 
length of a node  $\mathop{Avglen}
\limits_{u \in \mathcal{V}}(u)$ is given by:
$ \frac{1}{\mathcal V '}\sum_{v \in {\mathcal V}\neq u} I(u,v) * d_{min}(u,
v)$,
where $I(u,v)$ indicates whether there's a path from node u to node v. If node v is reachable for node u,
then $I(u,v)$ is equal to 1 otherwise it is 0. $\mathcal V'$ is equal to the total number of reachable nodes 
for node u,
more specifically, $\mathcal V'$ = $\sum_{v \in {\mathcal V}\neq u} I(u,v) $. $d_{min}(u,v)$ determines 
the shortest path
length from node u to node v.
Therefore, an entry $\textbf{I}(v), v \in \mathcal V$ can be written as:  $I
(v)$  = $Cons(v) \oplus  Deg(v) \oplus Clu(v) \oplus PR(v) \oplus Avglen(v)$ where $\oplus$ is the 
direct concatenation.
\vspace{-0.5cm}
\begin{figure}
  \centering
  \includegraphics[width=0.8\textwidth]{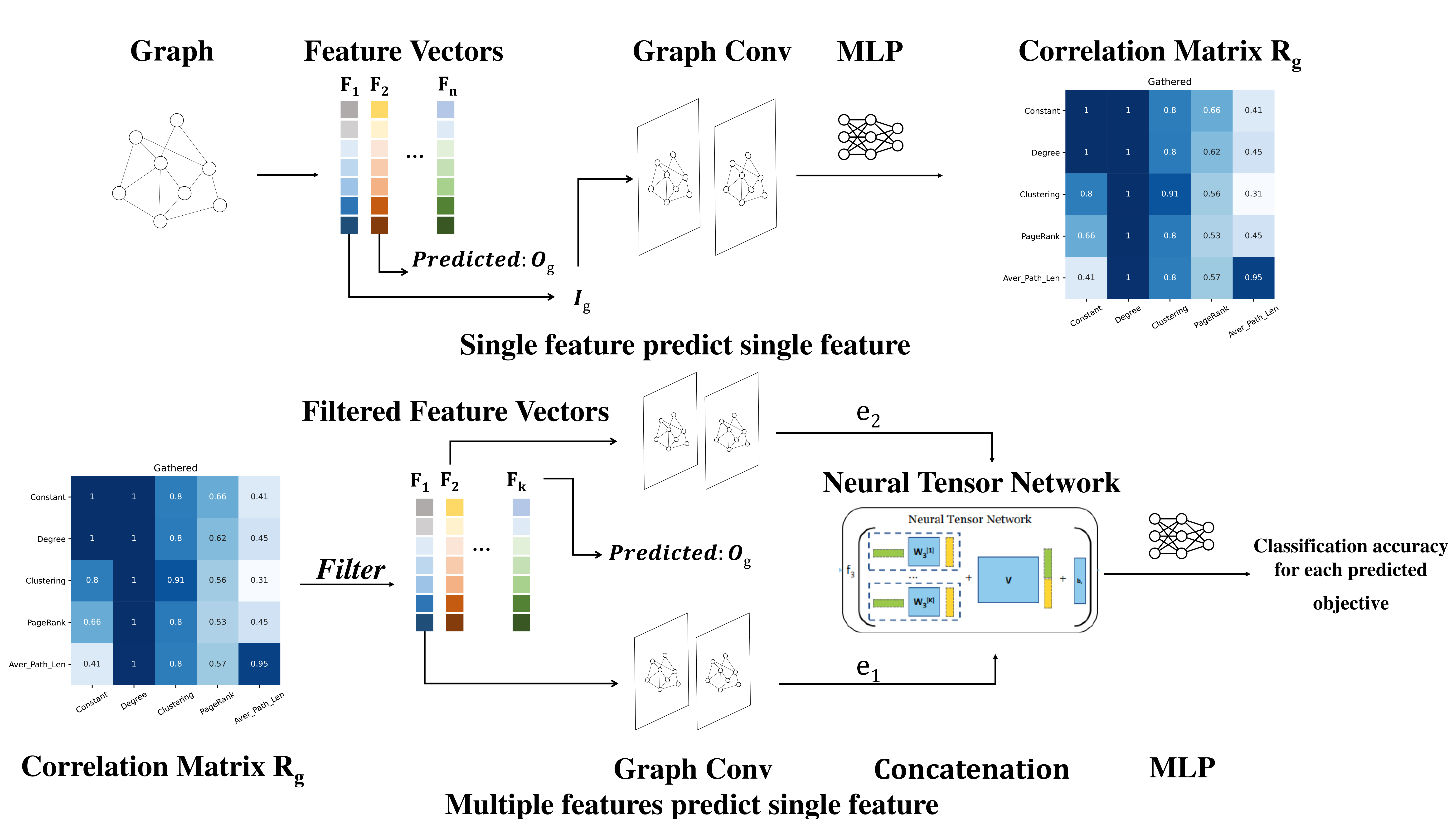}
  \caption{Pipeline model for feature mutual prediction task, where 
the neural tensor network is originated in NLP domain \cite
{NIPS2013_b337e84d} and sketch of the NTN block comes from \textit
{SimGNN} \cite{bai2020simgnn}. In the Fea2Fea-simple model, only single feature is regarded as the input while mulitple features which are un-correlated are regarded as the input when it comes to Fea2Fea-multiple model, which is presented at the bottom part of the image. Neural Tensor Network, an example of feature concatenation methods is not necessary but just taken as an example to be familiar with our pipeline. }
\end{figure}

\vspace{-1.2cm}
\subsection{Fea2Fea-single: Single feature predicts single feature}
\xhdr{Obtain a correlation matrix by GNN} We have obtained a feature matrix $\textbf{x}_{\mathcal G} \in  \bbbr^{|\mathcal{V}|\times 5}$. Specifically, graph features
have been indexed to simplify analysis. The feature order is given by constant feature, degree, clustering coefficient, pagerank and 
average path length, which are indexed from 1 to 5 respectively. Suppose we have built 
a model $\mathcal M$ which is based on one of the four graph embedding methods together with multi-layer perceptrons. We extract two features each time, one
for input $\textbf{I}_{\mathcal G}$ and the other for output $\textbf{O}_{\mathcal G}$. We set bins for the output
since we focus more on classification task rather than regression task. After outputs are classified into their own bins, we use model $\mathcal M$ to predict $\textbf{O}_{\mathcal G}$. The trained model 
will generate predictive output $\textbf{O}^{'}_{\mathcal G}$.  We use mean classification \textit{accuracy} on test datasets and \textit{negative loss likelihood loss}(NLLloss) as two main evaluators in this paper. 
A higher accuracy indicates that it is easier to predict from input feature to output feature. 
We will implement a feature correlation matrix $\mathcal R \in \bbbr^{5 * 5}$ since each feature can be taken both as input and output, except for the case of constant feature. We should note that constant feature can 
be only taken as input but not as output. It is the most basic feature for all graphs and will not lead to a 
classification problem. We can set the element in the matrix to infinity or let it symmetric to the  diagonal under this circumstance. Algorithm 1 in appendix A shows the complete process of how to construct a correlation matrix for feature mutual prediction. 
\vspace{-0.45cm}
\subsection{Fea2Fea-Multiple: Mutiple features predict single feature}

\xhdr{Threshold mechanism on choosing features} We obtain feature correlation matrix $\mathcal R$ from Fea2Fea-single. The value $\mathcal R(i,j)$ in the matrix $\mathcal R$ indicates whether feature $  f_{i}$ and feature $  f_{j}$ can be easy or difficult to predict each other. Moreover, we want to add more features to $  f_{i}$ to see if it will predict $ f_{j} $ more accurately. However we cannot add features arbitrarily since adding similar features may cause data redundancy. Therefore we need to collect all possible feature combinations without redundant information by applying a threshold mechanism.
Given a threhold $t$, we filter out and discard such feature combination : ($  f_{i}$, $  f_{j}$) where $\mathcal R(i,j)$ is greater than $t$. Consider two extreme cases.   All feature combinations are considered for training when $t$ is equal to 1 while no feature combinations are considered when $t$ is equal to 0. When t $\in$(0,1), the number of possible concatenations $\mathcal N$ $\in$[1,$2^d-1$).

 We initialize an array called \textit{Comb} which includes all concatenation between features. The time complexity of this generation is $\Omega (2^d)$, where d is equal to the input feature dimension. It works efficiently when input feature dimension is small. A filter is applied after generating the array. For each $f_{i}$ and $f_{j}$ in each element of $Comb$, if both $ \mathcal R(i, j)$ and $ \mathcal R(j, i)$ are less than threshold $t$, the combination of $f_{i}$ and $f_{j}$ is valid, otherwise this combination is moved from $Comb$. Finally we achieve a filtered $Comb$.

\xhdr{Feature concatenation}  
Given a valid combination $f_{1}$, $f_{2}$,...,$f_{k}$, where k $\in$\{2,3,4\}. We implement graph convolution layers for each feature to map them into graph embedding space: $e_{1}$, $e_{2}$,...,$e_{k}$. We provide three methods to concatenate those features after graph embeddings, which are simple concatenation, biliear concatenation and neural tensor network(NTN) \cite{NIPS2013_b337e84d} based concatenation. Assume that the graph embedding space of each feature is d. NTN concatenation contains both direct(simple) concatenation and bilinear concatenation, which is given by:
\begin{equation}
  g_{(t)} = \textbf{u}^{T} \textbf{h}(\underbrace{g_{(t-1)}^{T} \textbf{W}_{(t)} e_{t}}_{\text{bilinear    concatenation}} +
  \underbrace{g_{(t-1)} \oplus e_{t}}_{\text{simple concatenation}} + \textbf{b}), g_{(0)} = e_{1}\in \bbbr^{d}\\ 
\end{equation}
where $g_{(t)} \in \bbbr^{td}$ represents the temporal embedding after t-th concatenation, $\oplus$ represents the function that connects two features together with the summation of their dimensions, $g_{(0)}$ is initialized by $e_{1}$, the weight matrix $\textbf{W}_{(t)} \in \bbbr^{(t-1)d * td * d} $ with t$\geq$2, the obtained graph embedding $g_{(t)} \in \bbbr^{td}$,
weight vector $\textbf{u}^{T} \in \bbbr^{td}$ and $\textbf{h}$ represents the activation function: \textit{tanh} function as it is stated in original paper. However, we cannot conclude that NTN concatenation is superior to the other two concatenation methods according to equation(1). We should implement experiments with all three concatenation methods separately. Fea2Fea-Multiple will finally achieve the average accuracy of each combination, as well as the average accuracy of different numbers of input dimensions, which is the base of real world applications.
\begin{figure}[!tp]
  \centering
  \begin{center}
    \hspace*{-1in}
      \includegraphics[width=0.5\linewidth]{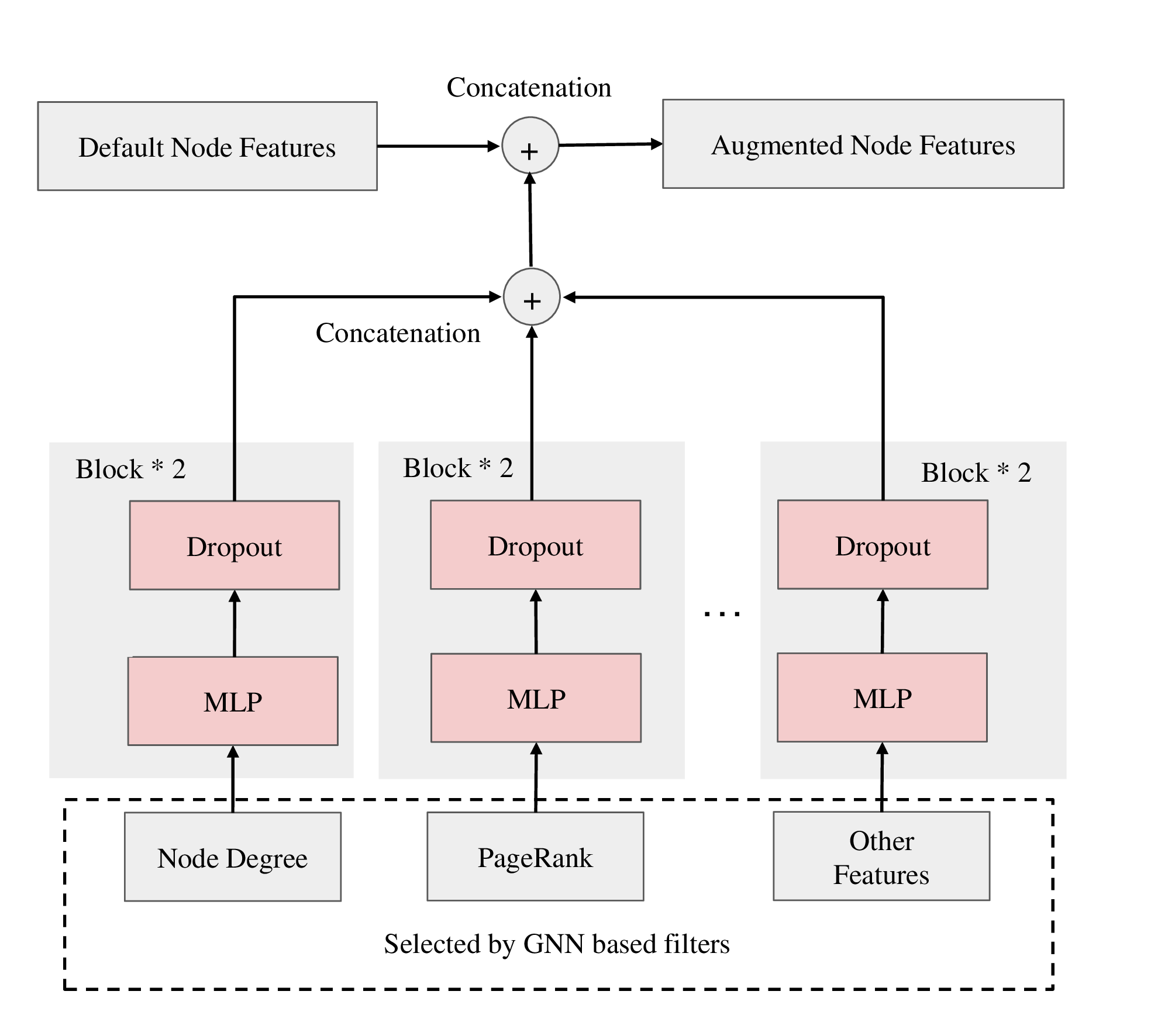}
      \includegraphics[width=0.5\linewidth]{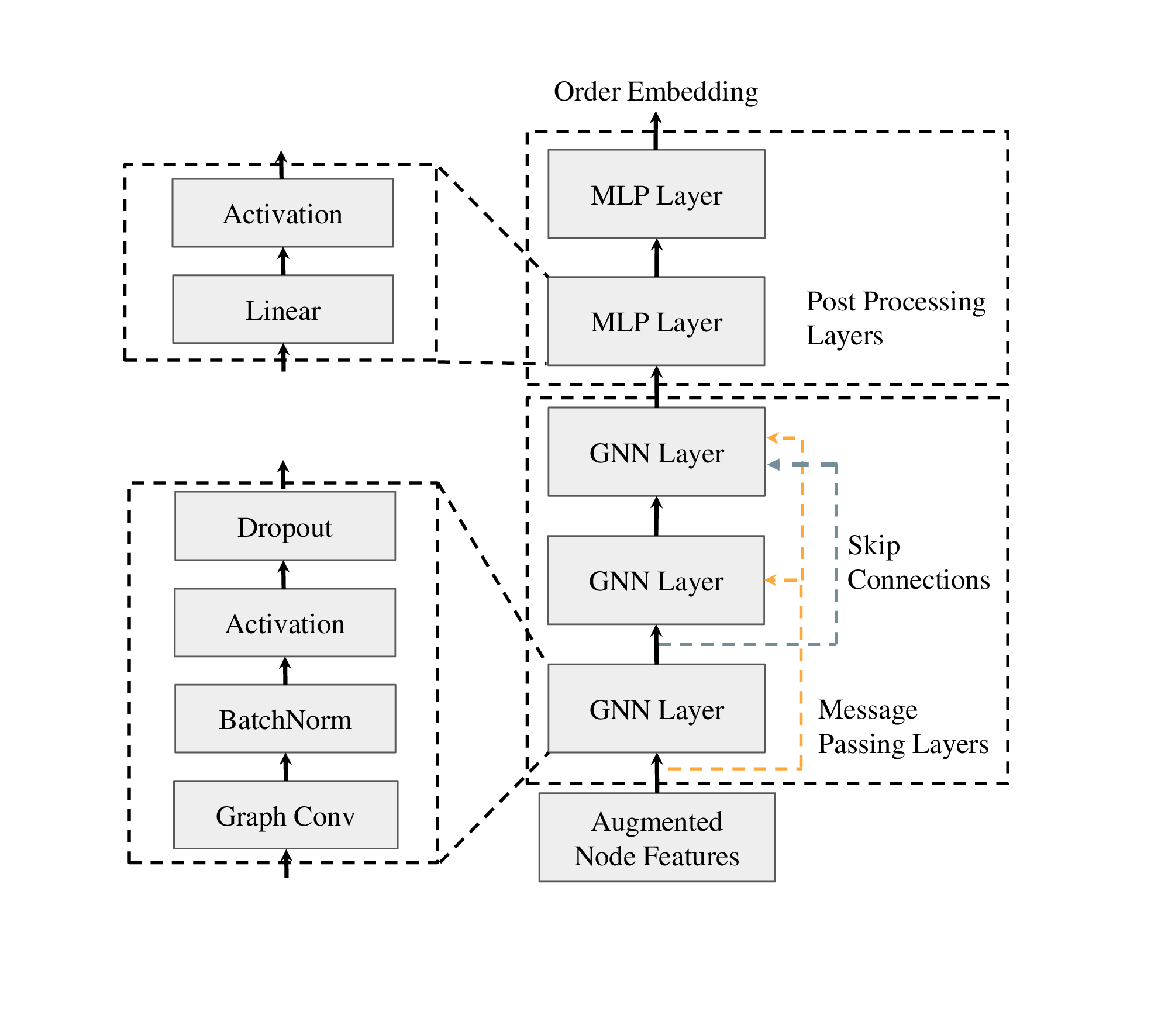}
      \hspace*{-1in}
      \caption{Left: Features which are selected by GNN filter will first go through two MLP layers individually, then concatenate with other features. Right: irredundant features are concatenated with initial node features as augmented node features, which will go through three GNN layers with skip connection and a two-layer MLP. }
    \end{center}
  \end{figure}
\vspace{-0.5cm}
\subsection{model architecture}
\vspace{-0.4cm}
\xhdr{Fea2Fea} For Fea2Fea-single pipeline, we implement two GNN layers followed by two multilayer-perceptron (MLP) layers. Hidden embedding size \textit{h} = 64. The number of bins $\mathcal B$ is set to 6 as default. Model depth $\mathcal D $ is set to 2 as default. We add batch-norm layers when peforming model depth hyperparameter tests. We add a two-layer MLP as our baseline model without graph convolution layers for comparison. For Fea2Fea-multiple pipeline, we observe the optimal graph convolution method from Fea2Fea-single with the same graph model depth and hidden embedding size. Default threshold $\mathcal T$ is set to 0.85, which is regarded as a hyperparameter. For feature concatenation, hidden dimension and number of neurons for NTN are set to 64. 

\xhdr{Application model} GNN based filters will select unredundant structural feature combinations. Each of the features will pass through a two-layer MLP and perform one of three concatenations, followed by simple concatenation with initial node features. Three GNN layers are applied with graph convolution, batch normalization, activation function(\textit{relu}) and dropout layer with droupout probability of 0.6. The hidden dimension is 64 in graph embedding space. Preprocessing layer, which is composed of two-layer MLP is applied after the graph embedding, which is shown in Figure 2.

\xhdr{Binning methods}
According to the distribution of structural features for \textit{Planetoid} dataset, \textit{degree} and \textit{clustering coefficient} features will cause class inbalance(Appendix C). Therefore, we divide the unbalanced data into one bin, and the other data 
into remaining bins, especially more than half of the nodes' \textit{clustering coefficient} are zero. \textit
{NCI1} dataset has a worse situation where most of the nodes in the graph have a zero 
\textit{clustering coefficient}. One solution is to remove this feature from
the feature set.
\vspace{-0.5cm}
\section{Experiments}
\vspace{-0.3cm}
For node datasets, we split the whole datasets into training, validation 
and test dataset with a fixed ratio in original setting \cite{hamilton2017inductive}. For graph datasets, we split graph 
dataset into training and test datasets with a ratio of 8:1 and follow a traditional 10-fold cross validation in training splits \cite{do2021twostage}.

\xhdr{Datasets} Six benchmark datasets are selected.  For node datasets, we choose \textit{Planetoid} \cite{hamilton2017inductive}, which includes {\sc Cora}, {\sc CiteSeer} and {\sc PubMed}. For graph datasets, we choose \textit{TUDataset} \cite{morris2020tudataset}, which is a collection of benchmark datasets including data from small molecules, bioinformatics, social networks, computer vision and 
some synthetic datasets. In this paper, we choose two datasets  {\sc Proteins} and {\sc 
Enzymes} from bioinformatics domain and {\sc Nci1} from molecule domain. After generating graph feature matrix for each dataset, we plot the distribution for each graph feature to search for imbalance(Appendix C). 
\vspace{-0.6cm}
\subsection{Results}
\vspace{-0.6cm}
\begin{table}\scriptsize
  \centering
  \caption{Feature to Feature Prediction on Cora and Proteins Datasets (bins = 6)}
  \begin{tabular}{*{11}{c}} \toprule
{Task}  & \multicolumn{5}{c}{{\sc Cora}} & \multicolumn{5}{c}{{\sc Proteins}} 
\\
\cmidrule(lr){2-6}\cmidrule(lr){7-11}
 & GCN & GIN & SAGE & GAT & MLP  &GCN & GIN & SAGE & GAT & MLP \\ \hline
{\textit{Cons} $\rightarrow$ \textit{Deg}}  &0.509 & \B 1.000 & 0.213 & 0.202 & 0.206 & 0.560 & 
\B0.662 & 0.469 & 0.469 & 0.469 \\
{\textit{Deg} $\rightarrow$ \textit{Deg}}  &0.741 & 1.000 & 0.967& 0.514 &\B 1.000   & 0.633 & 
\B 0.662 & 0.688 & 0.470 & 0.640 \\
{\textit{Clu} $\rightarrow$ \textit{Deg}}   &0.423 & \B 1.000 & 0.504 & 0.474 & 0.285 & 0.537 & 
\B0.652 & 0.482 & 0.469 & 0.467 \\
{\textit{PR} $\rightarrow$\textit{Deg}}   &0.308 & \B 1.000 & 0.311 & 0.223 & 0.197 & 0.430 & 
\B0.662 & 0.446 & 0.469 & 0.469 \\
{\textit{Avglen} $\rightarrow$ \textit{Deg}}   &0.409 & \B 1.000 & 0.499 & 0.228 & 0.274 & 0.522 
& \B0.657 & 0.482 & 0.469 & 0.469 \\
{\textit{Cons} $\rightarrow$ \textit{Clu}}  &0.523 & \B 0.533 & 0.461 & 0.461 & 0.461 & 0.299 & 
\B0.436 & 0.202 & 0.202 & 0.236 \\
{\textit{Deg} $\rightarrow$ \textit{Clu}}  &\B0.550 & 0.542 &  0.548 & 0.506 & 0.531 & 0.435 & 0.387 & \B0.454 & 0.312 & 0.346 \\
{\textit{Clu}  $\rightarrow$ \textit{Clu}}  &0.724 & 0.765 & \B 0.968 & 0.668 & 0.968 & 0.570 & 0.610 & \B0.707 & 0.549 & 0.723 \\
{\textit{PR} $\rightarrow$ \textit{Clu}}  &0.490 & \B 0.538 & 0.486 & 0.461 & 0.461 & 0.247 & \B0.330 & 0.278 & 0.219 & 0.236 \\
{\textit{Avglen} $\rightarrow$ \textit{Clu}}  &0.508 & \B 0.538 & 0.498 & 0.460 & 0.465 & \B0.276 
& 0.258 & 0.328 & 0.202 & 0.273 \\
{\textit{Cons} $\rightarrow$ \textit{PR}}  &0.639 & \B 0.756 & 0.160 & 0.160 & 0.160 & 0.645 & 
\B0.648 & 0.170 & 0.170 & 0.169 \\
{\textit{Deg} $\rightarrow$ \textit{PR}}  &0.573 & \B 0.792 & 0.750 & 0.392 & 0.603 & 0.622 & 0.699 & \B0.722 & 0.239 & 0.461 \\
{\textit{Clu} $\rightarrow$ \textit{PR}}  &0.427 & \B 0.695 & 0.403 & 0.199 & 0.395 & \B0.575 & 0.465 & 0.467 & 0.224 & 0.349 \\
{\textit{PR} $\rightarrow$ \textit{PR}}  &0.345 & \B 0.714 & 0.323 & 0.185 & 0.160 & 0.170 & \B0.403 & 0.251 & 0.170 & 0.176 \\
{\textit{Avglen} $\rightarrow$ \textit{PR}}  &0.450 & \B 0.741 & 0.490 & 0.202 & 0.247 & \B0.579 
& 0.453 & 0.395 & 0.170 & 0.175 \\
{\textit{Cons} $\rightarrow$ \textit{Avglen}}  &0.357 & \B 0.384 & 0.169 & 0.169 & 0.169 & \B0.182 & 0.171 & 0.171 & 0.171 & 0.171 \\
{\textit{Deg} $\rightarrow$ \textit{Avglen}}  &0.420 & 0.435 & \B 0.440 & 0.340 & 0.199 & \B0.212 
& 0.184 & 0.200 & 0.171 & 0.175 \\
{\textit{Clu} $\rightarrow$ \textit{Avglen}}  &0.286 & \B 0.310 & 0.263 & 0.219 & 0.202 & 0.227 & 0.196 & \B0.254 & 0.228 & 0.216 \\
{\textit{PR} $\rightarrow$ \textit{Avglen}}  &0.215 & \B 0.421 & 0.266 & 0.185 & 0.169 & 0.171 & 
\B0.172 & 0.171 & 0.171 & 0.171 \\
{\textit{Avglen} $\rightarrow$ \textit{Avglen}}  &0.503 & 0.445 & 0.774 & 0.490 & \B 0.958 & 0.549 & 0.483 & \B0.612 & 0.545 & 0.513 \\\hline
\bottomrule
  \end{tabular}
\end{table} 
\vspace{-1cm}
\xhdr{Model performance} In terms of node datasets, \textit{GIN} overall performs the best among all graph embedding method based models and baseline MLP model when performing Fea2Fea-single. Especially when predicting \textit{degree}, \textit{GIN}  can reach 100\% average test accuracy. \textit{GraphSAGE} and \textit{GCN} can predict features at an acceptable level but they are far from the performance of \textit{GIN}. The reason might be that GIN is focused on structural features, which leads to higher accuracy in graph classification problems.  \textit{GAT} does not perform well in most of the tasks. \textit{MLP} can achieve self-prediction, but it is poor at mutual prediction. On the view of graph datasets, \textit{GIN} performs the best while performance of \textit{GraphSAGE} and \textit{GCN} is much better than the performance on node datasets. Results of \textit{GAT} and \textit{MLP} are close to the optimal value at some tasks but generally can not reach the performance provided by \textit{GIN}. When it comes to the real world applications, improvements are made in graph datasets after we add irredundant features(Table 2), with an average improved accuracy of 4.8\%. It does not improve a lot in node datasets. The reason might be attributed to the high dimensionality of {\sc{Cora}} and {\sc{Citeseer}} dataset.

\xhdr{Difficulty of prediction} From table 1 or table 5, we figure out that  \textit{degree} is the easiest to predict. Predicting \textit{clustering coefficient} and \textit{average path length} is hard generally for both kinds of datasets. From the difficulty of prediction, we can also explain the property of each feature. For example, \textit{degree} feature is a basic feature other than \textit{constant feature} which is easier to predict from others. However, \textit{clustering coefficient} is sparse while \textit{average path length} and \textit{pagerank} require the entire information from the whole graph which may bring redundant information. The choices of Neural Tensor Networks or just simple or bilinear concatenation are discussed in Appendix C. Overall it depends on the classification tasks. 

\xhdr{Average accuracy of multi-to-one prediction}
We take the simple concatenation to illustrate an example. When predicting \textit{clustering coefficient} from other features in Planetoid dataset, possible combination number of features is only equal to 2 or 3. For Citeseer dataset, a 3-set input is better than a 2-set input. A possible 3-set concatenation for Cora dataset is \{\textit{Cons}, \textit{PR}, \textit{Avglen}\}, which is much better than a 1-set prediction for each of the feature in this 3-set. When predicting \textit{Pagerank}, the prediction accuracy is not stable. A 3-set feature combination is worse than a 2-set feature combination for {\sc{PubMed}} and {\sc{Citeseer}} and even worse than Fea2Fea-single. In \textit{TUdataset}, predicting \textit{pagerank} and \textit{average path length}
might exist the situation of 4-set feature set, but this 4-set feature set does not perform well and the average accuracy is descending. Figures are shown in appendix C(figure 9).
  \vspace{-0.6cm}
  \begin{table}\scriptsize
    \centering
    \caption{Performance on node and graph classification problems (average test accuracy in \%), where s,b,n are the abbreviations of "simple", "bilinear" and "NTN" methods respectively, followed by the number of irredundant features. A baseline Fea2Fea-null is based on GCN on node datasets and on GIN when it comes to graph datasets.}
    \vspace{-0.3cm}
    \begin{tabular}{*{7}{c}} \toprule
  {Model}  & \multicolumn{1}{c}{{\sc Cora}} & \multicolumn{1}{c}{{\sc Citeseer}} & \multicolumn{1}{c}{{\sc Pubmed}}  & \multicolumn{1}{c}{{\sc Enzymes}} & \multicolumn{1}{c}{{\sc Proteins}}  & \multicolumn{1}{c}{{\sc NCI1}}   \\ \hline

  MLP           & 59.0 $\pm$ 1.0& 59.6 $\pm$ 0.5 & 70.7 $\pm$ 2.2  & 32.7 $\pm$  5.2 & 65.9 $\pm$ 3.6   & 58.0 $\pm$ 0.8   \\
  GraphSAGE       &  76.0 $\pm$ 4.4&  66.8 $\pm$ 2.8 & 73.5 $\pm$ 1.5 & 37.8 $\pm$ 3.0  &  66.2 $\pm$ 2.5  & 64.7 $\pm$ 2.3   \\
  GCN             & 80.0 $\pm$ 0.9 & 68.1 $\pm$ 0.9 & 74.2 $\pm$ 1.1 & 36.0 $\pm$ 5.0  &  66.2 $\pm$ 0.8  & 61.3 $\pm$ 0.9  \\
  GAT             & 79.7 $\pm$ 1.2 & 69.2 $\pm$ 0.9 &  74.0 $\pm$ 1.4 &  31.0 $\pm$ 5.6  &   65.9 $\pm$ 2.4  & 60.9 $\pm$ 2.2 \\          
  Fea2Fea-null    & 80.0 $\pm$ 0.9 & 68.1 $\pm$ 0.9 & 74.2 $\pm$ 1.1 & 47.2 $\pm$ 3.2 & 67.9 $\pm$ 1.2  & 71.8  $\pm$ 0.6 \\ \hline
  Fea2Fea-s2       & 79.7 $\pm$ 0.8 &  65.0 $\pm$ 2.4 & 77.4 $\pm$ 0.8 & \B48.5 $\pm$ 4.5  &\B77.8 $\pm$ 0.9  &  74.2 $\pm$ 0.8       \\
  Fea2Fea-b2   & 77.3 $\pm$  3.1 & 64.7 $\pm$ 4.0 &77.3 $\pm$ 1.1& 45.8 $\pm$ 3.2  & 76.4 $\pm$ 1.2  & 70.8 $\pm$ 2.6\\
  Fea2Fea-n2  & 77.0 $\pm$  1.6 & 62.6 $\pm$ 4.0 & 75.7 $\pm$ 2.6  & 42.8 $\pm$ 3.4 & 74.9 $\pm$ 2.7 & 68.5 $\pm$ 0.7\\
  Fea2Fea-s3       & 79.6 $\pm$ 0.9  &66.2 $\pm$  1.8 & \B78.5 $\pm$ 1.8  &  \B48.0 $\pm$ 4.4  & 76.8 $\pm$ 1.4 & \B74.9 $\pm$ 0.9   \\           \hline

  \bottomrule
    \end{tabular}
  \end{table} 

\section{Conclusions and future works}
\vspace{-0.4cm}
In this paper, we introduce Fea2Fea-single to analyze correlations between structural features. We implement Fea2Fea-multiple to remove redundant features. We take the advantage of enriched graph embedding methods and feature concatenation to complete the experiments. Experiments show that adding structural features with GNN based pre-selection is necessary before training. In our future works, we plan to add more graph features and large-scale Open Graph Benchmark(OGB) datasets to enrich feature sets and to sum up a commonality on structural feature correlations. Graph embedding methods will also be enriched
to make sure that \textit{GIN} is the best graph embedding method in Fea2Fea-single. 
The conclusion of low self-prediction on pagerank is still under controversy which requires further explorations.
\bibliographystyle{splncs04}
\bibliography{references}

\appendix
\newpage
\section{Algorithm Pseudocode}
\vspace{-1cm}
\begin{algorithm}[!htp]
  \caption{Get Feature Correlation Matrix}
  \KwIn{full-scale feature matrix 
    $\mathbf{x}_{\mathcal{G}} $;
    model architecture $\mathcal M$; metrics $\mathcal P$ }
  \KwOut{ Feature correlation matrix $\mathcal R
$}
  $\mathcal R \gets \textbf{0}, \mathcal K \gets 5$\\
  \For{$i \gets 1$ \textbf{to} $\mathcal K$}{
      $\textbf{I}_{\mathcal G} \gets \textbf{x}_{\mathcal G}(:,i)$\\
  
      \For{ $j \gets 2$ \textbf{to} $\mathcal K$
      }{
      $\textbf{O}_{\mathcal G} \gets \textup{Binning}\left(\textbf{x}_{\mathcal G}(:,j)
\right)$\\
      $\textbf{O}^{'}_{\mathcal G} \gets$ $\mathcal{M}\left(\textbf{I}_{\mathcal G}, 
\textbf{O}_{\mathcal G}\right)$\\
      $\mathcal{R}(i,j) \gets \mathcal P(\textbf{O}^{'}_{\mathcal G},\textbf{O}_
{\mathcal G} )$
      }
      \If{i == j == $\mathcal{K}$}{
        \Return{$\mathcal R$}\;
        }
  }
  \end{algorithm}
\vspace{-0.8cm}
The case when j is equal to 1 is not shown in the description of pseudocode, but as we have mentioned in the main paper, we should not ignore the case of predicting constant feature. In real practice, the first column is symmetrical to the first row to simplify analysis. For example, predicting constant feature from node degree is equivalent to predicting node degree from constant feature.
\vspace{-0.5cm}
\begin{algorithm}[!htp]
  \caption{Get $\mathcal A_{\mathcal G}$ which records multiple feature to 
single 
feature prediction results}
  \KwIn{ 
Feature Correlation Matrix   $\mathbf{x}_{\mathcal{G}}$; model architecture $\mathcal M'$; metrics $\mathcal P$; output 
feature 
index \textbf{idx}; threshold $\mathcal T = 0.85$ }
  \KwOut{ Array $\mathcal A_{\mathcal G}$}
  Initialize $Comb \gets \{\{0,1\},...,\{0,1,2,3,\mathcal 
K\}\}$ \\

  \For{$comb$ $\in$ $Comb$}{
      \For{$ {\forall}$ $\mathcal F_{i}, F_{j}$ $\in$ $comb$}{
        \If{$\mathcal R(F_{i},F_{j})$ $\geq$ $\mathcal T$ or $\mathcal R(F_{j},F_{i})$ $\geq$ $\mathcal 
T$ or  
\textbf{idx} in comb }{
          Remove $comb$ from $Comb$;
        }
      }
    }
  %Achieve all possible concatenation and perform feature concatenation\\
  \For{$comb$ $\in$ $Comb$}{
      $\textbf{I}_{\mathcal G} \gets \textbf{x}_{\mathcal G}(:,comb)$\\
      $\textbf{O}_{\mathcal G} \gets \textbf{x}_{\mathcal G}(:,\textbf{idx})$\\
      $\textbf{O}^{'}_{\mathcal G} \gets$ $\mathcal{M}^{'}\left(\textbf{I}_
{\mathcal G}, 
\textbf{O}_{\mathcal G}\right)$\\
      $\mathcal A_{\mathcal G}(comb | \textbf{idx}) \gets \mathcal P(\textbf{O}
^{'}_
{\mathcal G},\textbf{O}_{\mathcal G} )$\\

  }
  \end{algorithm}
  
\vspace{-0.8cm}
Concatenation method is included in model $\mathcal M'$, which is not shown in the pseudocode but is instead shown in figure 2. We should also notice that the generation of total combination is not efficient if we consider adding large more structural features. One possible solution is to use reservoir sampling and do add-drop iterations until all pairs of feature correlation are under the threshold.

  \section{Hyper-parameter tunning}
  \vspace{-0.3cm}
  Three hyperparameters are the number of bins, depth of graph convolution layers and threshold. We evaluated the influence of hyperparameters on \textit{Citeseer} and \textit{ENZYMES} datasets. Tasks are: from \textit{pagrank} predict \textit{average path length} and from \textit{average path length} predict \textit{clustering coefficient}. From the results of experiments, we find out that average accuracy is high when there're only two bins since the partition of classes is not strict. When the number of bins increases, the average prediction accuracy is descending. When model depth is shallow, the prediction effect is not good. Increasing the number of graph embedding layers with batch normalization will result in a better test generalization. We implement \textit{SkipLayerGNN} \cite{li2019deepgcns} when layers are deep($\geq$3). 

  \begin{center}
\vspace{-0.5cm}
    \begin{table}[!htp]\scriptsize
      \centering
      \caption{Hyper-parameter tests on node and graph datasets}
      \begin{tabular}{*{6}{c}} \toprule
    {Tasks} & Param  & \multicolumn{2}{c}{{\sc Citeseer}} & \multicolumn{2}{c}{{\sc Enzymes}} \\
    \cmidrule(lr){3-4}\cmidrule(lr){5-6}
    & order  & PR$\rightarrow$AvgLen & AvgLen$\rightarrow$Clu  & PR$\rightarrow$AvgLen & AvgLen$\rightarrow$Clu \\ \hline
    
    \multirow{8}{4em}{Bins} 
    &2 & \B 0.774 $\pm$ 0.008 & \B0.836 $\pm$ 0.003 &\B 0.856 $\pm$ 0.028 & \B0.550 $\pm$ 0.001 \\
    &3 & 0.727 $\pm$ 0.006 & 0.781 $\pm$ 0.003 & 0.751 $\pm$ 0.007 & 0.433 $\pm$ 0.058\\
    &4 & 0.649 $\pm$ 0.010 & 0.754 $\pm$ 0.008 & 0.642 $\pm$ 0.016 & 0.327 $\pm$ 0.053\\
    &5 & 0.616 $\pm$ 0.009 & 0.722 $\pm$ 0.004 & 0.551 $\pm$ 0.004 & 0.284 $\pm$ 0.023\\
    &6 & 0.584 $\pm$ 0.006 & 0.719 $\pm$ 0.002 & 0.400 $\pm$ 0.157 & 0.294 $\pm$ 0.044\\
    &7 & 0.558 $\pm$ 0.008 & 0.711 $\pm$ 0.003 & 0.390 $\pm$ 0.130 & 0.246 $\pm$ 0.002\\
    &8 & 0.505 $\pm$ 0.006 & 0.703 $\pm$ 0.003 & 0.306 $\pm$ 0.158 & 0.251 $\pm$ 0.021\\
    &9 & 0.491 $\pm$ 0.010 & 0.696 $\pm$ 0.002 & 0.293 $\pm$ 0.122 & 0.262 $\pm$ 0.026\\
    &10 & 0.478 $\pm$ 0.008 & 0.694 $\pm$ 0.003 & 0.180 $\pm$ 0.119 & 0.249 $\pm$ 0.024\\\hline
    \bottomrule 
    \multirow{4}{4em}{Depth} 
    &2 & 0.579 $\pm$ 0.005 & 0.720 $\pm$ 0.003 & \B0.505 $\pm$ 0.007 &  0.281 $\pm$ 0.038\\
    &4 & 0.387 $\pm$ 0.057 & 0.713 $\pm$ 0.004 & 0.237 $\pm$ 0.021 &  0.330 $\pm$ 0.050\\
    &6 & 0.512 $\pm$ 0.055 & 0.707 $\pm$ 0.007 & 0.275 $\pm$ 0.032 &  0.330 $\pm$ 0.038\\
    &8 & 0.638 $\pm$ 0.059 & 0.709 $\pm$ 0.006 & 0.348 $\pm$ 0.041 &  0.336 $\pm$ 0.030\\
    &10 & \B0.651 $\pm$ 0.042 & \B0.722 $\pm$ 0.007 & 0.370 $\pm$ 0.059 & \B0.347 $\pm$ 0.013\\\hline

    \bottomrule
      \end{tabular}
    \end{table}
    \end{center}
    \vspace{-1.2cm}
    We set two thresholds: 0.6 and 0.8, which are based on Fea2Fea-multiple. Experiments are executed 10 times. We obtain average accuracy from the experiments. When threshold value is equal to 0.6, valid combined input feature sets are: (\textit{Cons},\textit{AvgLen}). Mean accuracy is higher than the four combinations when threshold is equal to 0.8. For \textit{ENZYMES} dataset, a higher threshold helps to obtain better results. Therefore, it infers that more feature combinations are not always the best which shows that we must try all combinations.
    \vspace{-0.8cm}
     \begin{center}
      \begin{table}[!htp]\scriptsize
        \centering
        \caption{Threshold tests on node and graph datasets on predicting pagerank, the number in 
combination is in-ordered feature serial number as we mentioned before .}
        \label{tab:commands}
        \begin{tabular}{*{6}{c}} \toprule
      {Tasks} & Param  & \multicolumn{2}{c}{{\sc Citeseer}} & \multicolumn{2}{c}{{\sc Enzymes}} \\
      \cmidrule(lr){3-4}\cmidrule(lr){5-6}
      & order  & Valid Combination & Accuracy  & Valid Combination & Accuracy  \\ \hline
      
      \multirow{2}{4em}{Threshold} 
      &0.6 & (1,5)&  \B0.611 $\pm$ 0.039 & (1,3),(1,5),(3,5),(1,3,5) & 0.405 $\pm$ 0.081 \\
      &0.8 & (1,3),(1,5),(3,5),(1,3,5)&  0.534 $\pm$ 0.067 &(1,3),(1,5),(2,3),(3,5),(1,3,5)) & 
\B0.440 $\pm$ 0.079 \\
      \hline
      
      \bottomrule
        \end{tabular}
      \end{table}
      \end{center}
\vspace{-1cm}
\section{Visualization}
We plot the distribution for each structural feature in each dataset. \textit{Pagerank} and \textit{Average Path Length} is normally distributed for most datasets. Degree and clustering coefficient have shown imbalance of data for most datasets. Therefore, binning method plays an important role in partitioning resonable classes.
  \begin{figure}
    \centering
    \begin{center}
     \hspace*{-1.5in}
        \includegraphics[width=0.27\linewidth]{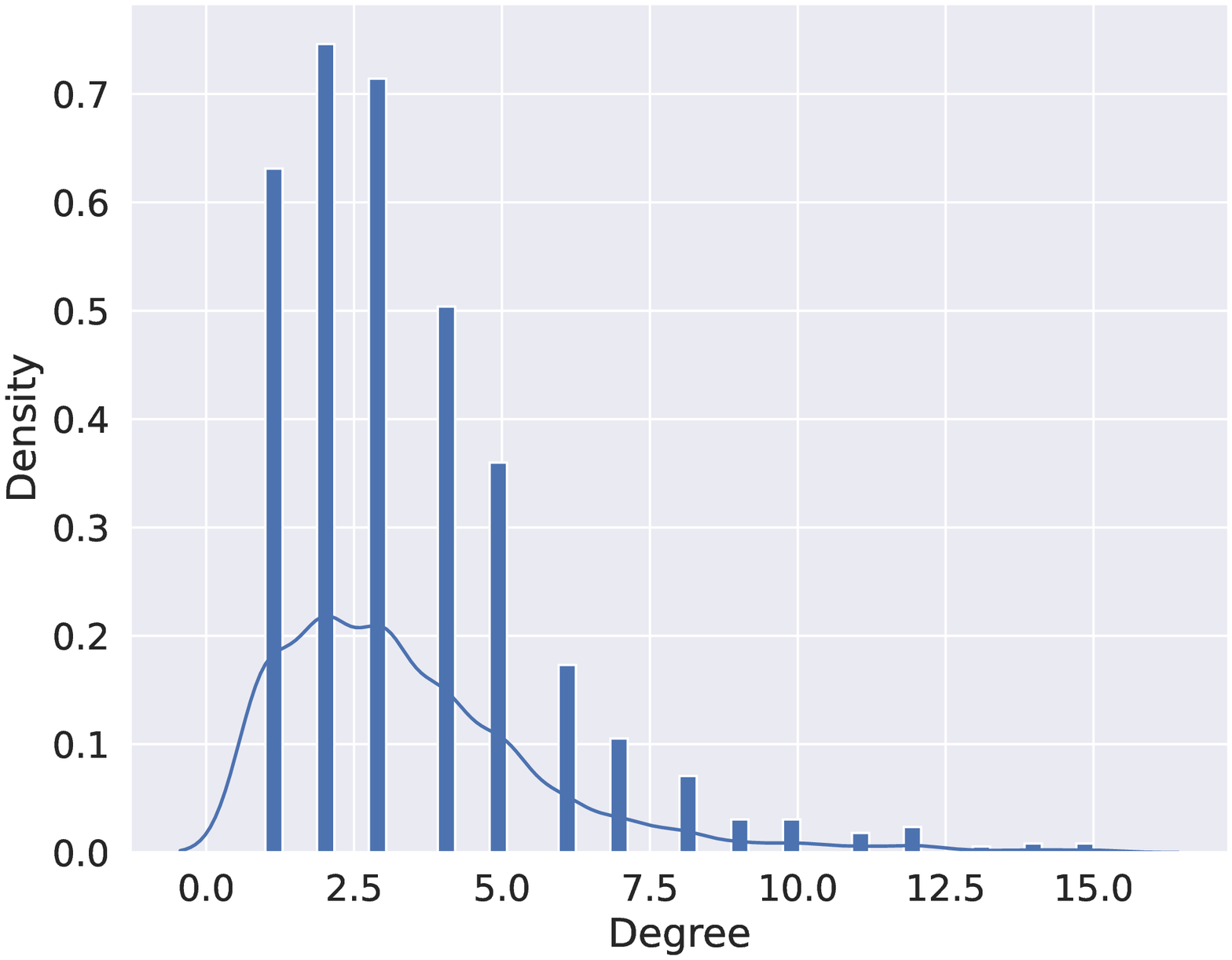}
        \includegraphics[width=0.27\linewidth]{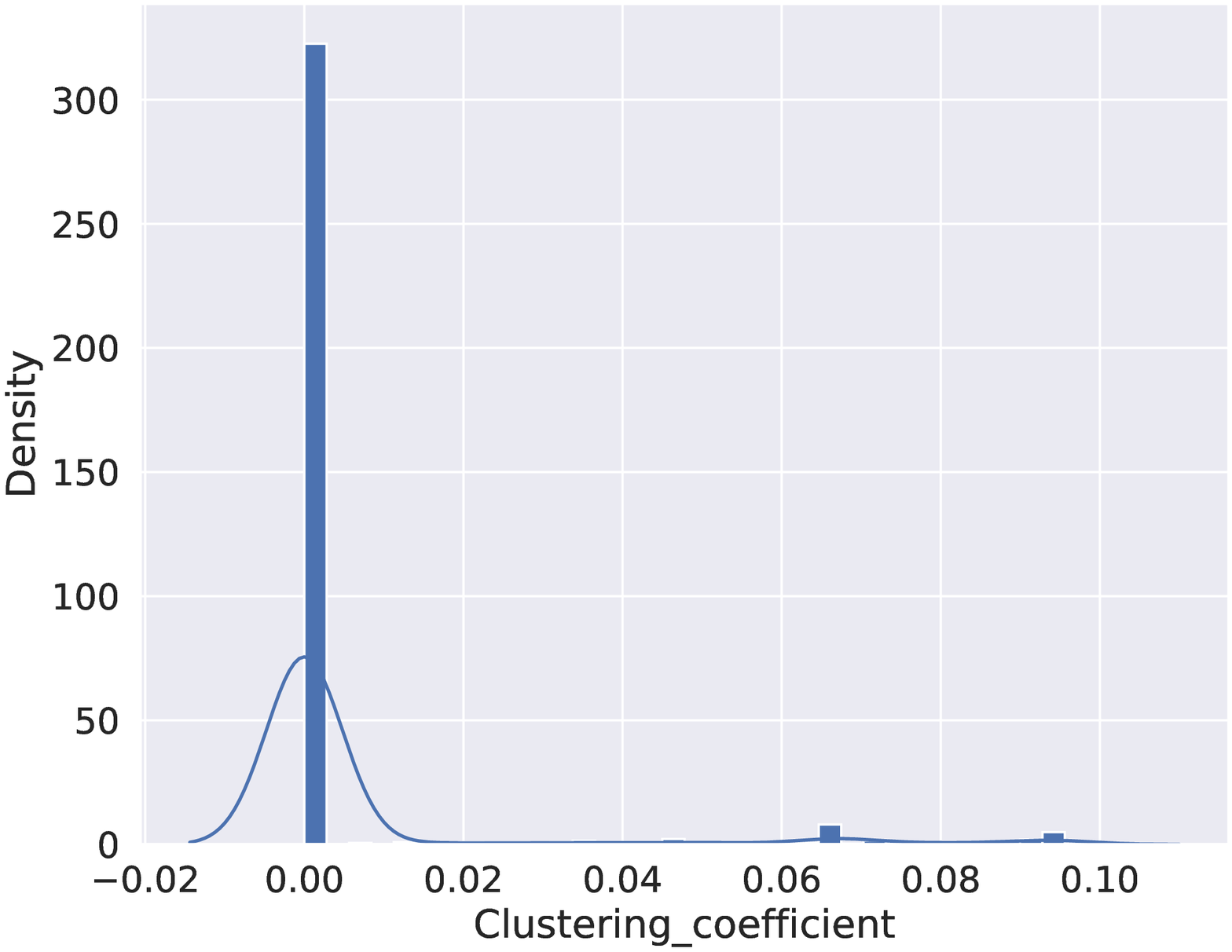}
        \includegraphics[width=0.27\linewidth]{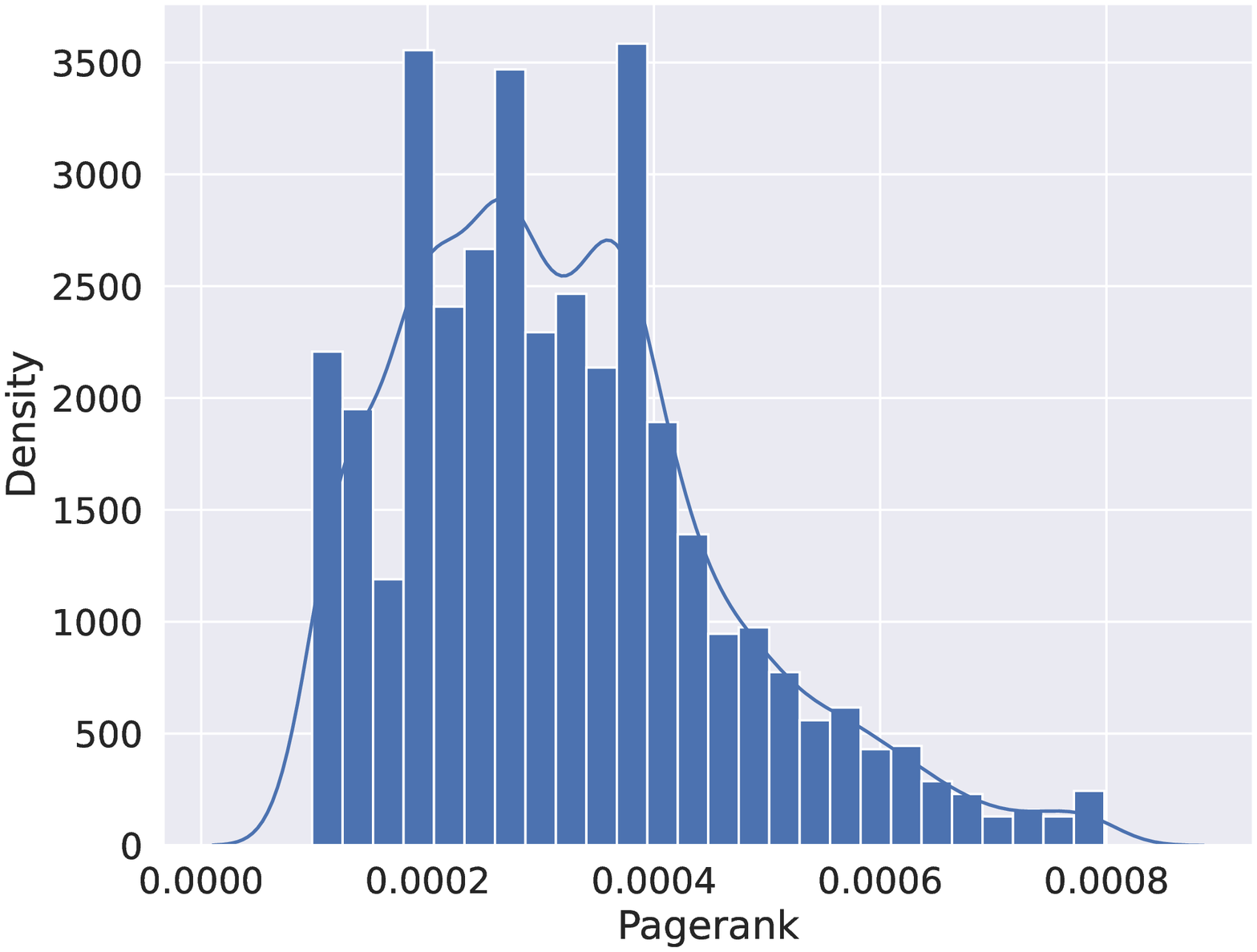}
        \includegraphics[width=0.27\linewidth]{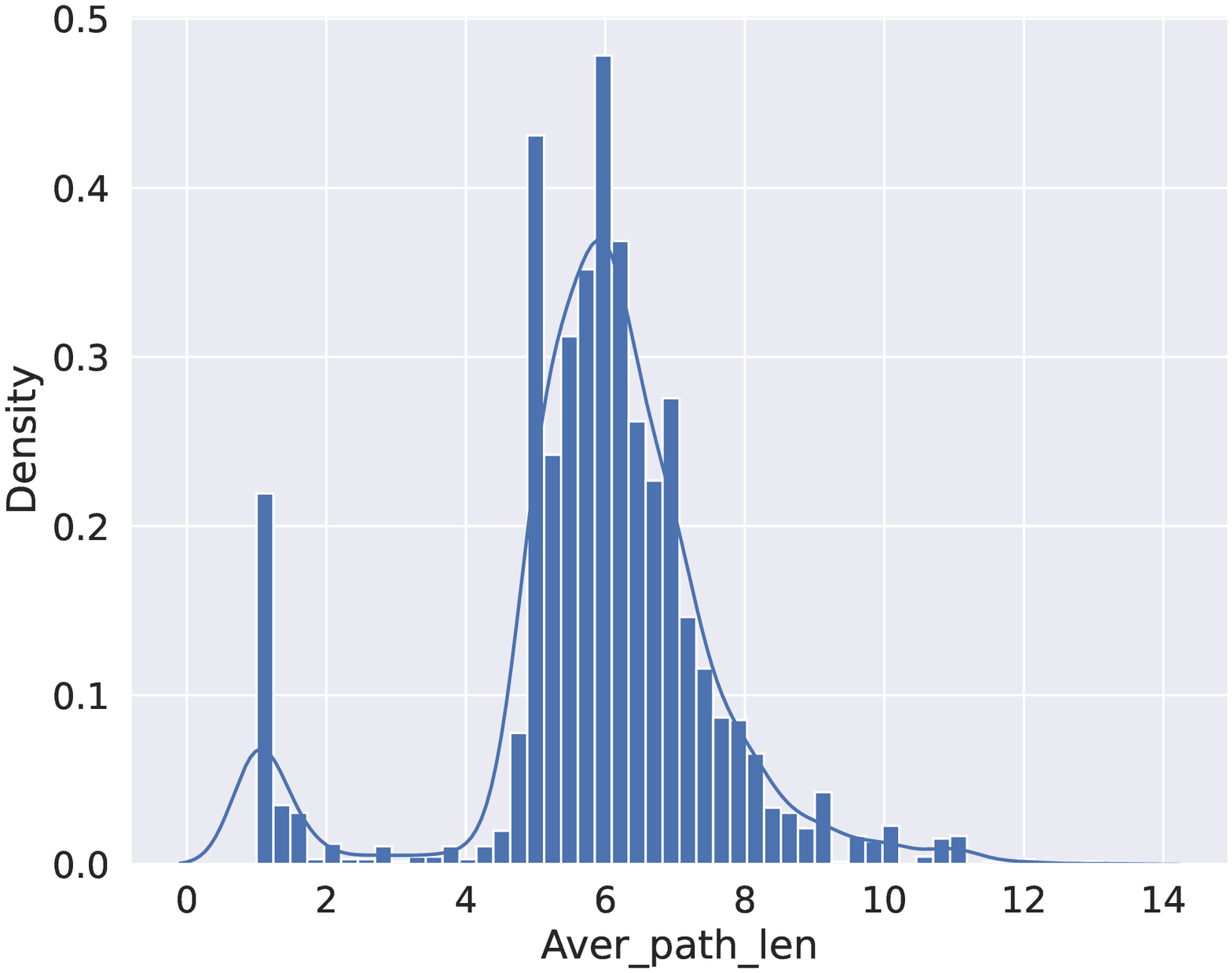}
        \hspace*{-1.5in}
        \caption{Distribution of graph feature of Cora Dataset}
    \end{center}
      \centering
      \begin{center}
       \hspace*{-1.5in}
          \includegraphics[width=0.27\linewidth]{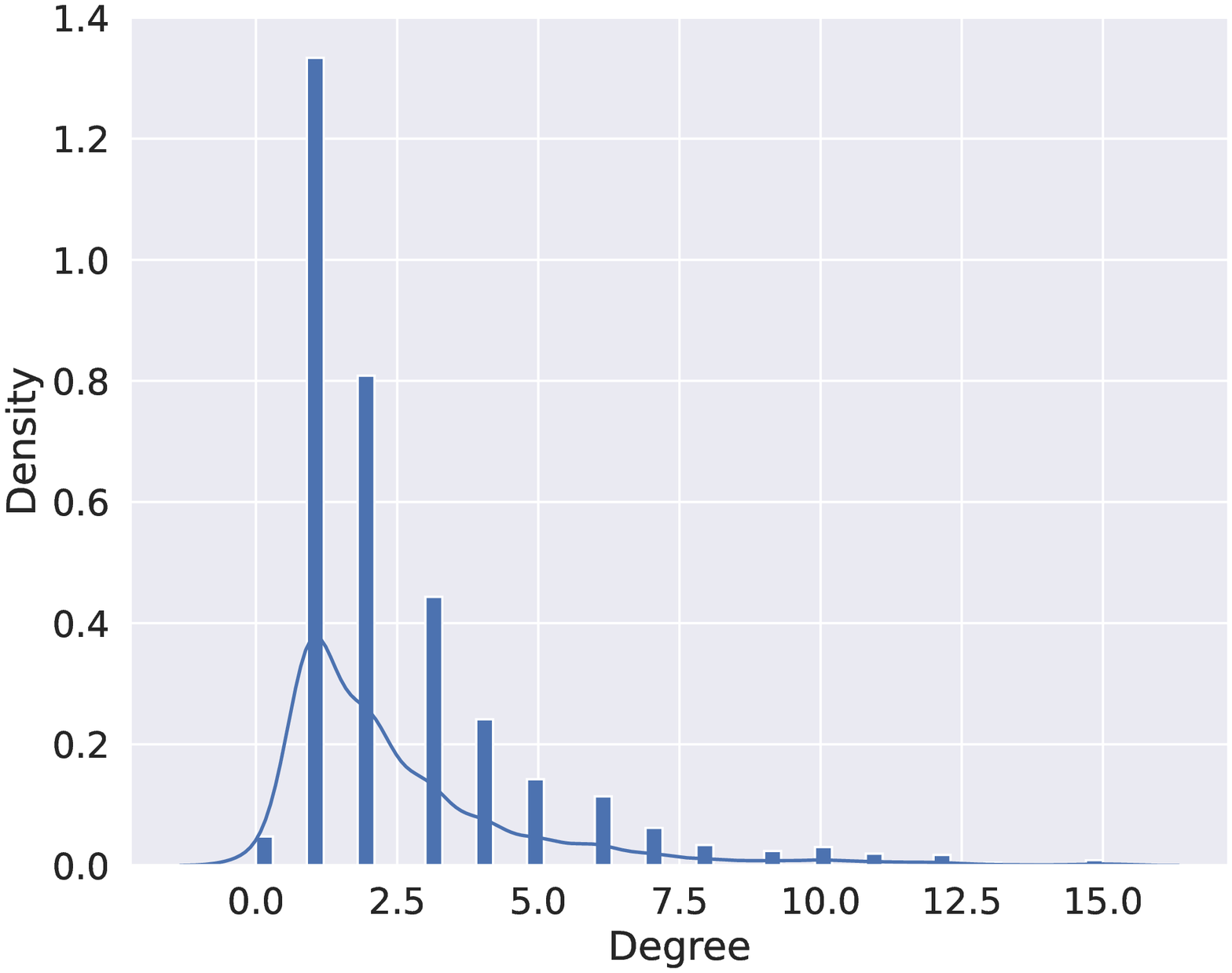}
          \includegraphics[width=0.27\linewidth]{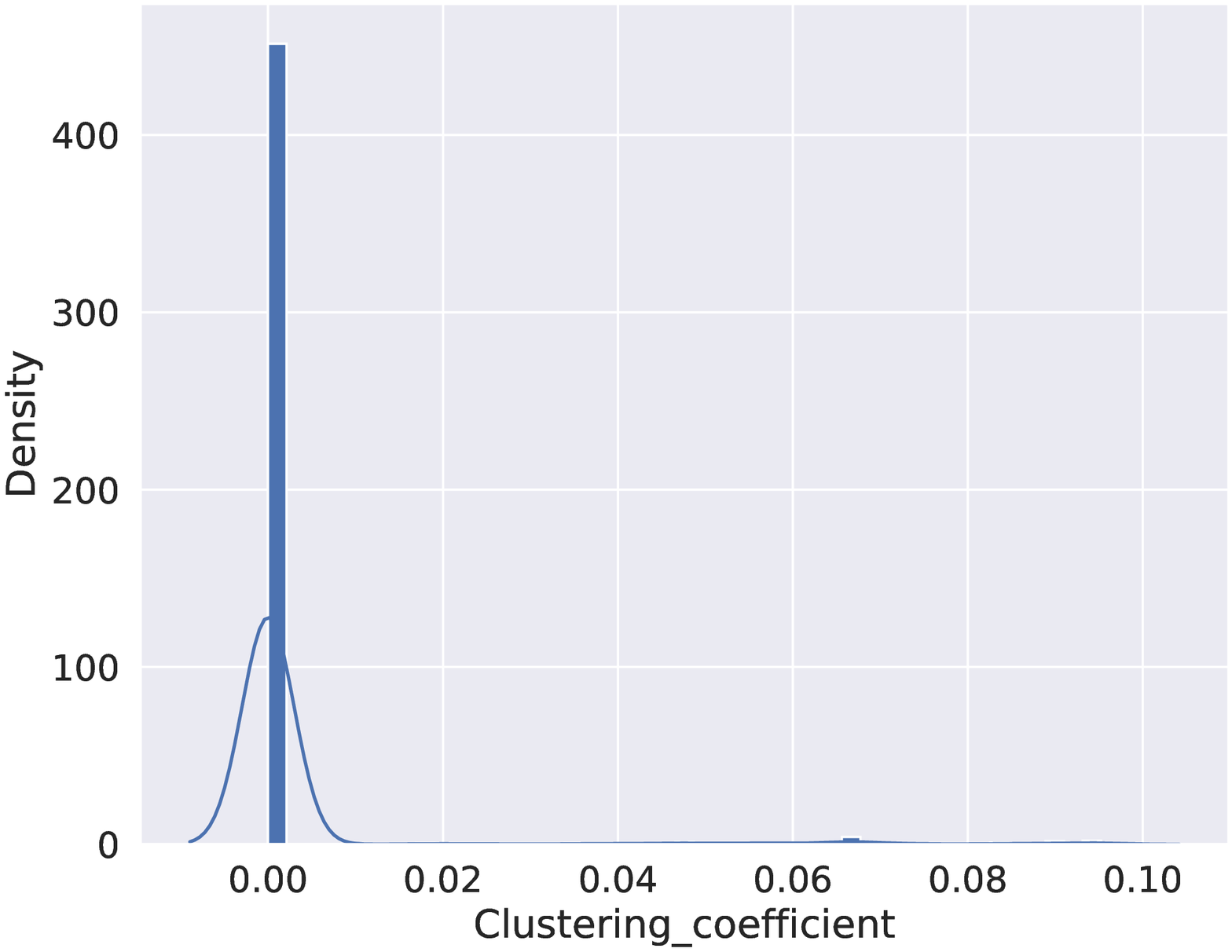}
          \includegraphics[width=0.27\linewidth]{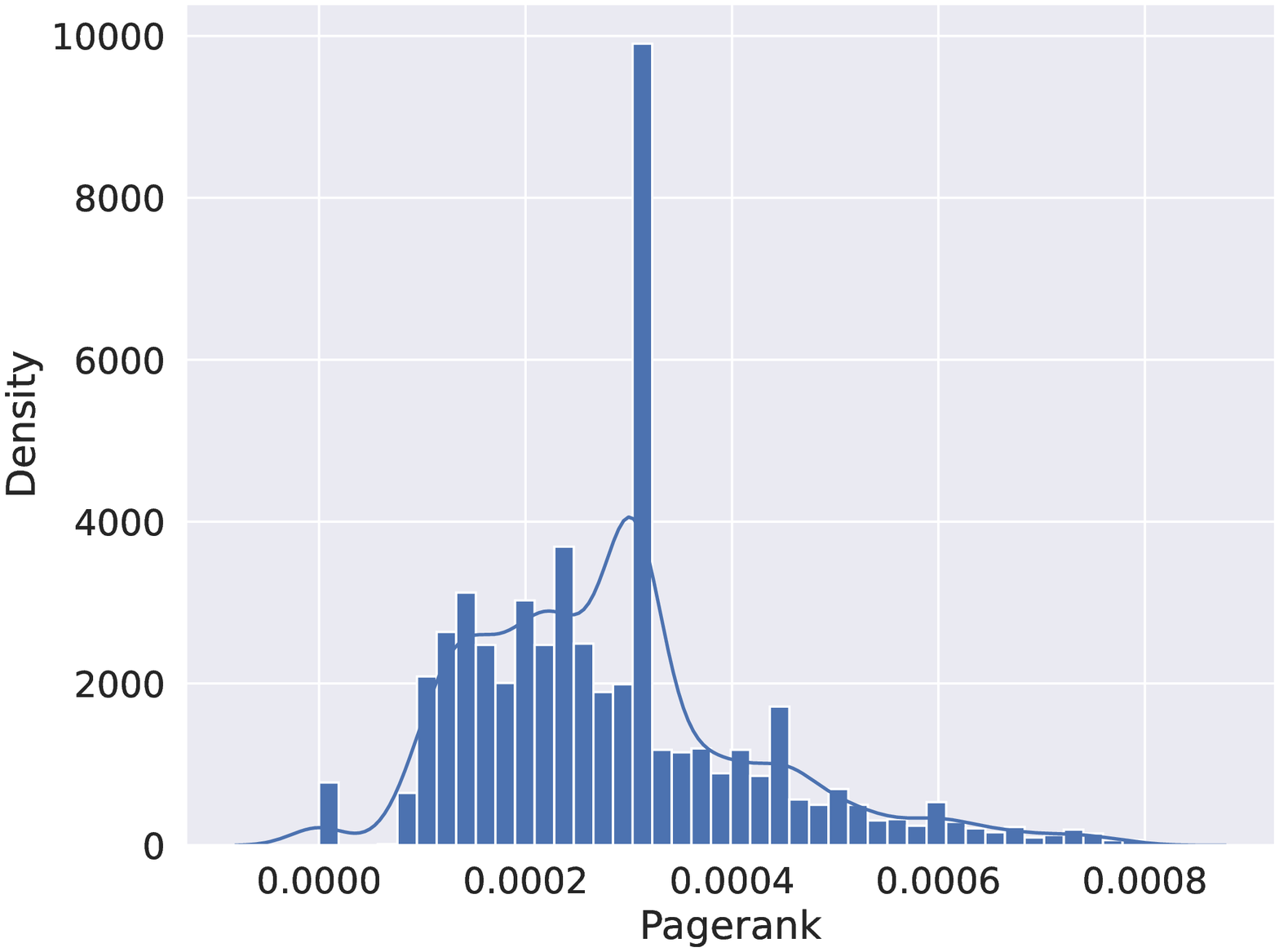}
          \includegraphics[width=0.27\linewidth]{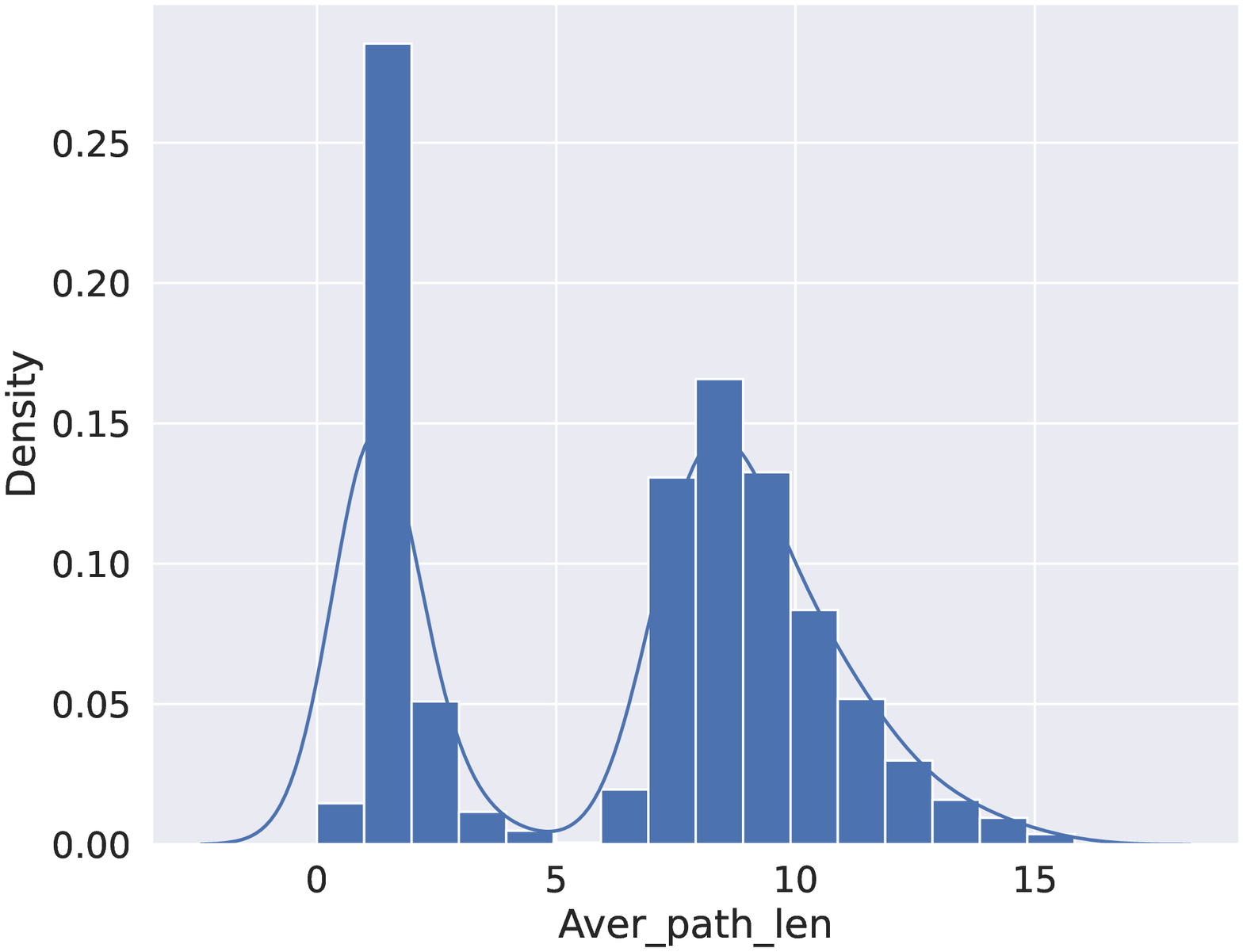}
          \hspace*{-1.5in}
          \caption{Distribution of graph feature of Citeseer Dataset}
      \end{center}

        \centering
        \begin{center}
         \hspace*{-1.5in}
            \includegraphics[width=0.27\linewidth]{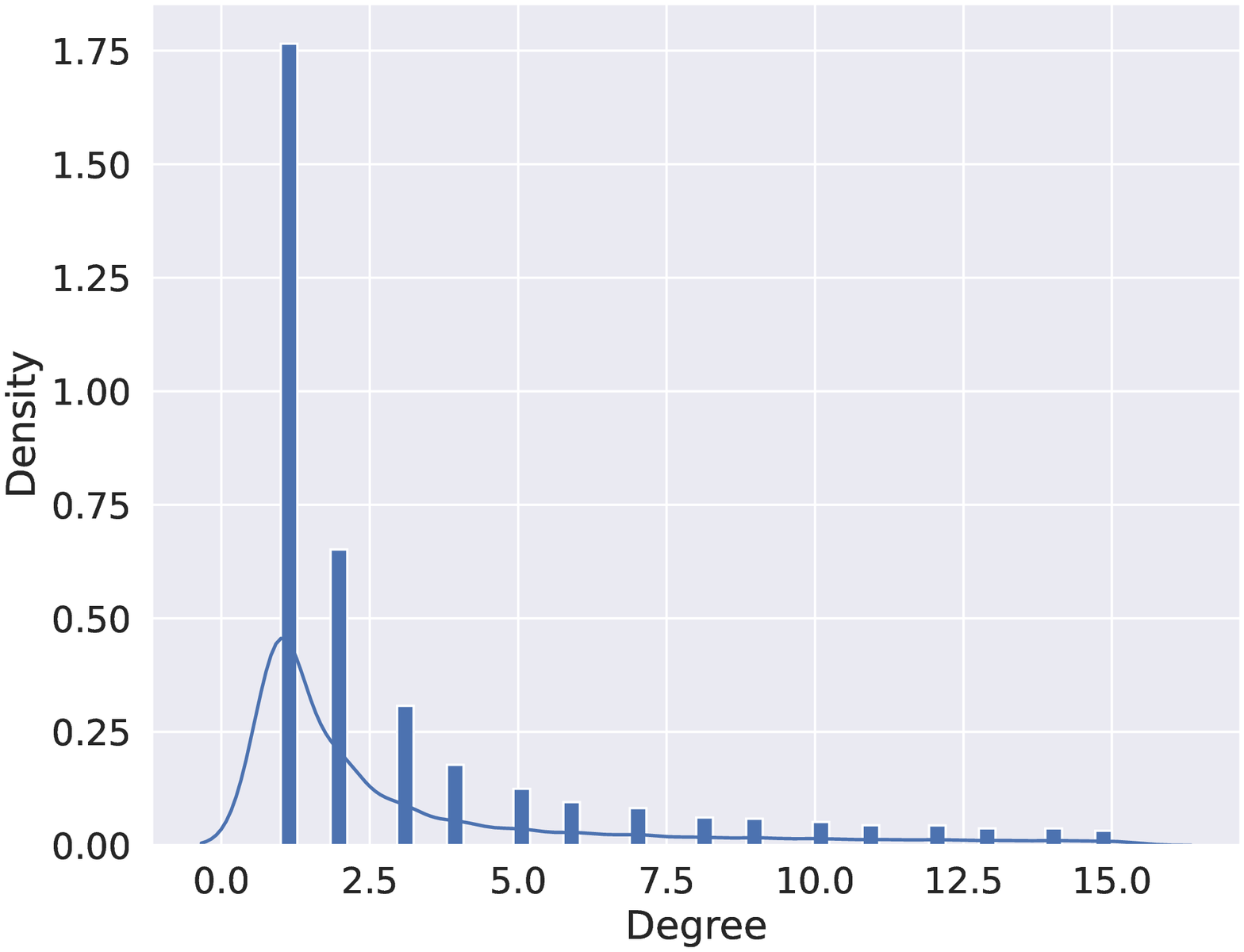}
            \includegraphics[width=0.27\linewidth]{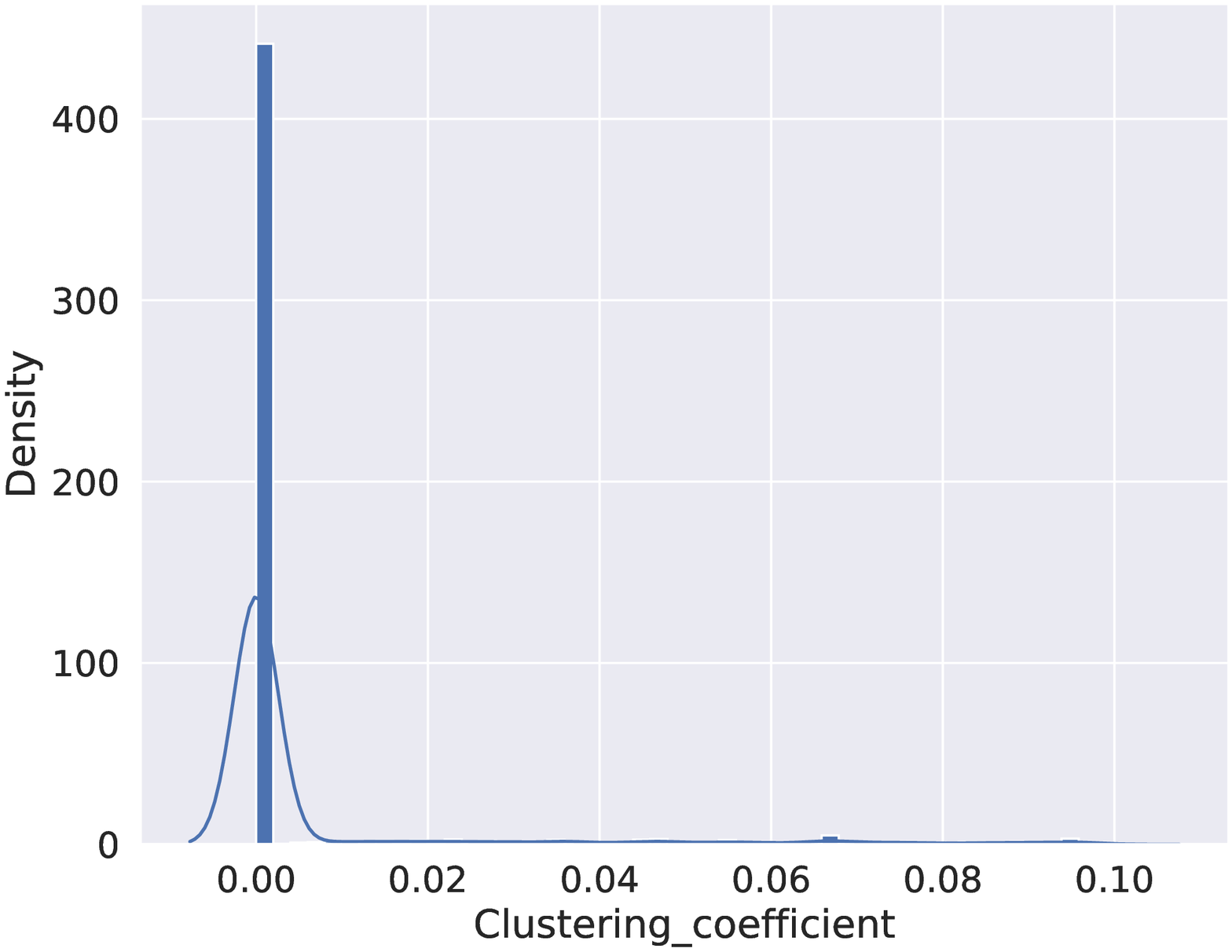}
            \includegraphics[width=0.27\linewidth]{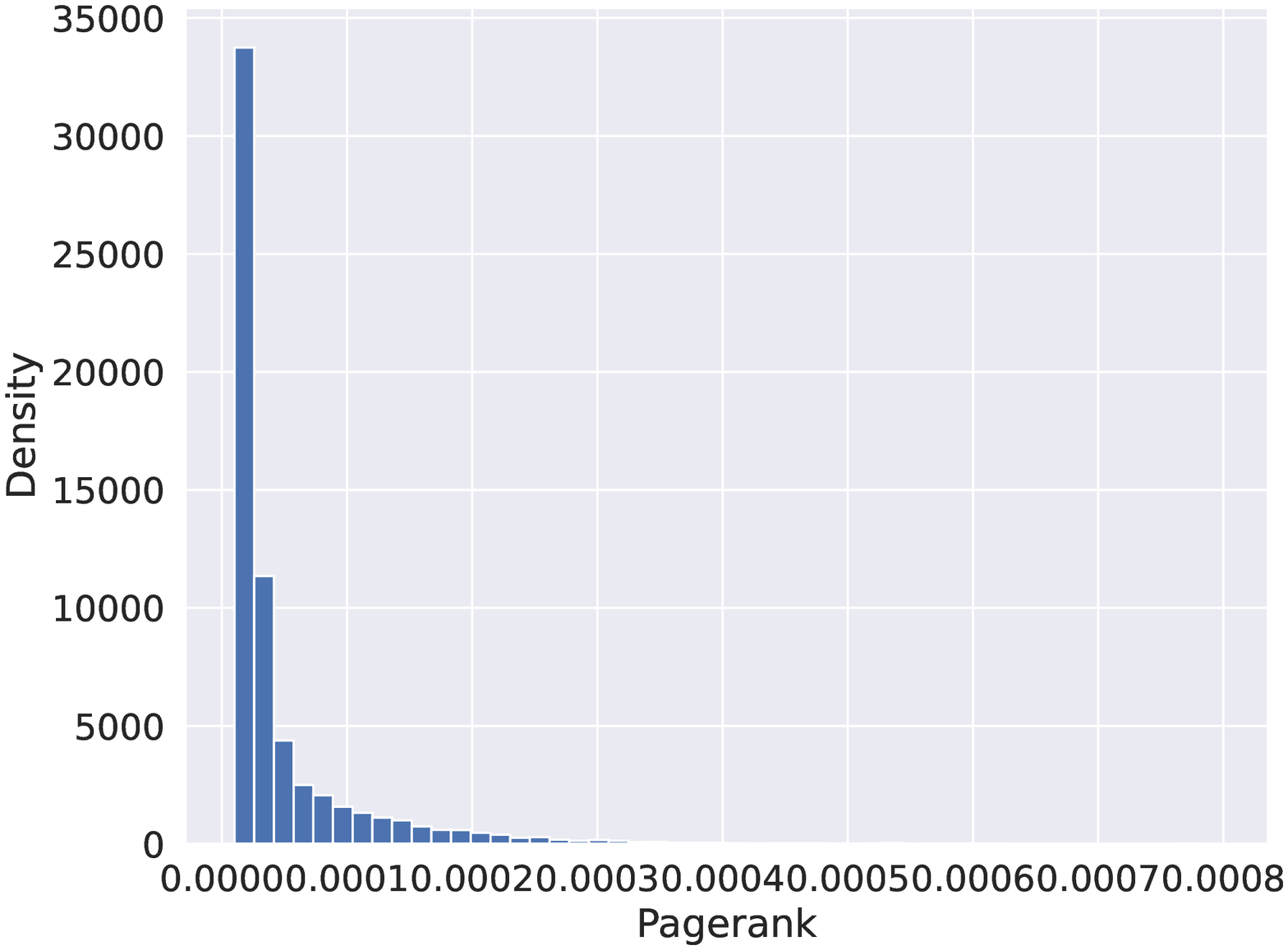}
            \includegraphics[width=0.27\linewidth]{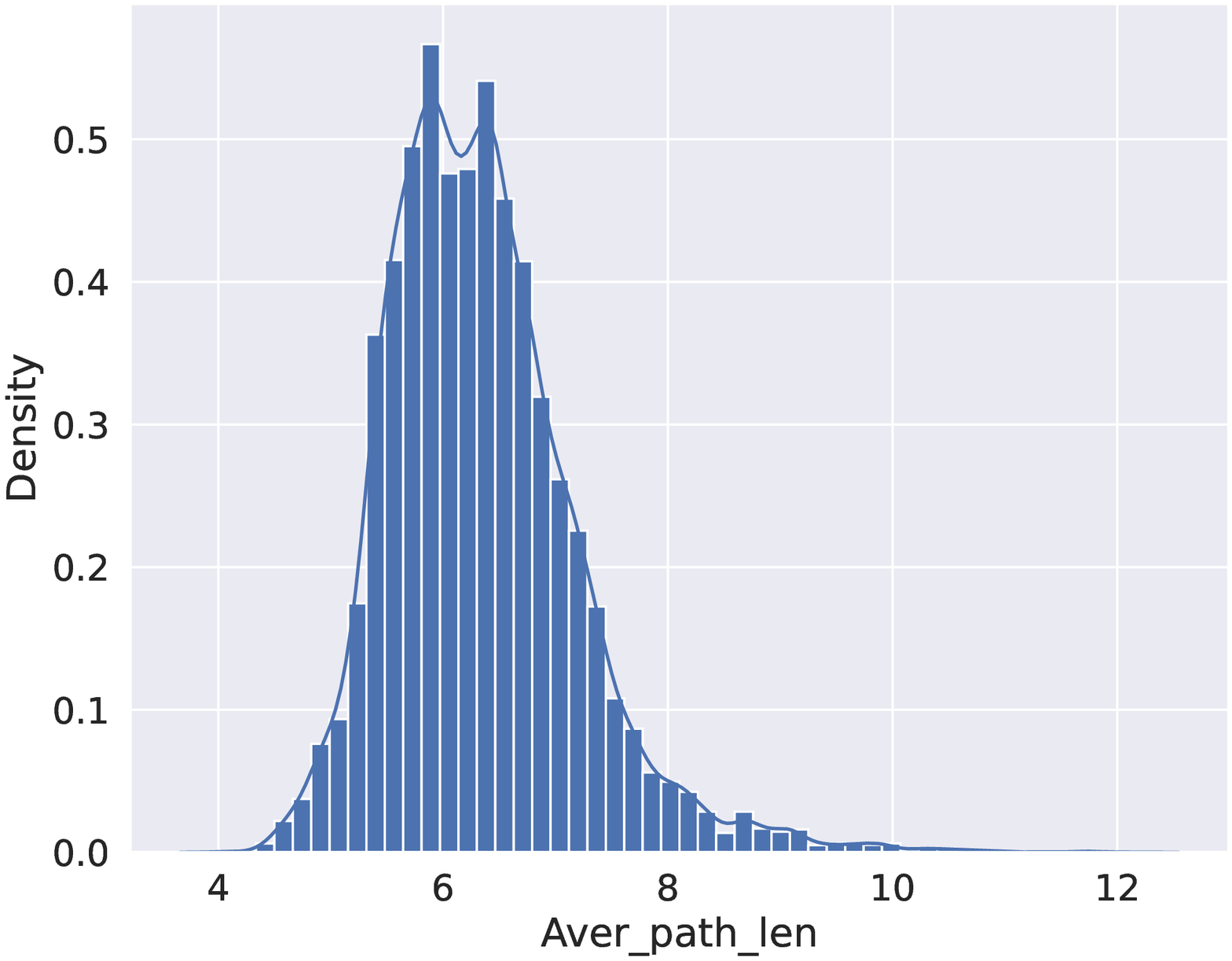}
            \hspace*{-1.5in}
            \caption{Distribution of graph feature of PubMed Dataset}
        \end{center}

      \centering
      \begin{center}

       \hspace*{-1.5in}
          \includegraphics[width=0.27\linewidth]{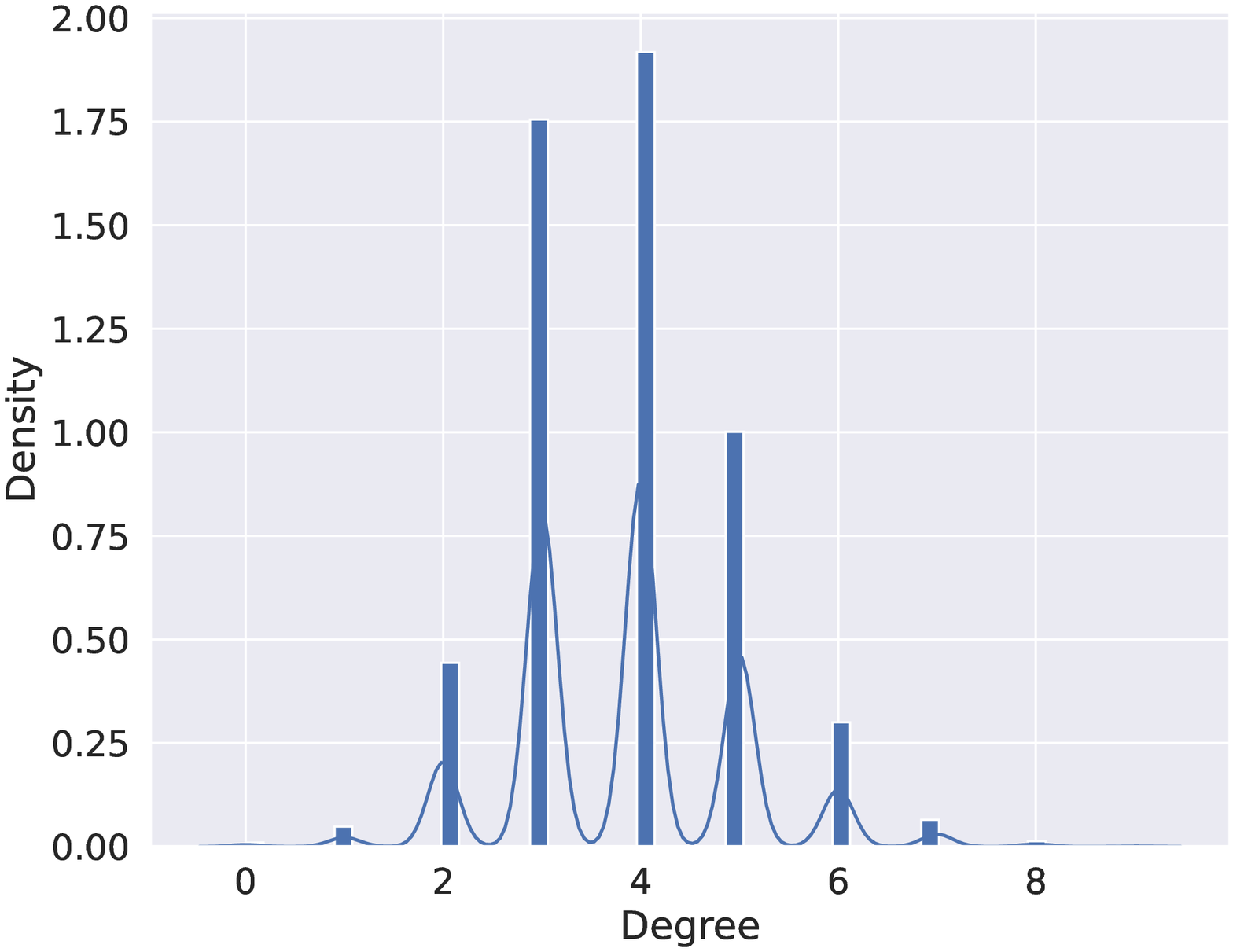}
          \includegraphics[width=0.27\linewidth]{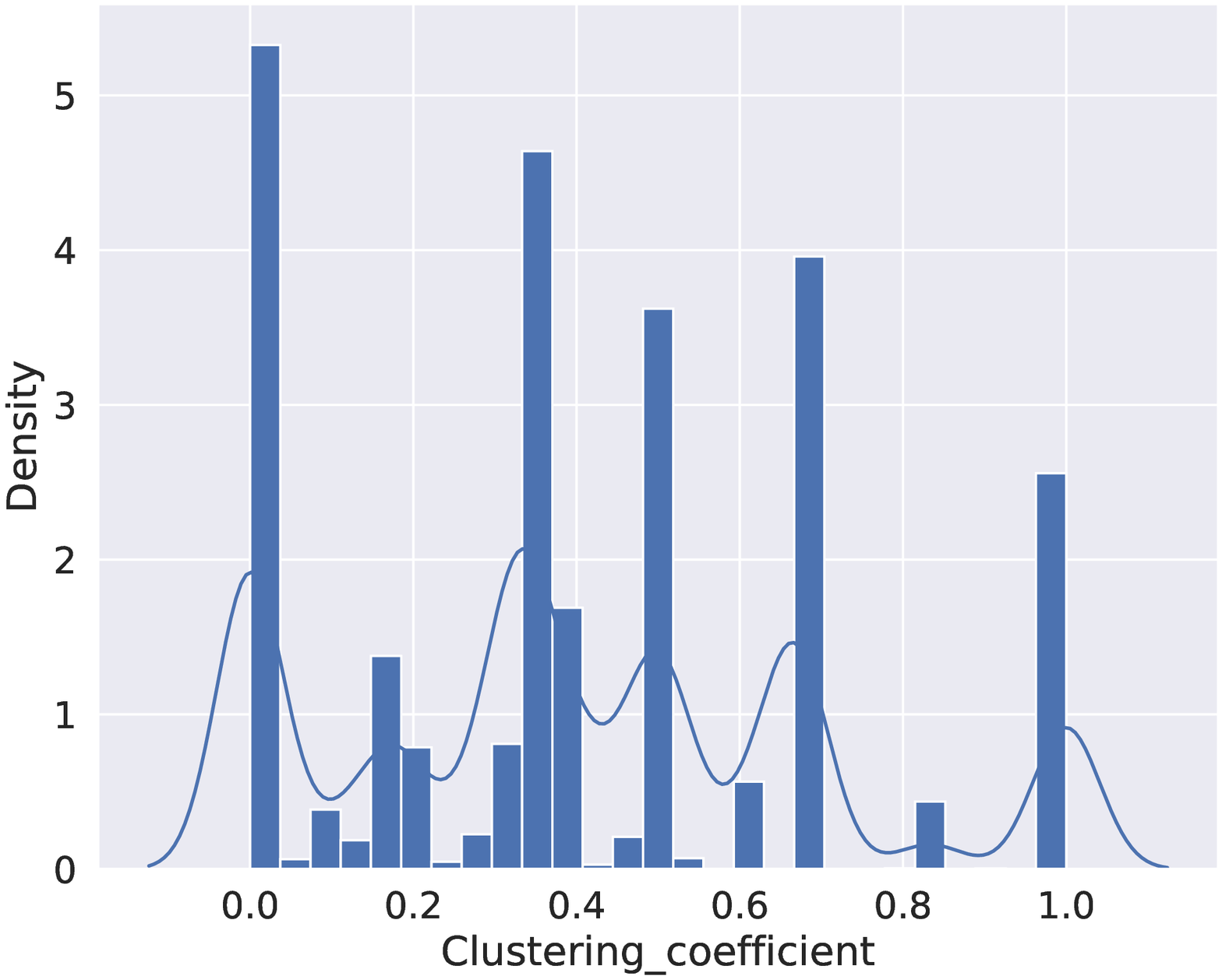}
          \includegraphics[width=0.27\linewidth]{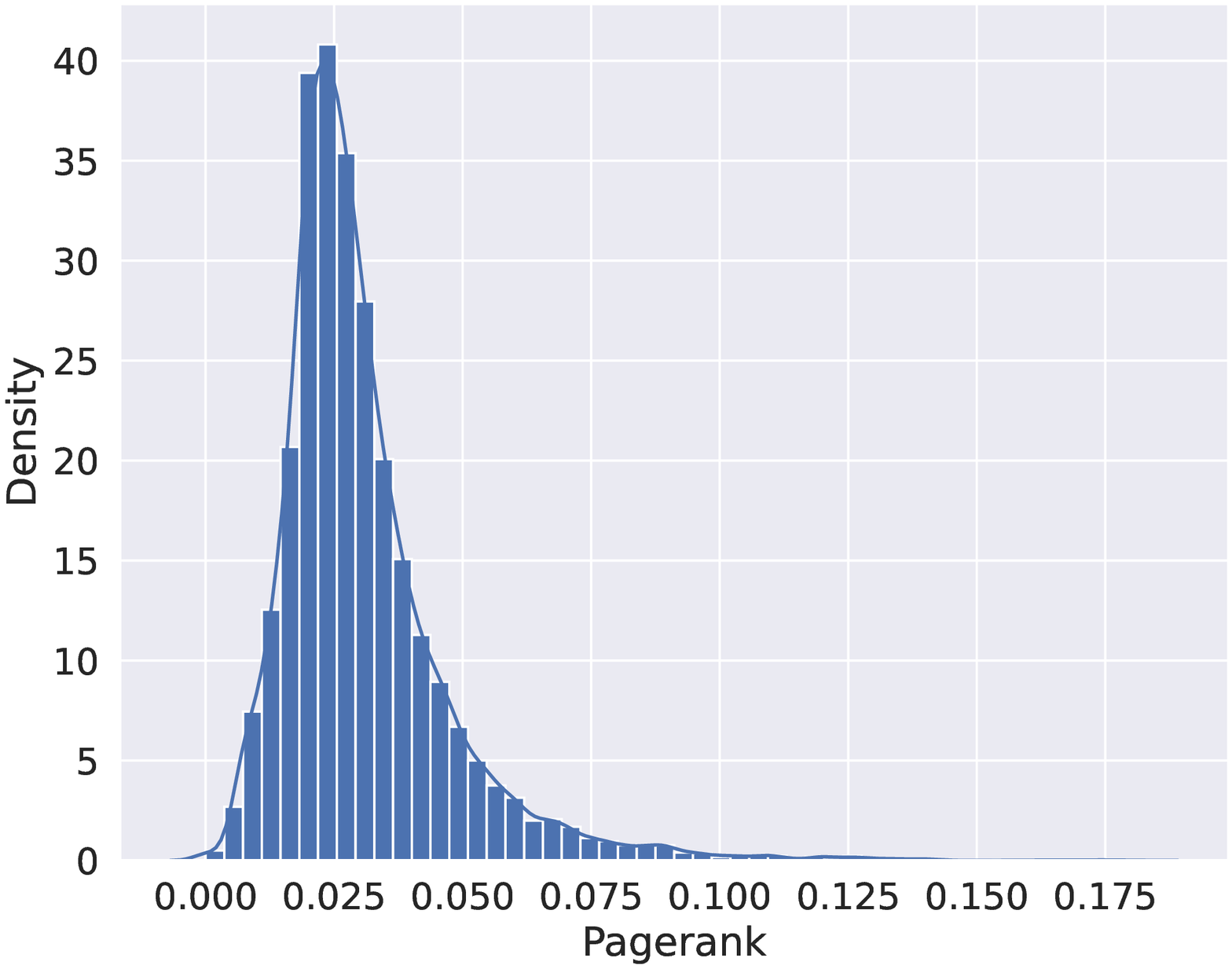}
          \includegraphics[width=0.27\linewidth]{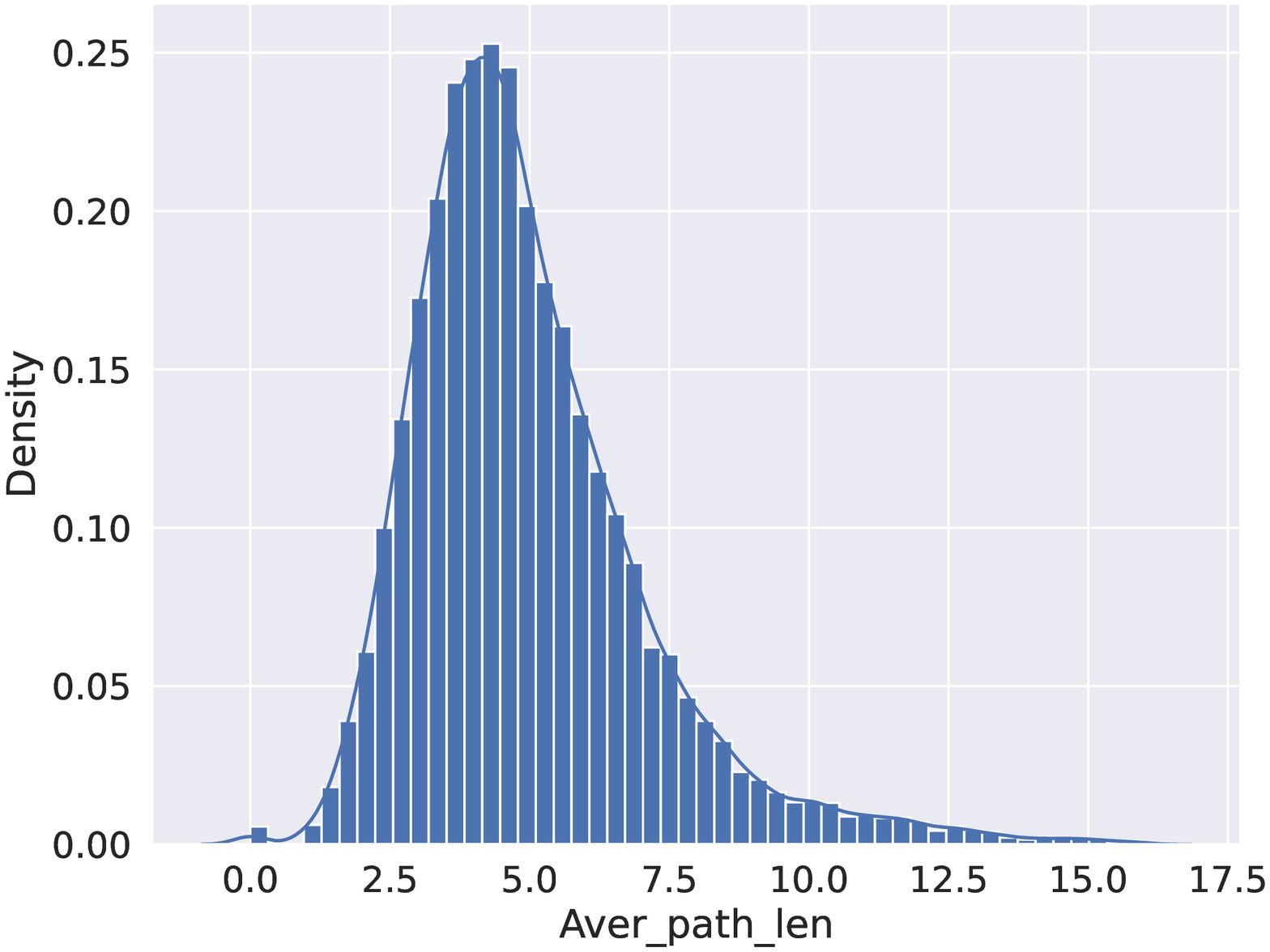}
          \hspace*{-1.5in}
          \caption{Distribution of graph feature of ENZYMES Dataset}
      \end{center}

      \centering
  \begin{center}
 \hspace*{-1.5in}
    \includegraphics[width=0.27\linewidth]{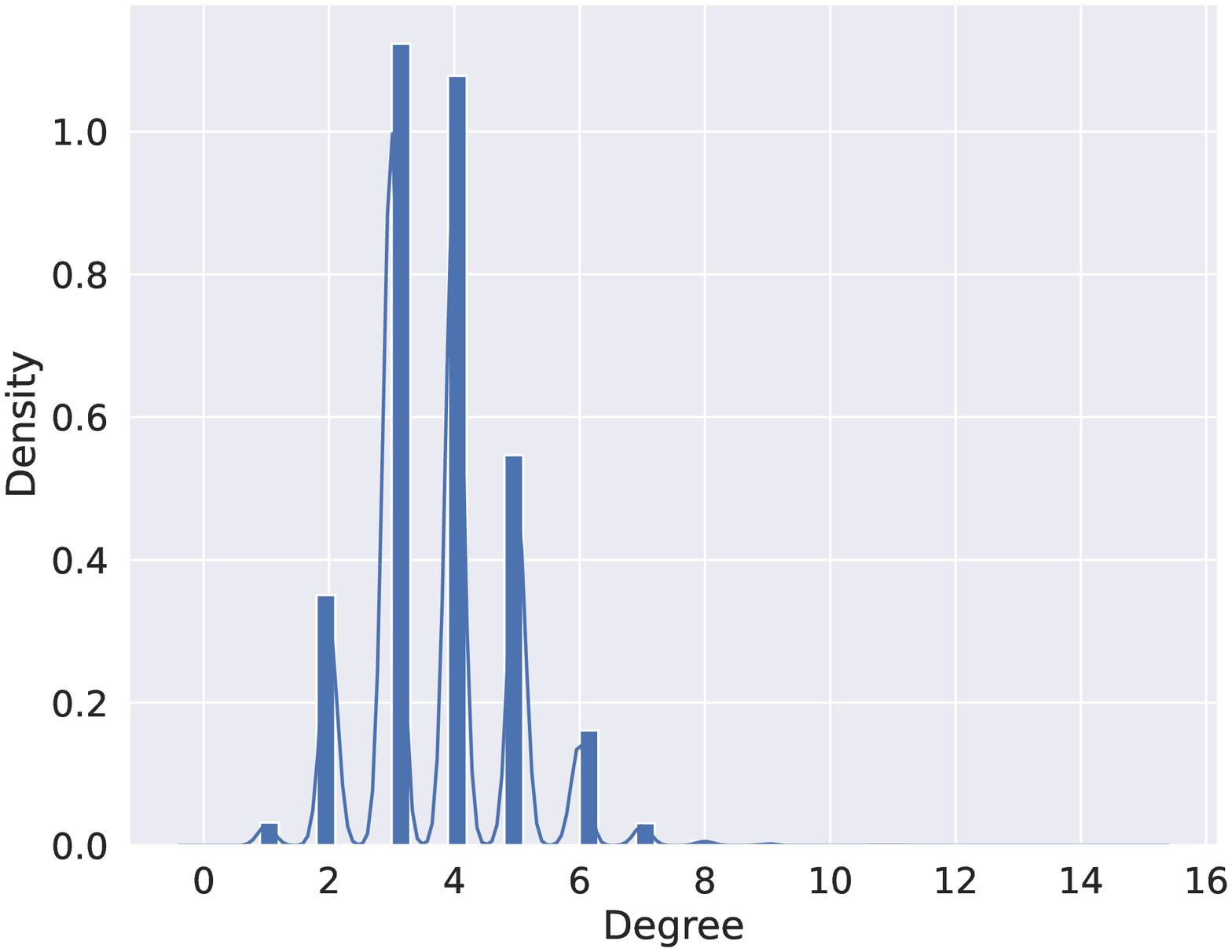}
    \includegraphics[width=0.27\linewidth]{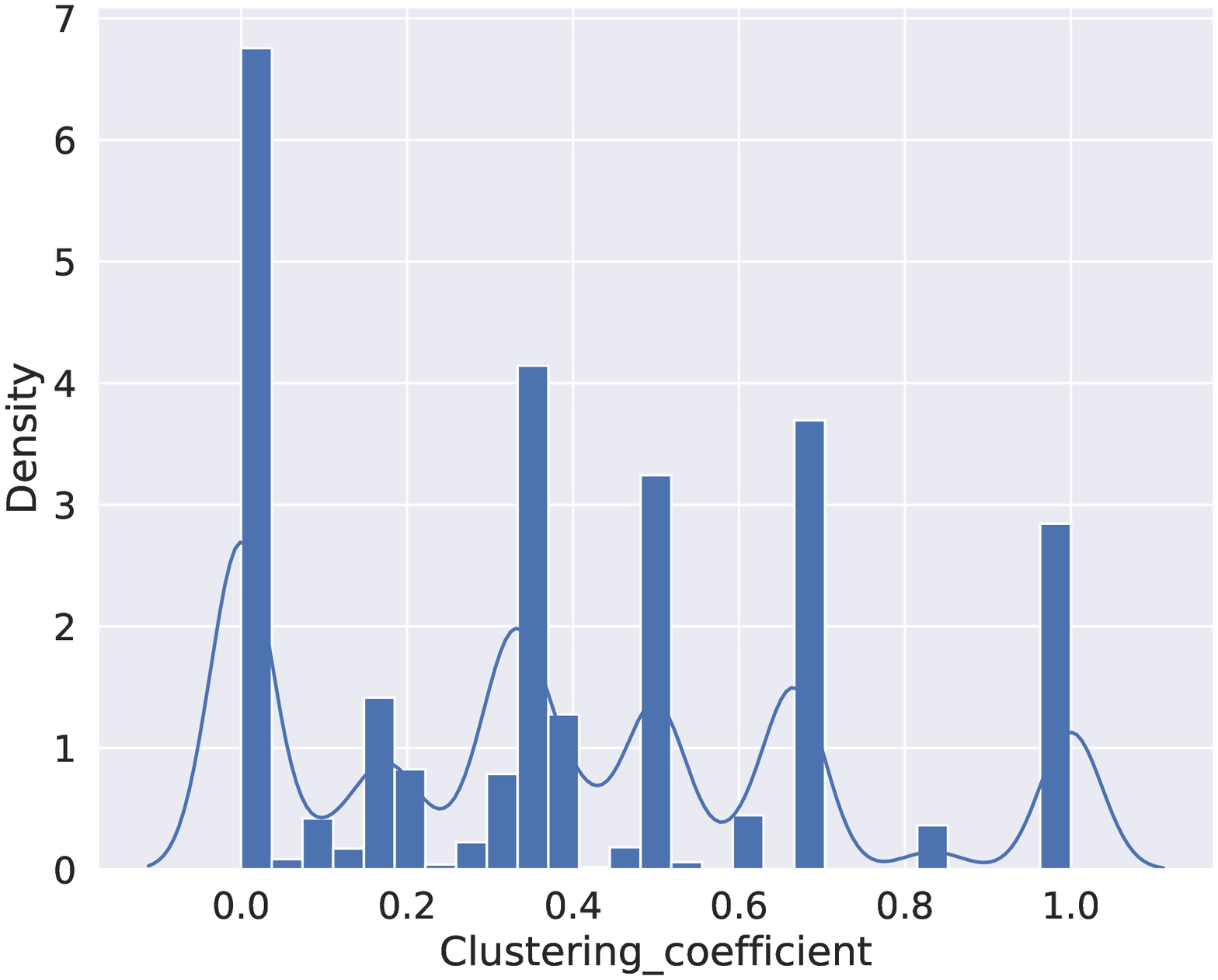}
    \includegraphics[width=0.27\linewidth]{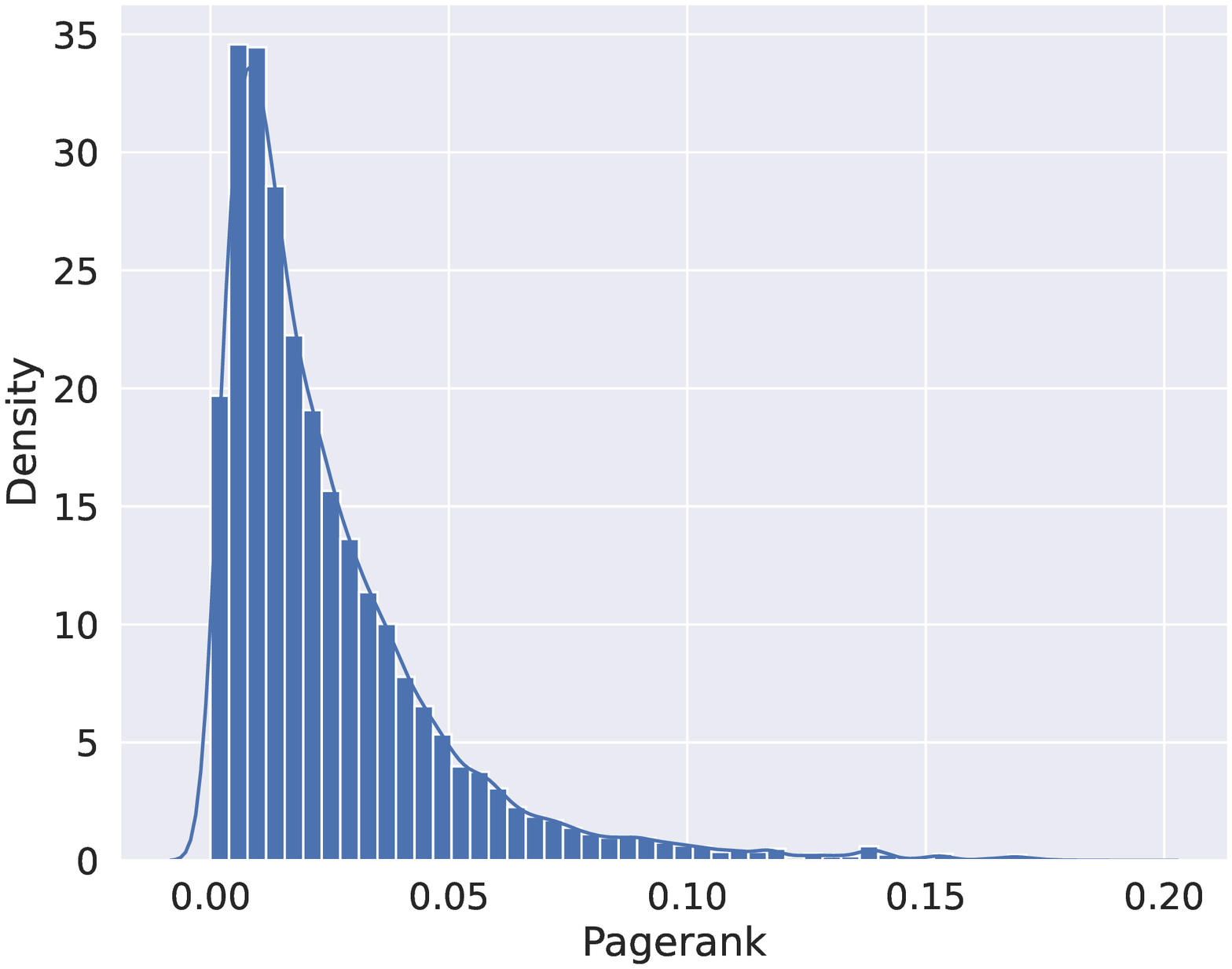}
    \includegraphics[width=0.27\linewidth]{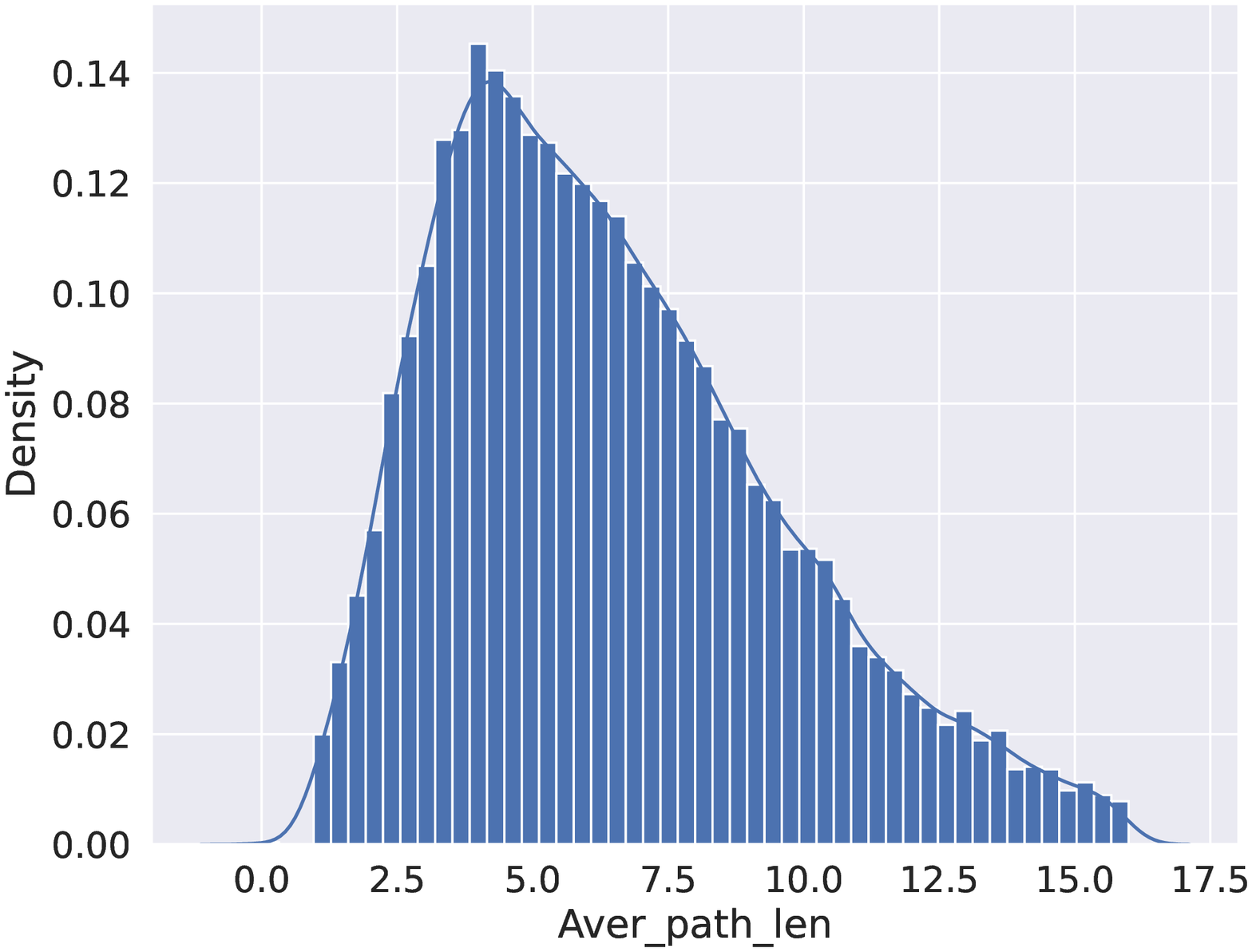}
    \hspace*{-1.5in}
    \caption{Distribution of graph feature of PROTEINS Dataset}
  \end{center}

  \centering
  \begin{center}
 \hspace*{-1.5in}
    \includegraphics[width=0.27\linewidth]{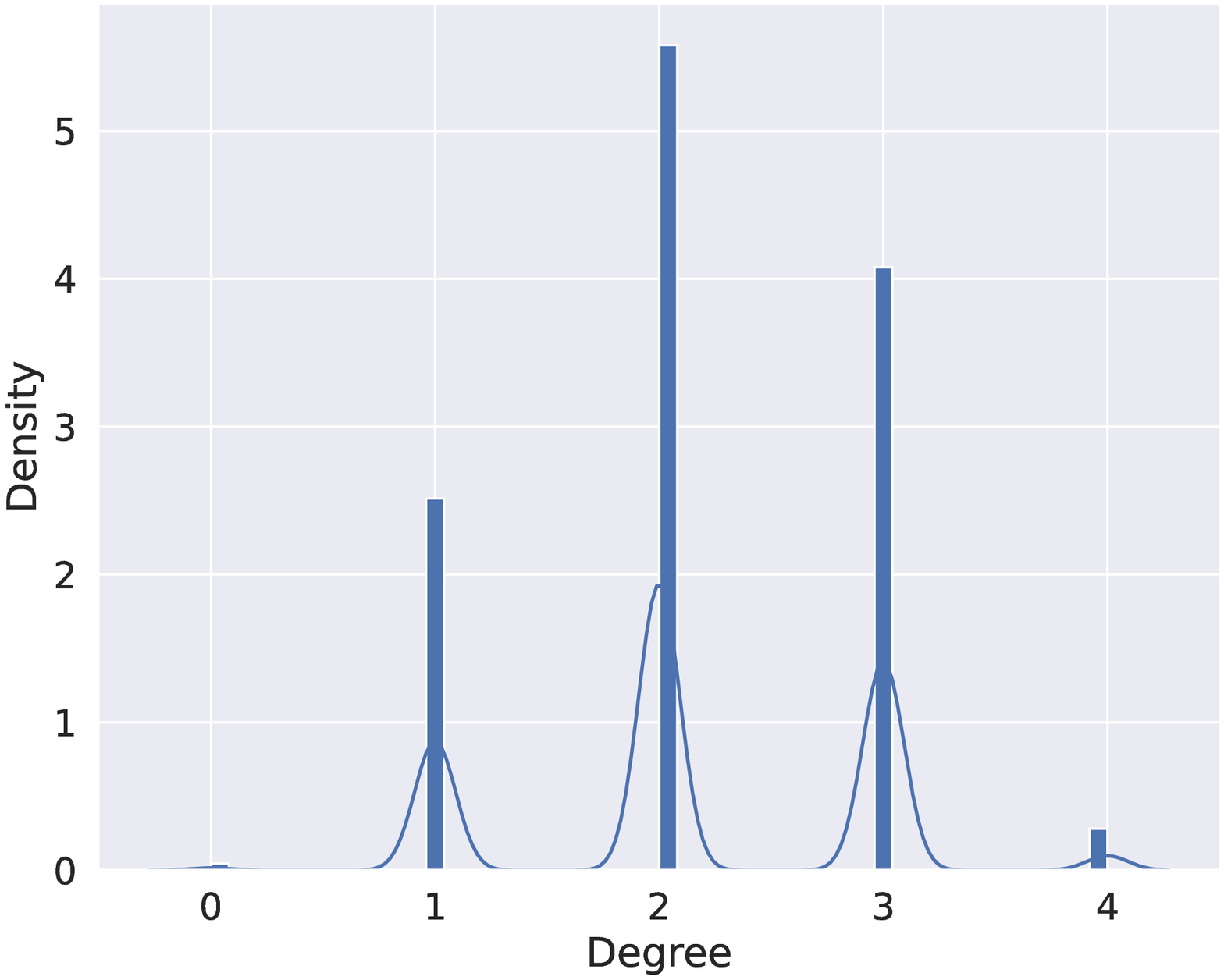}
    \includegraphics[width=0.27\linewidth]{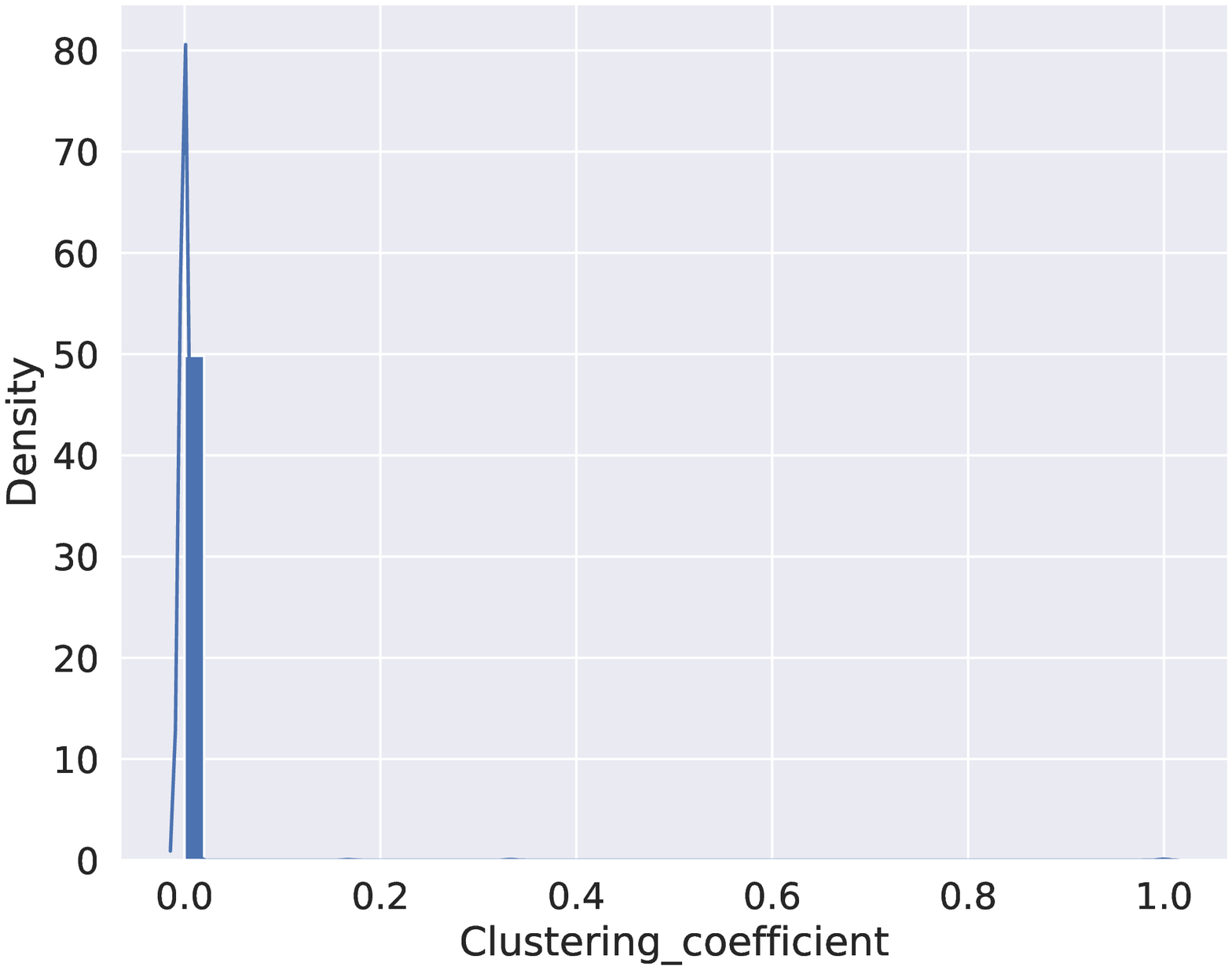}
    \includegraphics[width=0.27\linewidth]{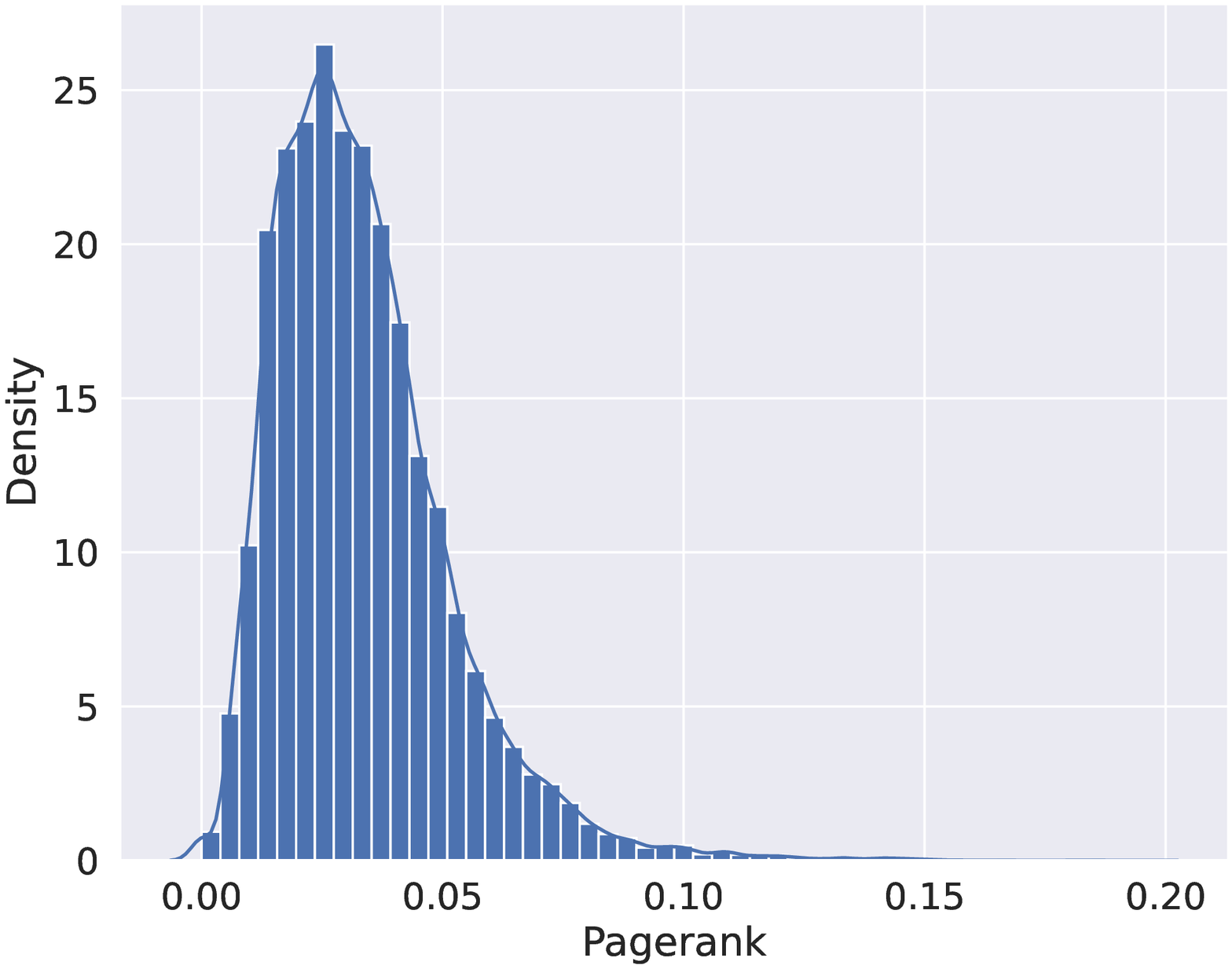}
    \includegraphics[width=0.27\linewidth]{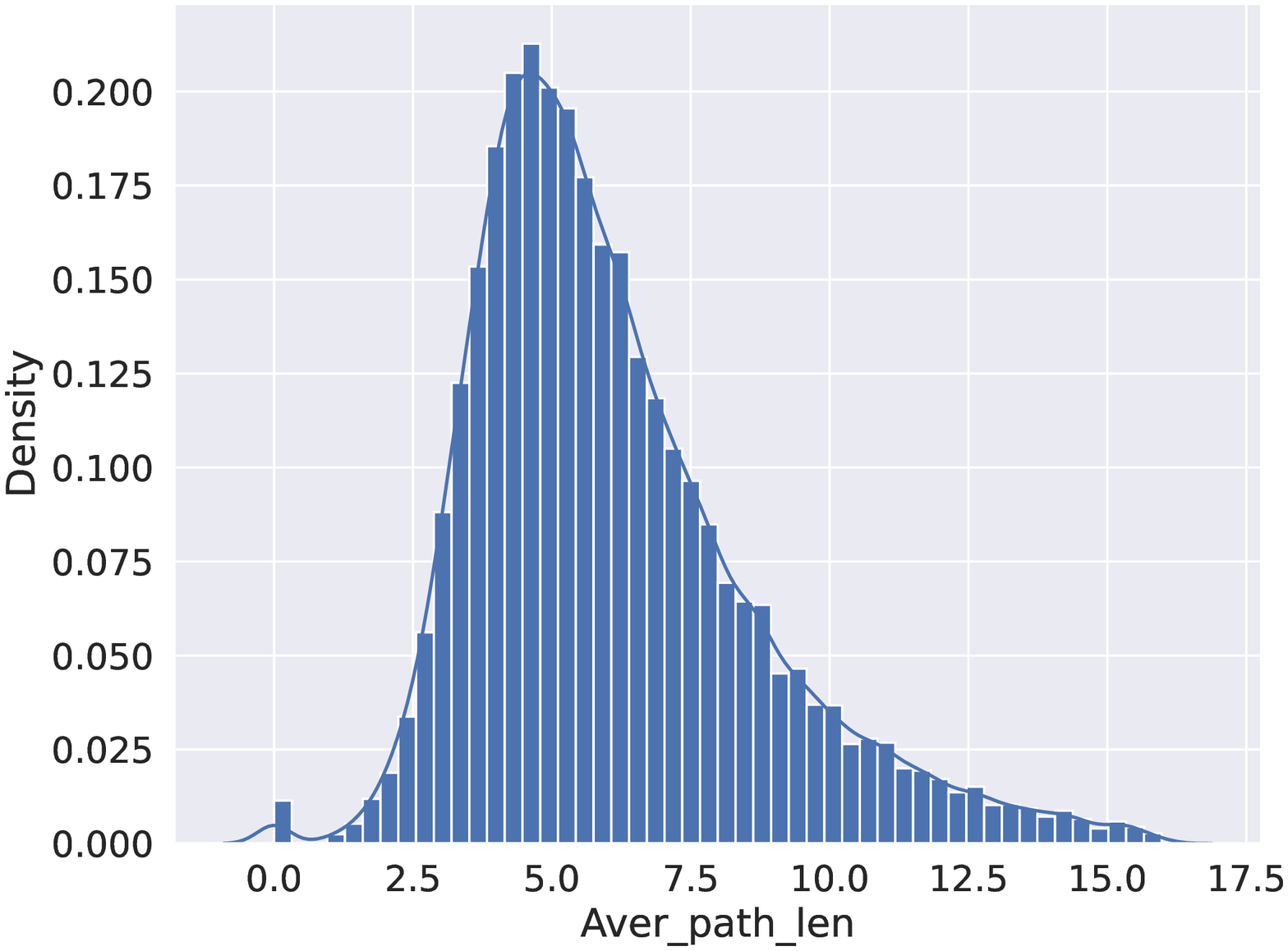}
    \hspace*{-1.5in}
    \caption{Distribution of graph feature of NCI1 Dataset}
  \end{center}
      \end{figure}

\vspace{-0.5cm}
  \section{Supplementary Results on Feature Prediction }

\vspace{-0.5cm}

  \xhdr{concatenation comparison}
\textit{NTN} method is worse than simple concatenation in the process of training 
the predicted \textit{pagerank} of Cora dataset, and it generalizes worse on test 
datasets. The \textit{bilinear} method faces the problem of over-fitting . In 
general, the \textit{simple concatenation} has the best performance on test 
datasets. When predicting \textit{average path length} on \textit{Cora} dataset, 
\textit{NTN} is far better than \textit{bilinear} and \textit{simple 
concatenation} but they show  similar results on test datasets. More specifically, 
\textit{Bilinear} and \textit{NTN} have a serious fluctuation problem on accuracy, 
overall \textit{NTN} is the optimal method according to Fea2Fea-multiple. These experiments indicate that 
complex concatenation method such as \textit{NTN} does not always have the best 
performance. Sometimes the simplest way is the best way.

  \begin{figure}[!h]
    \centering
    \begin{center}
    \hspace*{-1in}
        \includegraphics[width=0.25\linewidth]{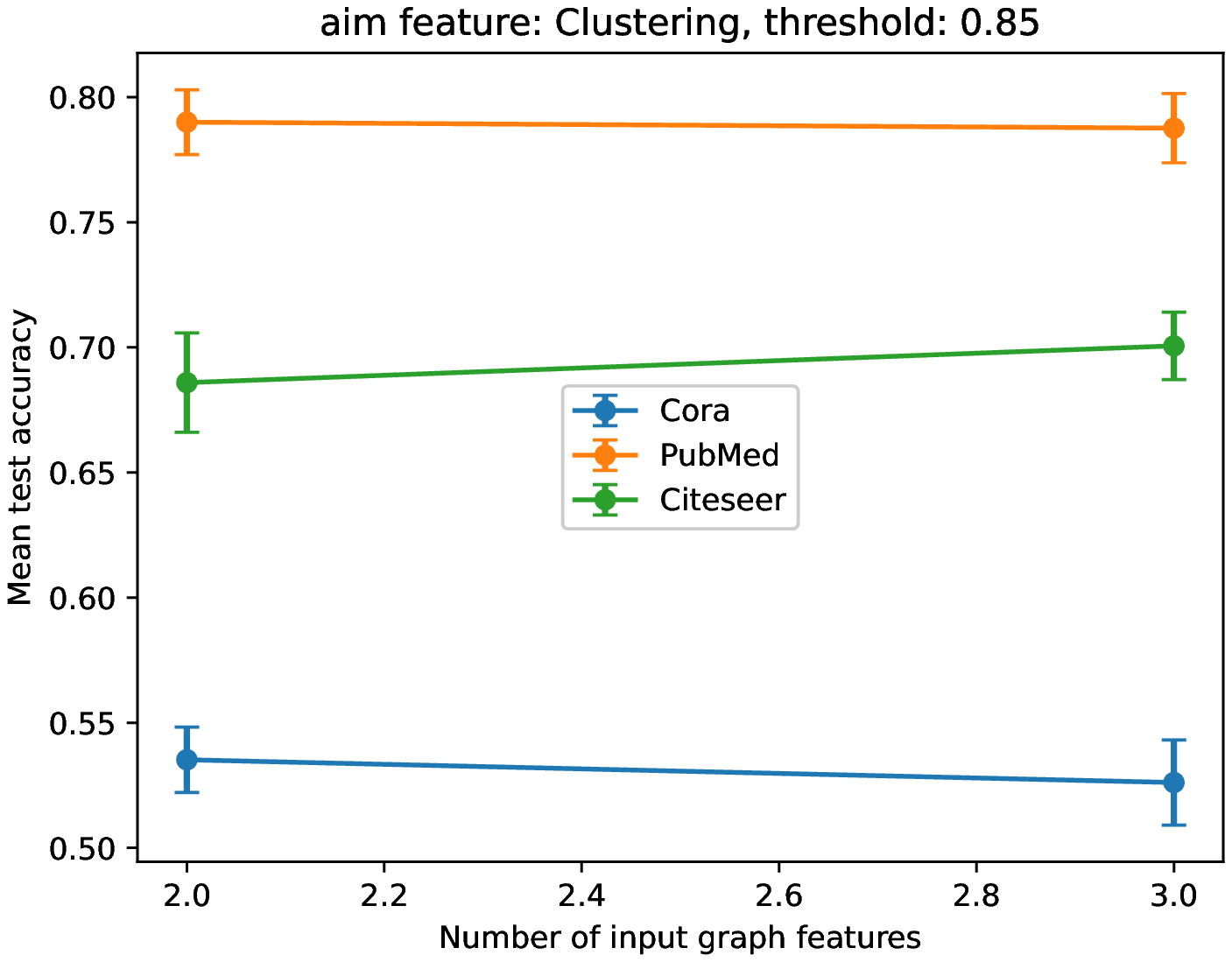}
        \includegraphics[width=0.25\linewidth]{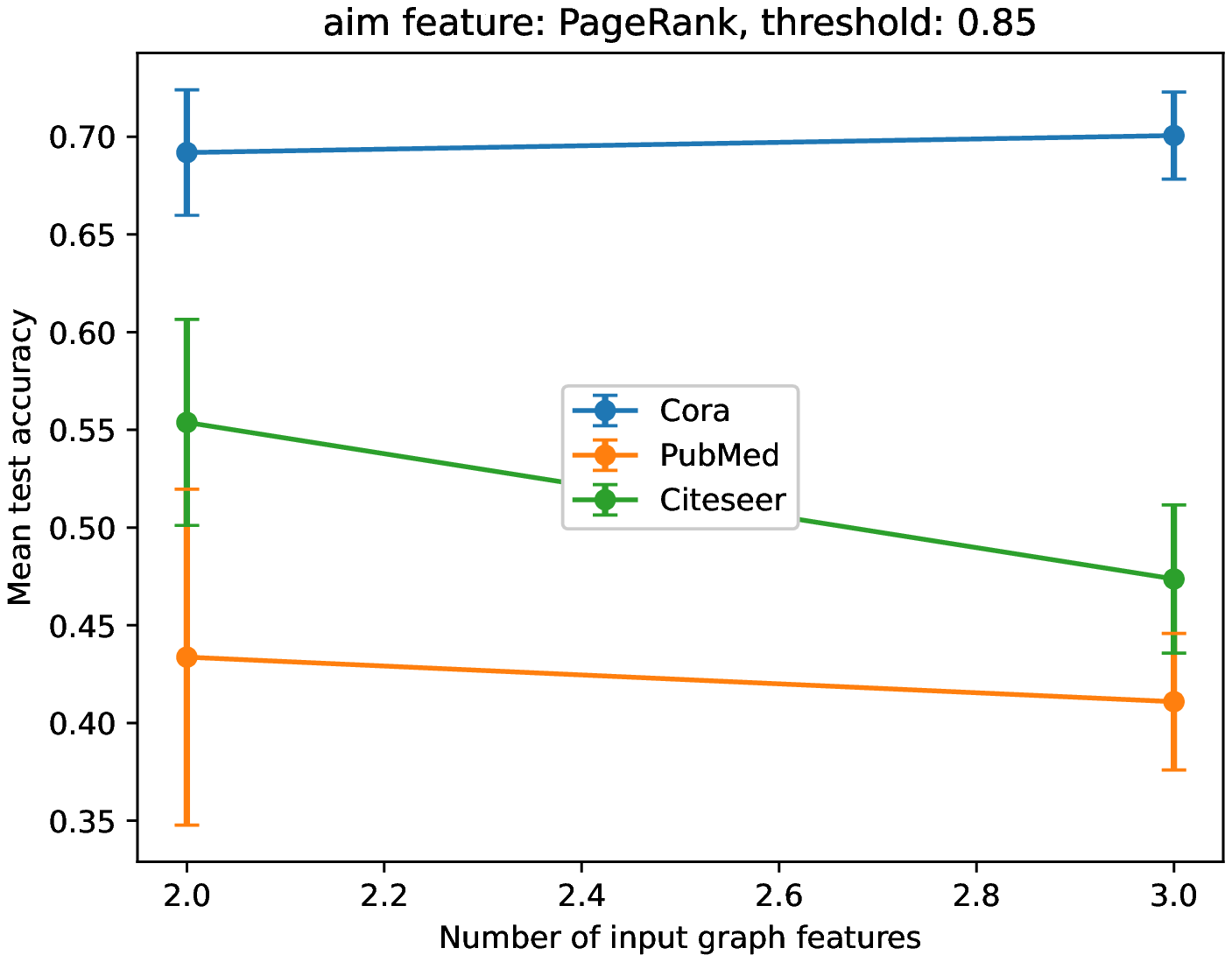}
        \includegraphics[width=0.25\linewidth]{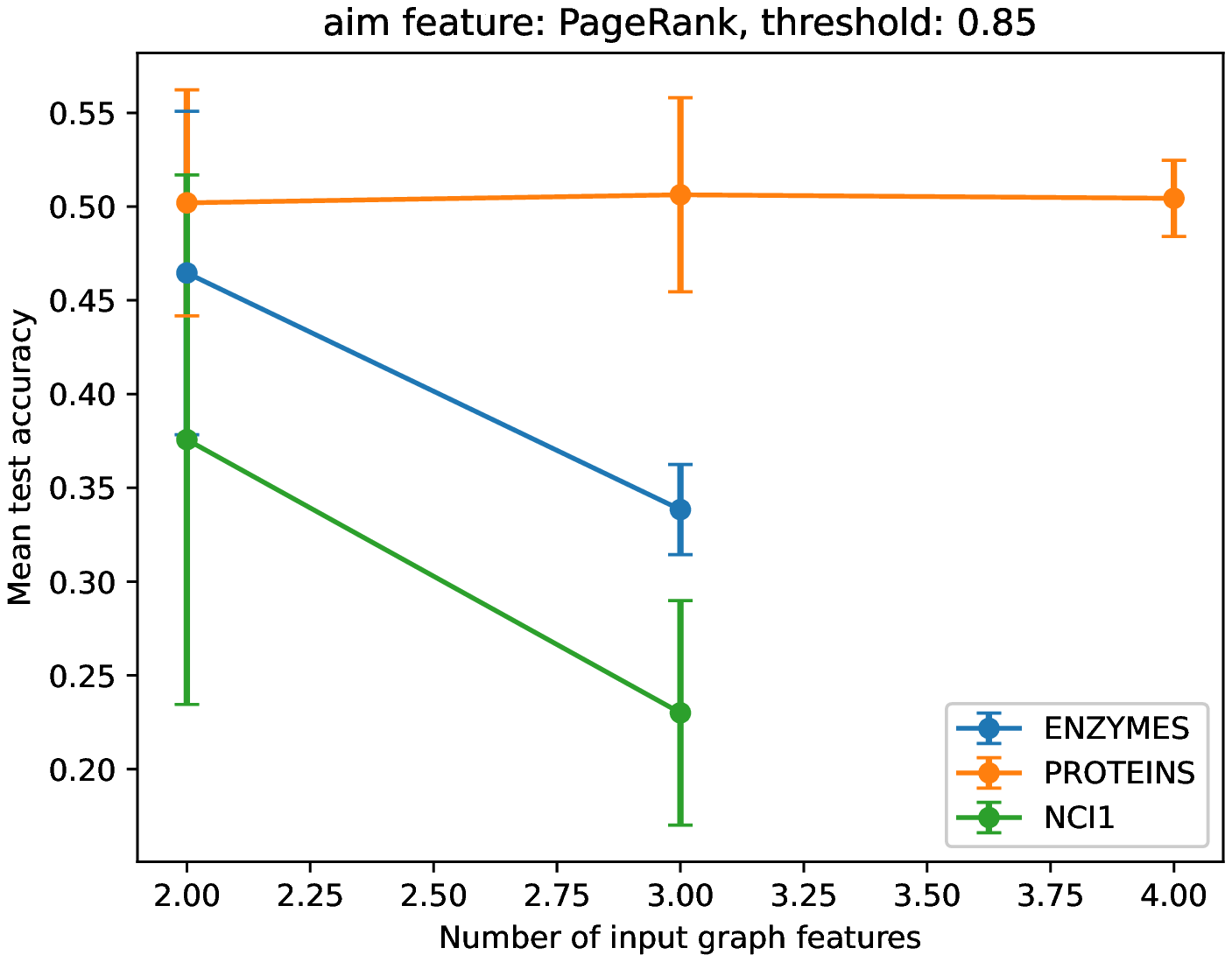}
        \includegraphics[width=0.25\linewidth]{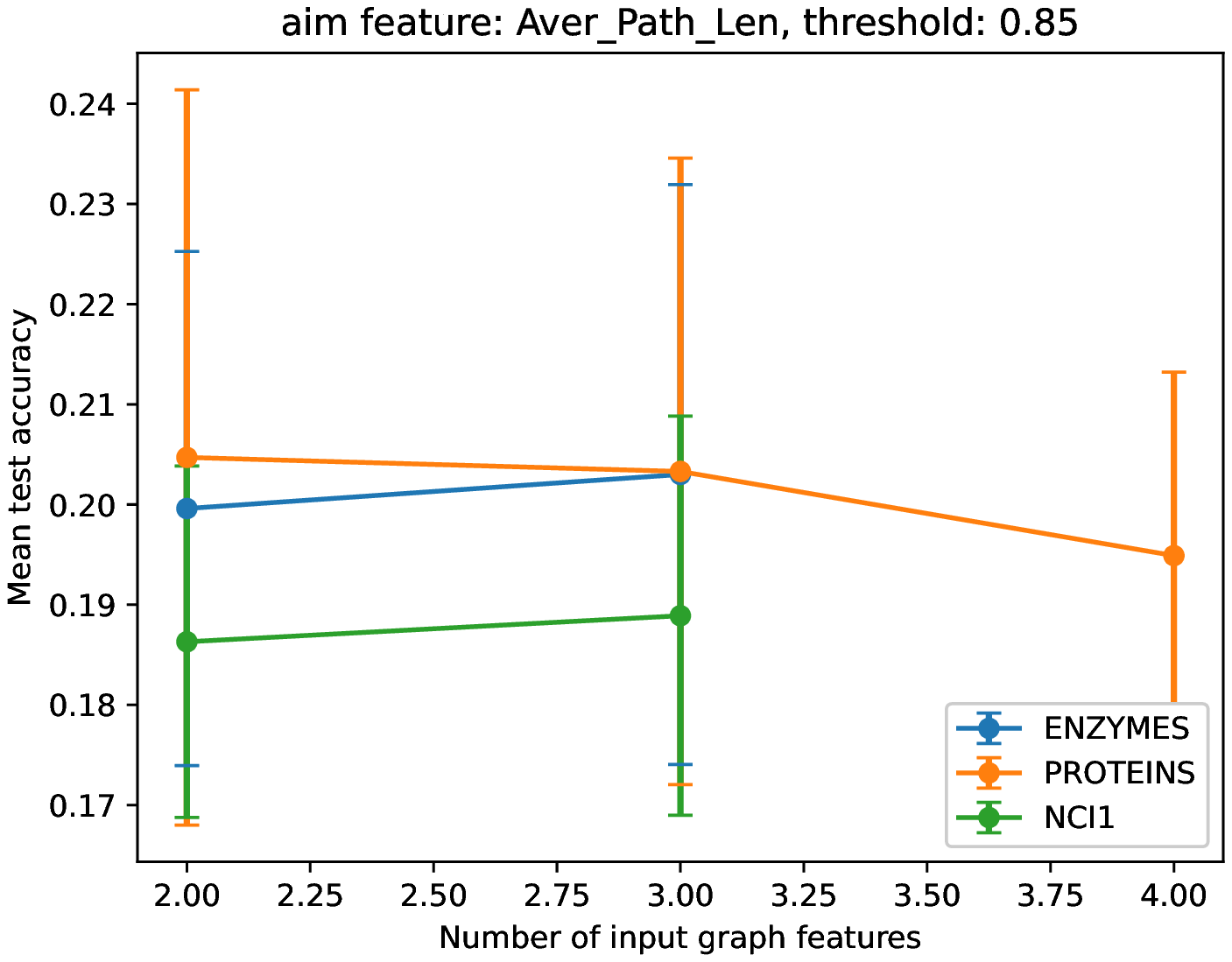}
        \hspace*{-1in}
        \caption{Different numbers of features contribute to different prediction 
  results for node and graph datasets. We perform 10 experiments and take the
  average accuracy with standard deviation shown in the figures.}

    \end{center}
    \end{figure}

    \begin{figure}[!htp]
  \centering
  \begin{center}
  \hspace*{-1in}
      \includegraphics[width=0.25\linewidth]{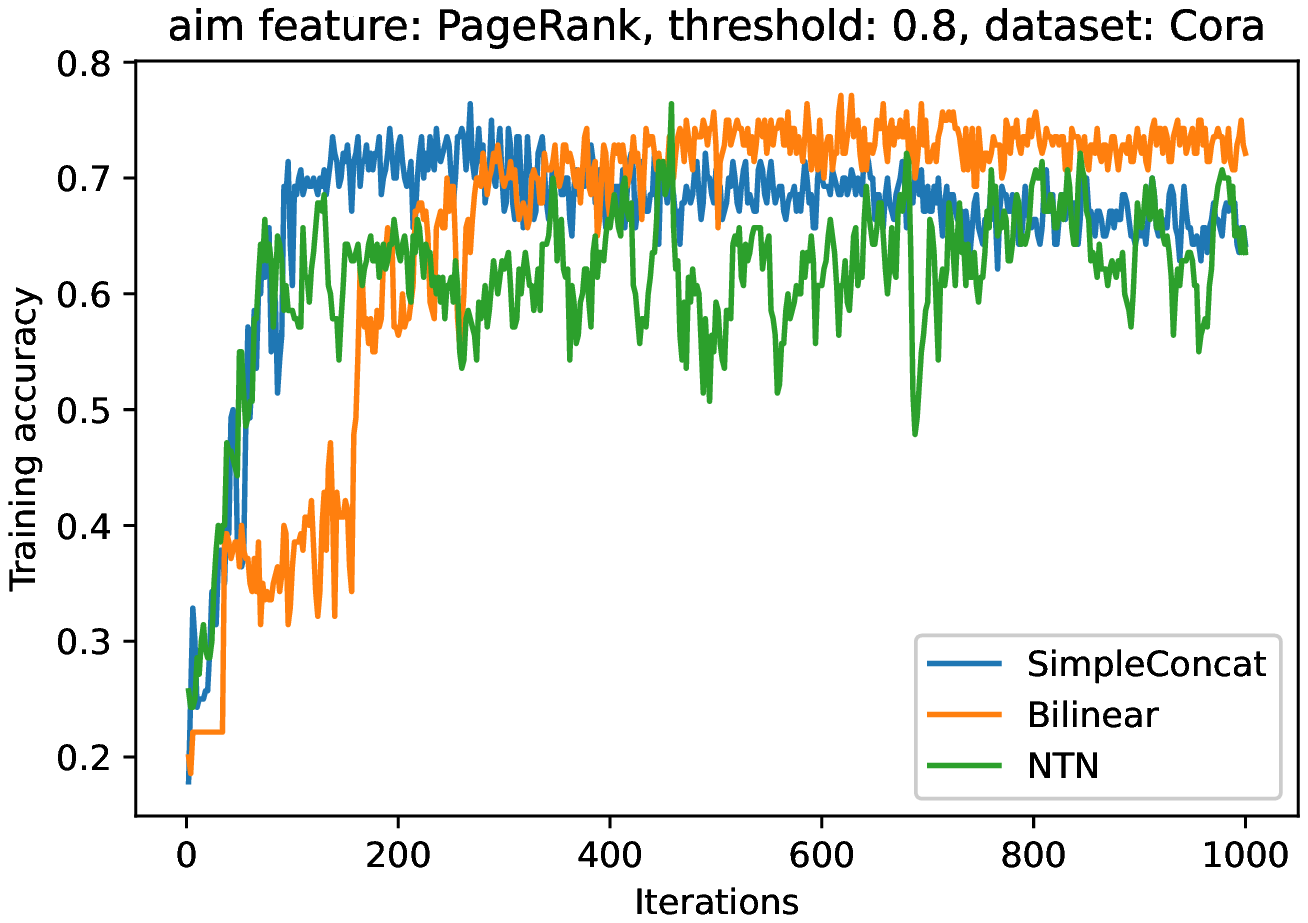}
      \includegraphics[width=0.25\linewidth]{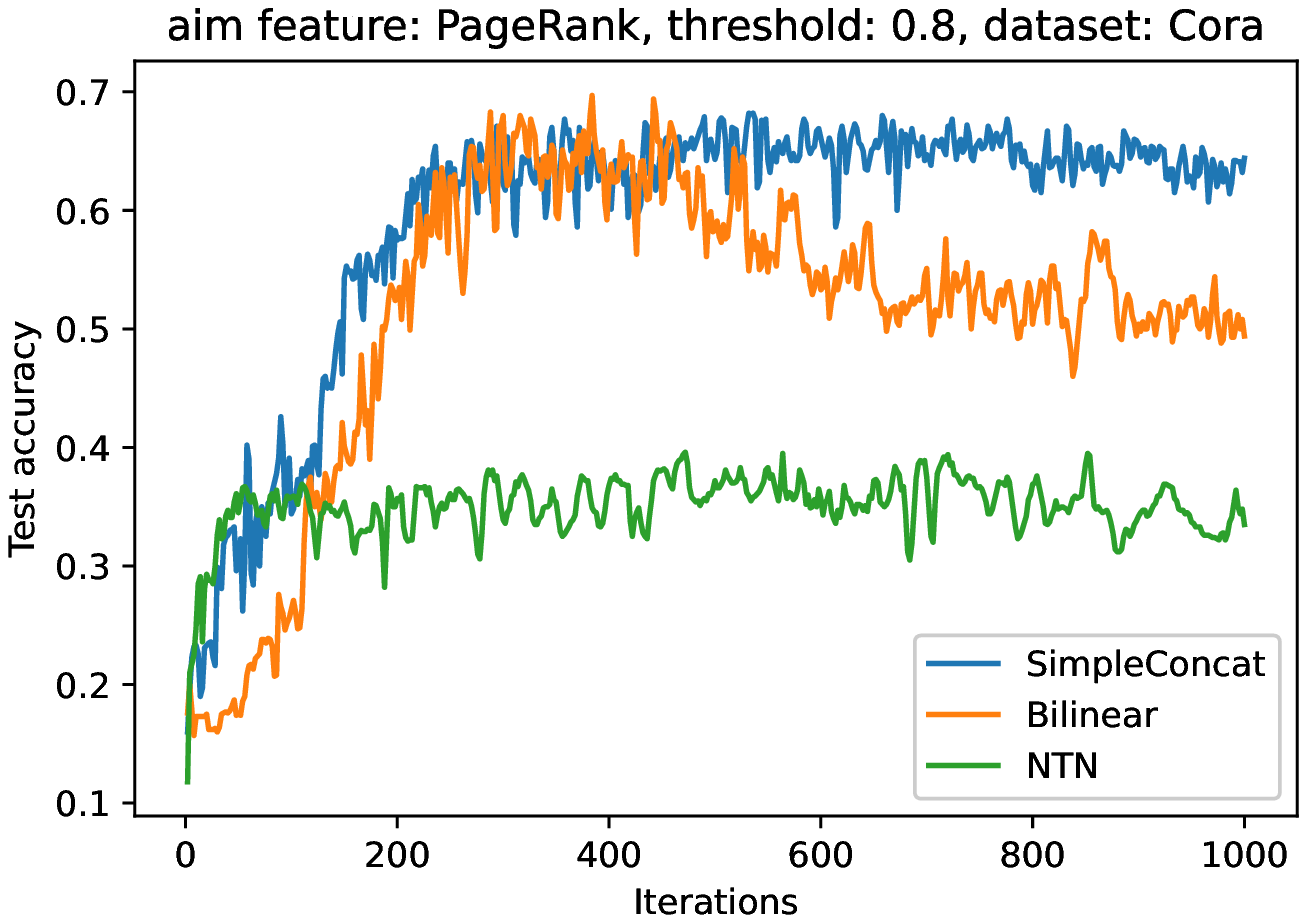}
      \includegraphics[width=0.25\linewidth]{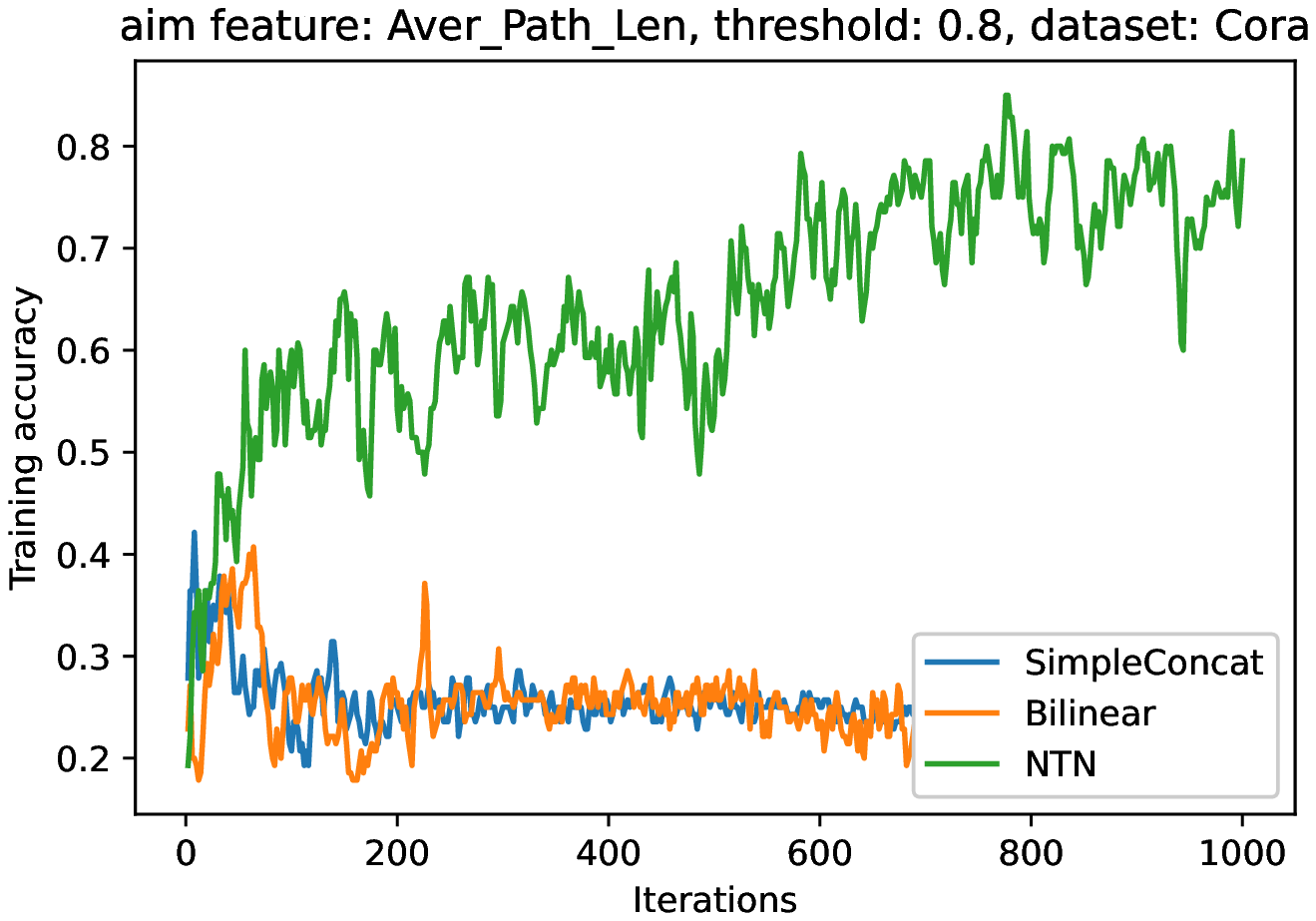}
      \includegraphics[width=0.25\linewidth]{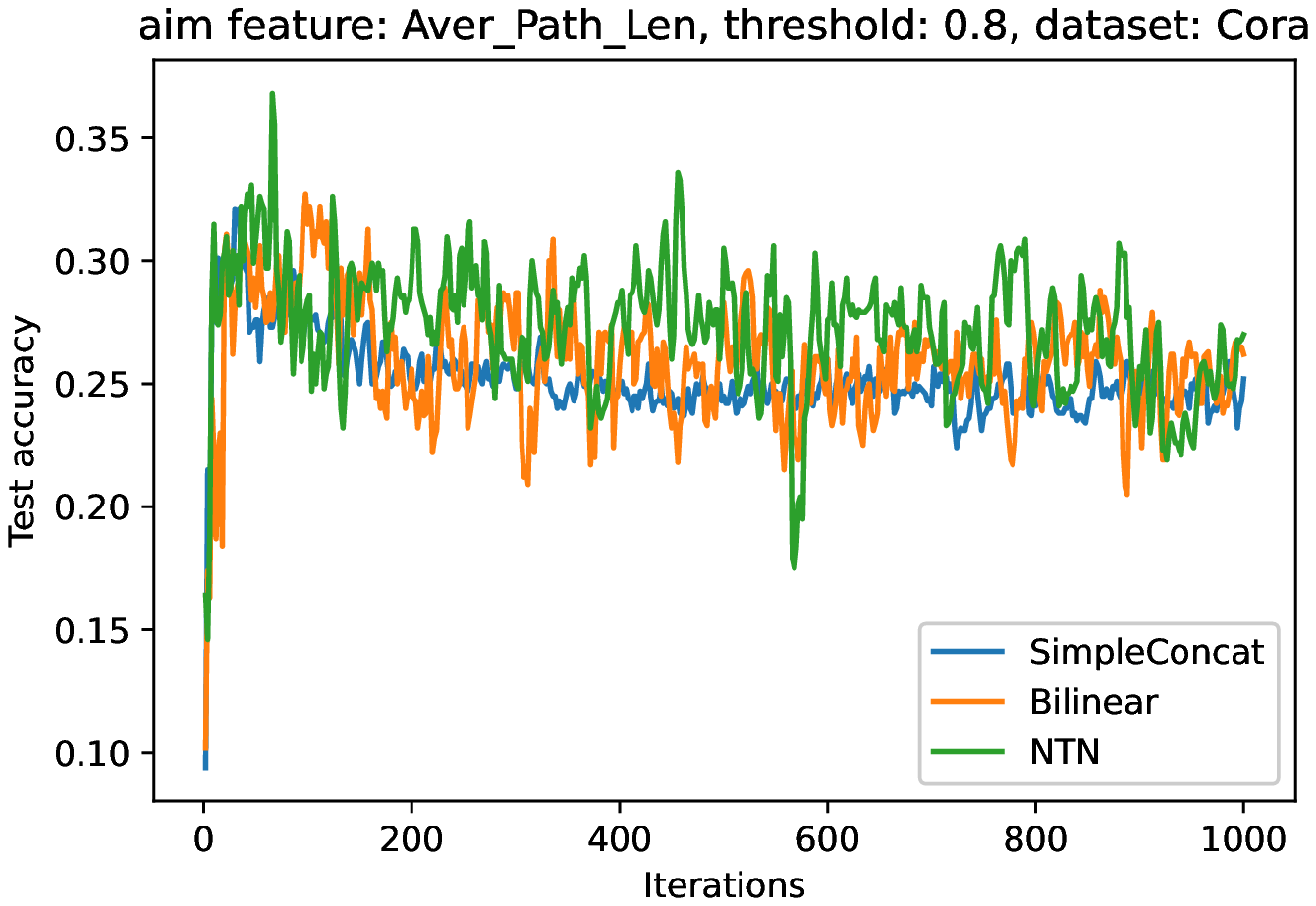}
      \hspace*{-1in}
      \caption{Training and test accuracy versus epochs for different concatenation results when 
predicting clustering coefficient and pagerank in Fea2Fea-multiple}

  \end{center}
  \end{figure}
 
\begin{table*}[!htb]
  \centering
  \caption{Feature to Feature Prediction on Planetoid Datasets (bins = 6)}
  \begin{tabular}{*{11}{c}} \toprule
{Aim}  & \multicolumn{5}{c}{{\sc Citeseer}} & \multicolumn{5}{c}{{\sc PubMed}} 
\\
\cmidrule(lr){2-6}\cmidrule(lr){7-11}
 & GCN & GIN & SAGE & GAT & MLP  &GCN & GIN & SAGE & GAT & MLP \\ \hline
{\textit{Cons} $\rightarrow$ \textit{Deg}}     & 0.587 & \B1.000 & 0.379 & 0.379 & 0.379 & 0.678 & \B0.996 & 0.478 & 0.478 & 0.478 \\
{\textit{Deg} $\rightarrow$ \textit{Deg}}      & 0.899 & \B1.000 & 0.994 & 0.395 & 1.000 & 0.725 & \B1.000 & 0.958 & 0.477 & 1.000 \\
{\textit{Clu} $\rightarrow$ \textit{Deg}}      & 0.628 & \B1.000 & 0.631 & 0.492 & 0.670 & 0.643 & \B1.000 & 0.603 & 0.471 & 0.574 \\
{\textit{PR} $\rightarrow$\textit{Deg}}        & 0.466 & \B1.000 & 0.409 & 0.376 & 0.379 & 0.539 & \B1.000 & 0.600 & 0.478 & 0.478 \\
{\textit{Avglen} $\rightarrow$ \textit{Deg}}   & 0.536 & \B0.997 & 0.686 & 0.476 & 0.473 & 0.696 & \B1.000 & 0.785 & 0.478 & 0.524 \\
{\textit{Cons} $\rightarrow$ \textit{Clu}}     & 0.671 &  \B0.693 & 0.658 & 0.658 & 0.658 & 0.799 &  \B0.805 & 0.780 & 0.780 & 0.780 \\
{\textit{Deg} $\rightarrow$ \textit{Clu}}      & 0.688 & 0.714 &  \B0.736 & 0.658 & 0.726 & 0.794 &  \B0.804 &  \B0.804 & 0.780 & 0.805 \\
{\textit{Clu}  $\rightarrow$ \textit{Clu}}     & 0.857 & 0.882 & 0.980 & 0.846 &  \B0.992 & 0.831 & 0.839 &  \B0.939 & 0.762 & 0.932 \\
{\textit{PR} $\rightarrow$ \textit{Clu}}       & 0.672 &  \B0.687 & 0.670 & 0.658 & 0.658 & 0.792 & \B 0.805 & 0.780 & 0.780 & 0.780 \\
{\textit{Avglen} $\rightarrow$ \textit{Clu}}   & 0.681 &  \B0.696 & 0.684 & 0.658 & 0.676 &  \B0.794 & 0.785 & 0.790 & 0.780 & 0.780 \\
{\textit{Cons} $\rightarrow$ \textit{PR}}      &  \B0.702 & 0.671 & 0.202 & 0.190 & 0.190 &  \B0.669 & 0.529 & 0.161 & 0.141 & 0.161 \\
{\textit{Deg} $\rightarrow$ \textit{PR}}       & 0.637 & 0.750 &  \B0.752 & 0.266 & 0.443 & 0.564 &  \B0.629 & 0.617 & 0.175 & 0.565 \\
{\textit{Clu} $\rightarrow$ \textit{PR}}       & 0.549 &  \B0.575 & 0.435 & 0.279 & 0.315 & 0.437 &  \B0.559 & 0.409 & 0.196 & 0.326 \\
{\textit{PR} $\rightarrow$ \textit{PR}}        & 0.192 &  \B0.635 & 0.415 & 0.263 & 0.190 & 0.478 &  \B0.554 & 0.336 & 0.161 & 0.161 \\
{\textit{Avglen} $\rightarrow$ \textit{PR}}    & 0.529 &  \B0.691 & 0.602 & 0.312 & 0.333 & 0.541 &  \B0.591 & 0.537 & 0.274 & 0.263 \\
{\textit{Cons} $\rightarrow$ \textit{Avglen}}  & 0.442 &  \B0.503 & 0.178 & 0.166 & 0.173 & 0.294 &  \B0.394 & 0.168 & 0.168 & 0.168 \\
{\textit{Deg} $\rightarrow$ \textit{Avglen}}   & 0.528 & 0.542 &  \B0.553 & 0.285 & 0.330 & 0.415 &  \B0.443 & 0.437 & 0.153 & 0.313 \\
{\textit{Clu} $\rightarrow$ \textit{Avglen}}   &  \B0.420 & 0.377 & 0.387 & 0.248 & 0.246 & 0.296 &  \B0.330 & 0.300 & 0.171 & 0.184 \\
{\textit{PR} $\rightarrow$ \textit{Avglen}}    & 0.268 &  \B0.466 & 0.207 & 0.310 & 0.173 & 0.316 &  \B0.459 & 0.198 & 0.197 & 0.168 \\
{\textit{Avglen} $\rightarrow$ \textit{Avglen}}& 0.734 & 0.597 & 0.937 & 0.596 &  \B0.979 & 0.450 & 0.378 & 0.860 & 0.270 &  \B0.984 \\

\bottomrule
  \end{tabular}
\end{table*}

  \section{Graph Embedding Analysis}
  We extract graph embedding vectors and linear layer embedding vectors to perform tsne visualization and comparsion.
  When predicting \textit{pagerank} from node \textit{degree} on {\sc{Citeseer}} dataset, the graph embedding is similiar to the mlp embedding. On {\sc{Proteins}} dataset, graph embedding cannot separate classes well compared with mlp embedding. When performing multiple features to single featuere prediction, input embedding indicates that input features are mixed which cannot classify the class well but mlp embedding can separate the class well. It emphasizes the importance of a MLP after the GNN block.

  \begin{figure}
    \centering
    \begin{center}i
    \hspace*{-1in}
        \includegraphics[width=0.33\linewidth]{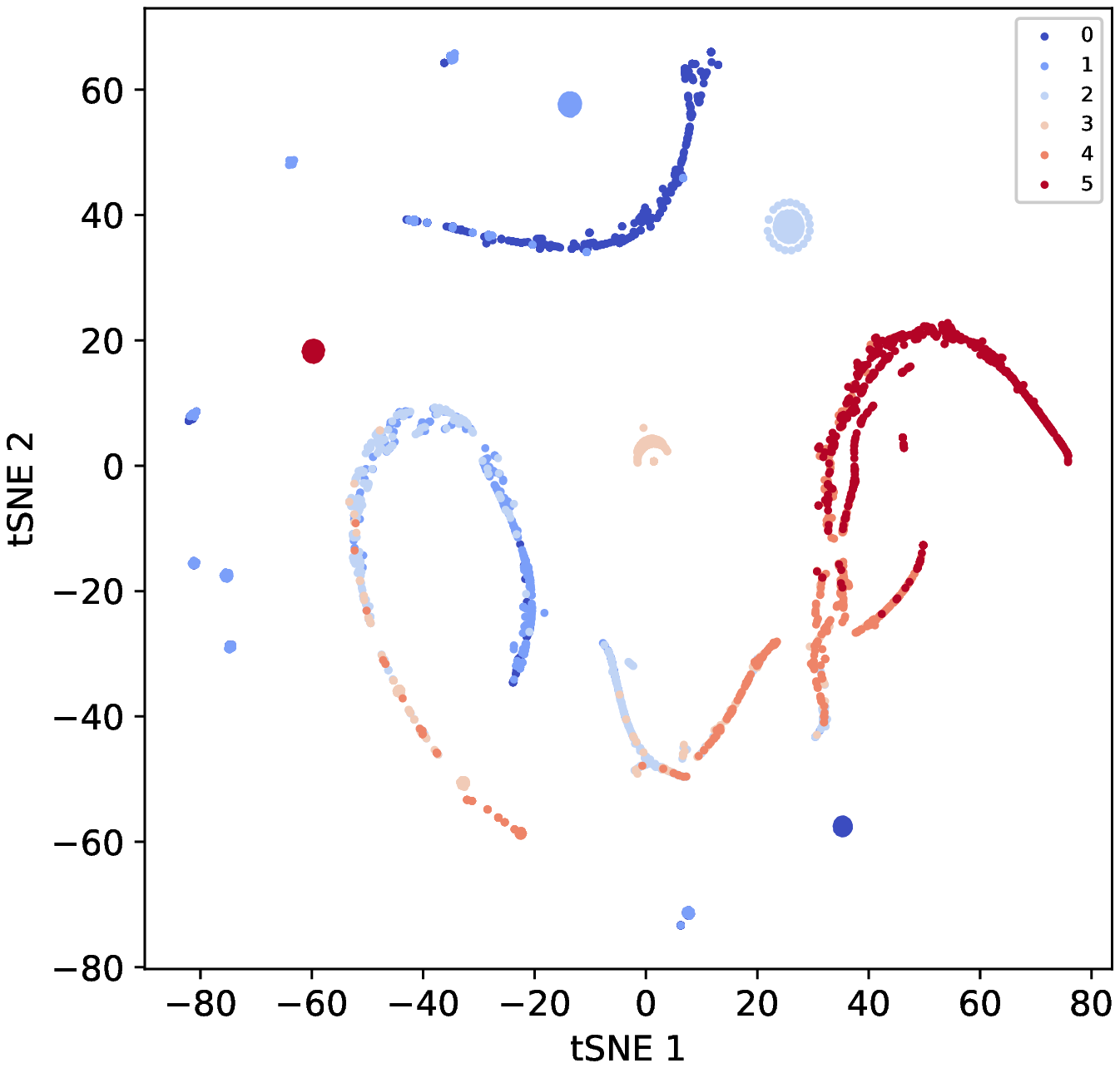}
        \includegraphics[width=0.33\linewidth]{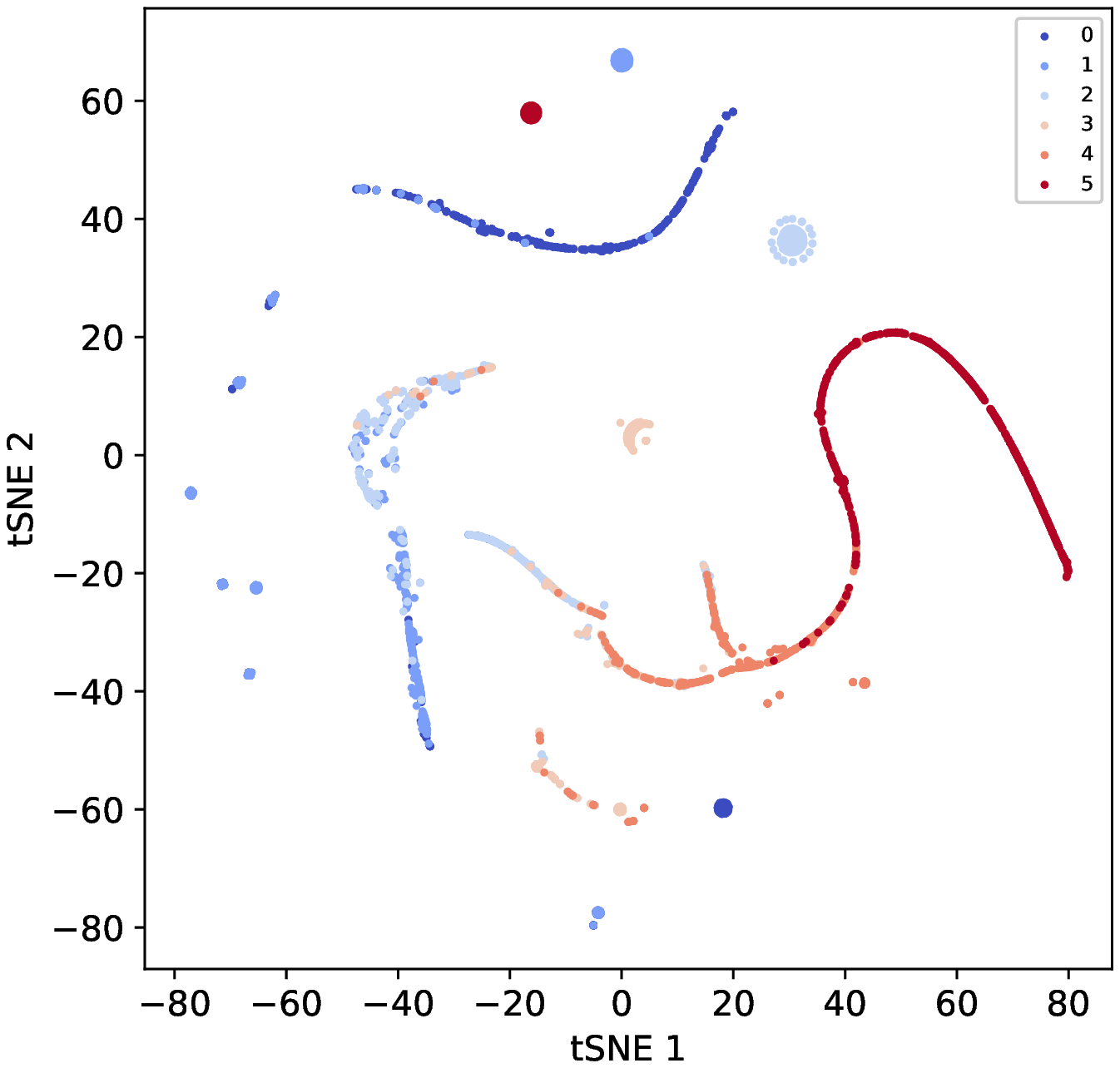}
        \hspace*{-1in}
  
      \hspace*{-1in}
        \includegraphics[width=0.33\linewidth]{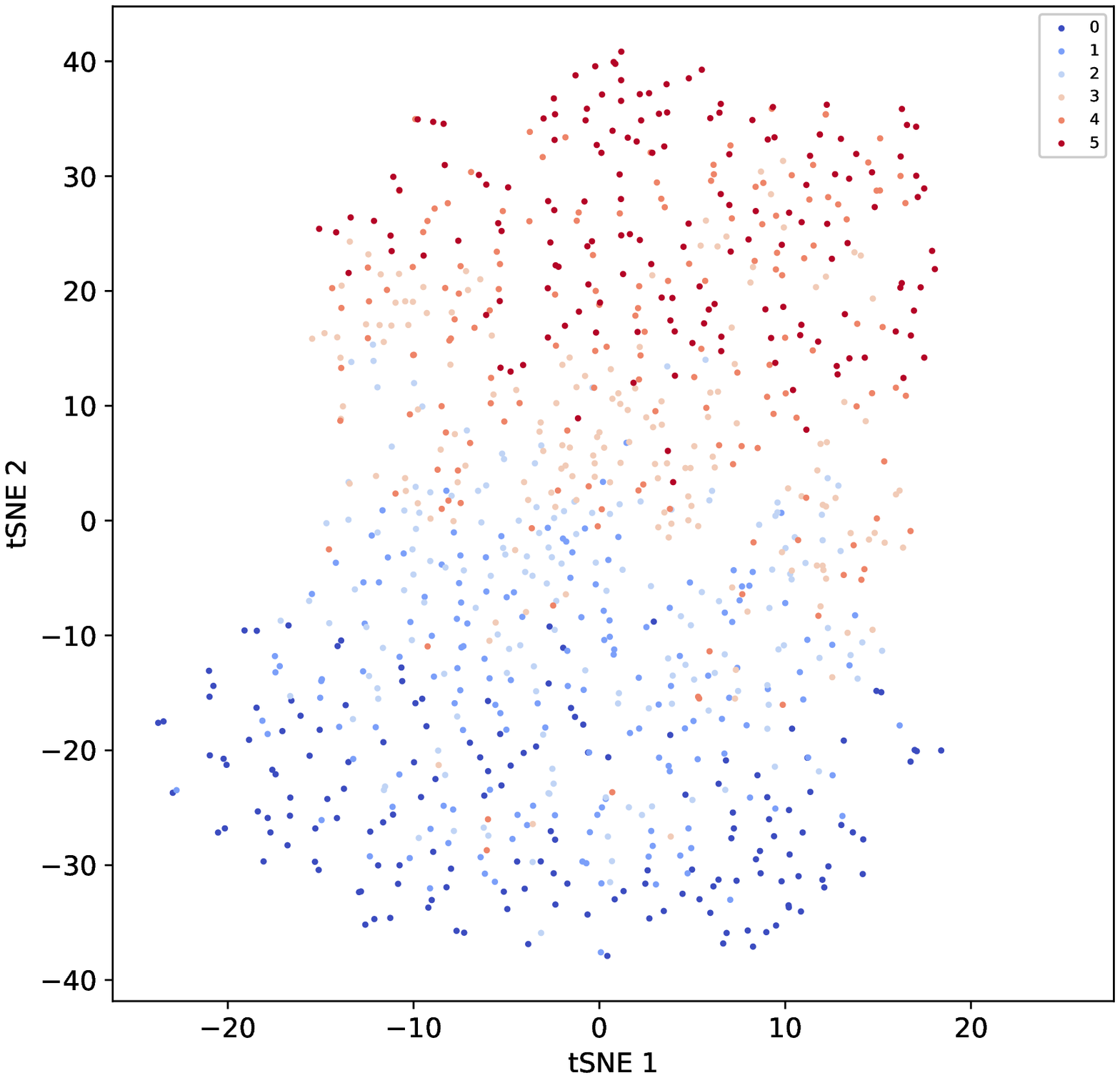}
        \includegraphics[width=0.33\linewidth]{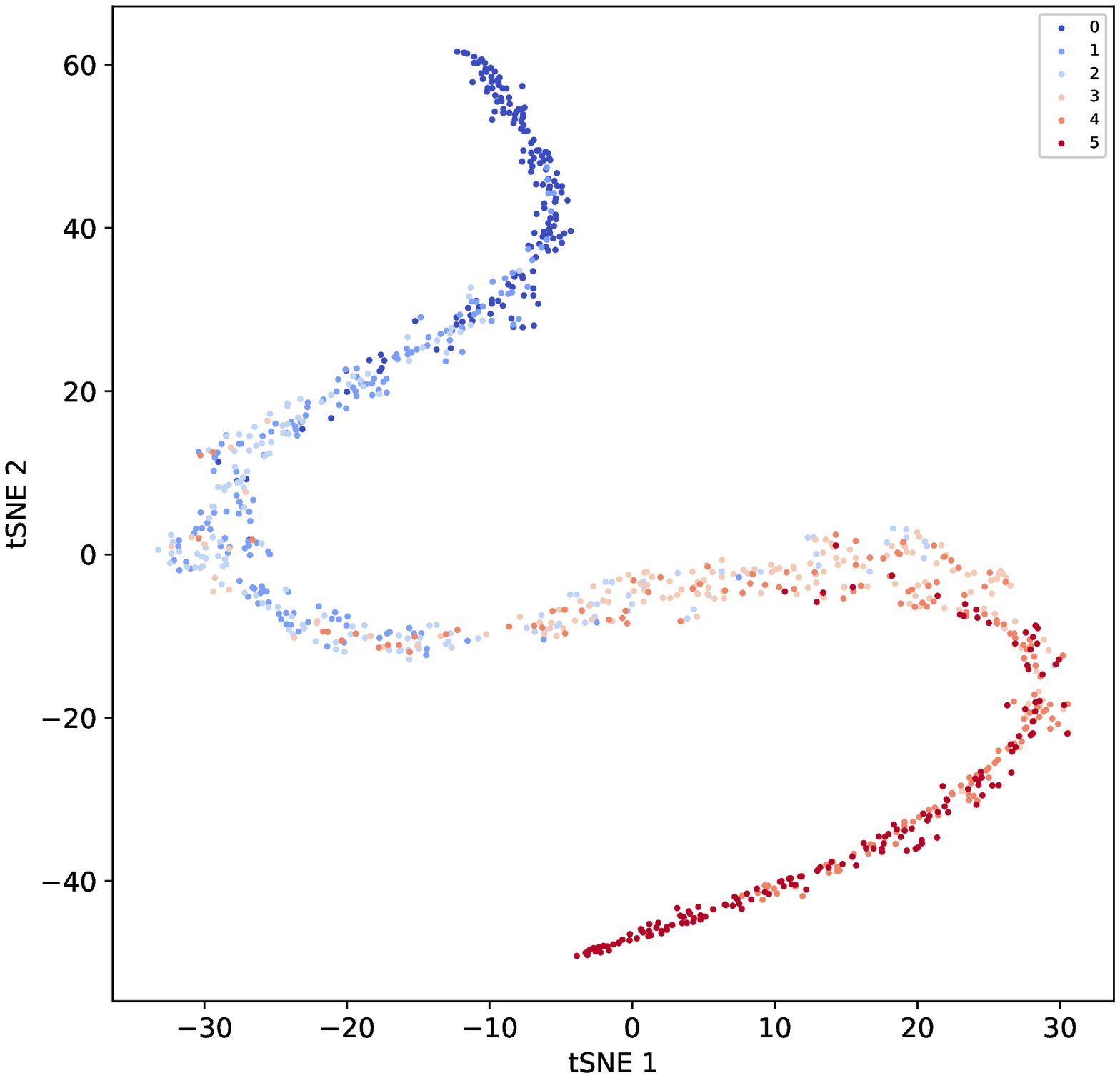}
        \hspace*{-1in}
        \caption{tSNE on graph and MLP embeddings with Degree predicting PageRank, test on
        CiteSeer and Proteins datasets(from left to right: graph embedding, MLP embedding, from top to bottom: Citeseer dataset, Proteins dataset)}
  
        \hspace*{-1in}
        \includegraphics[width=0.33\linewidth]{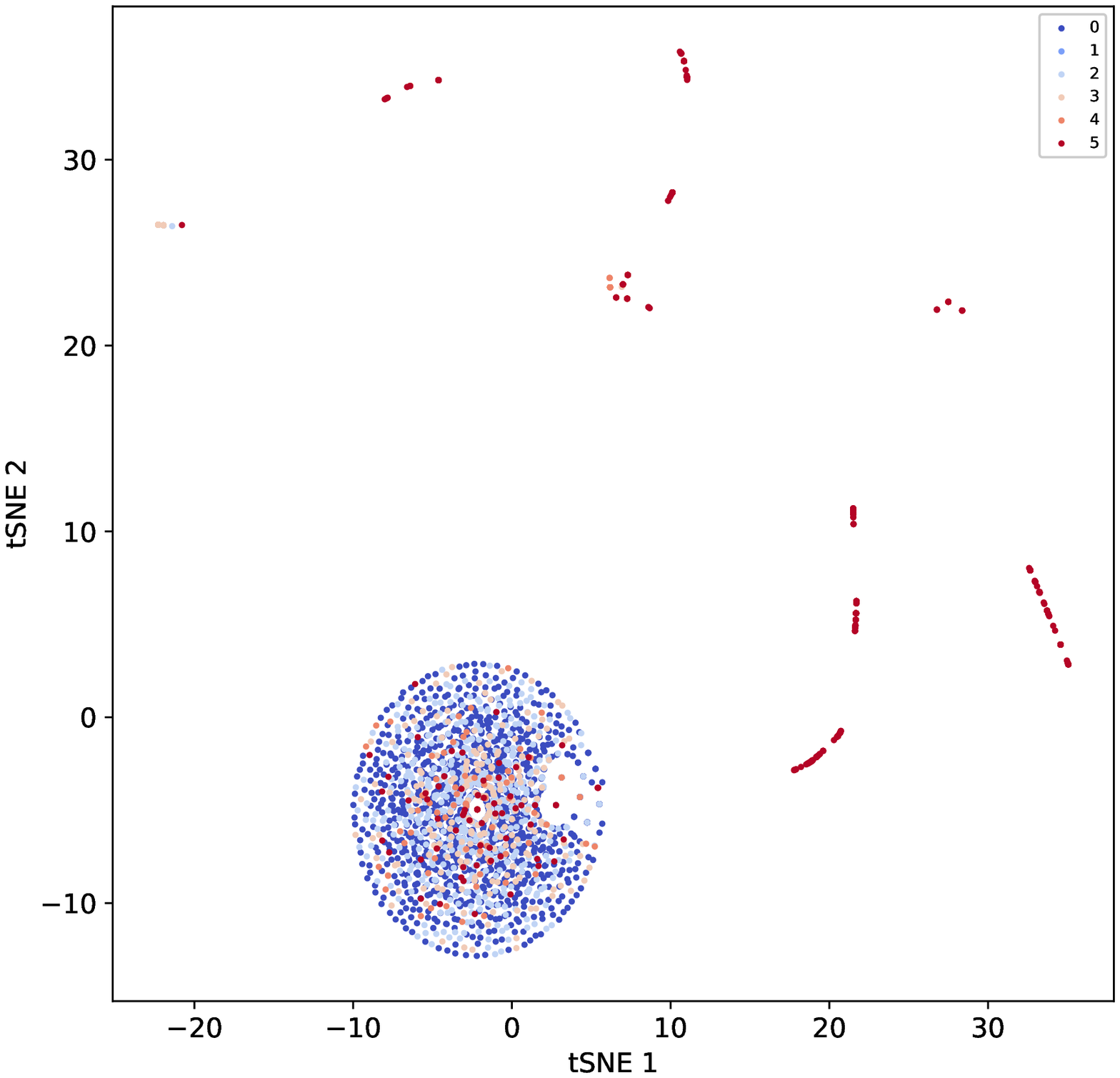}
        \includegraphics[width=0.33\linewidth]{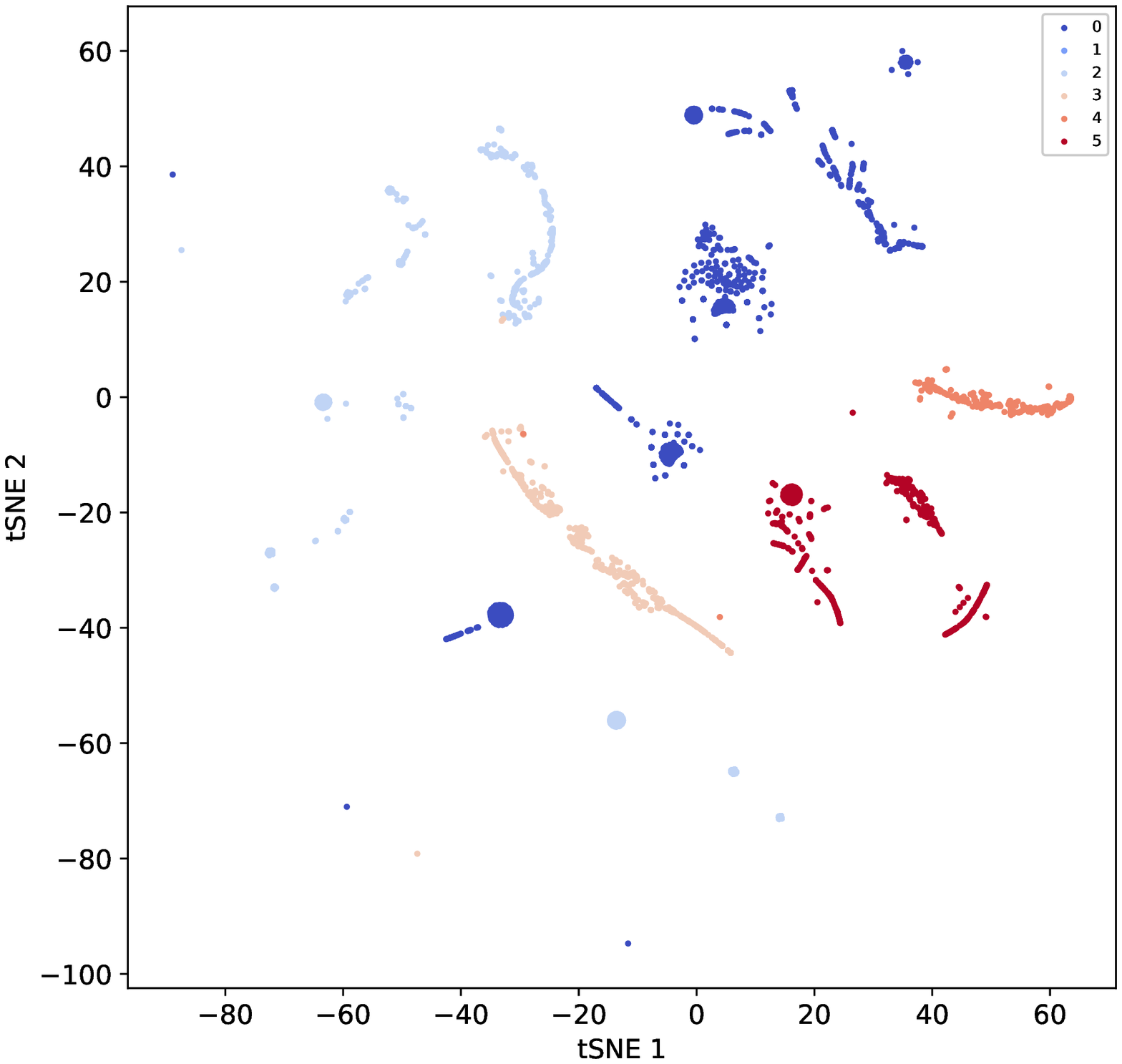}
        \includegraphics[width=0.33\linewidth]{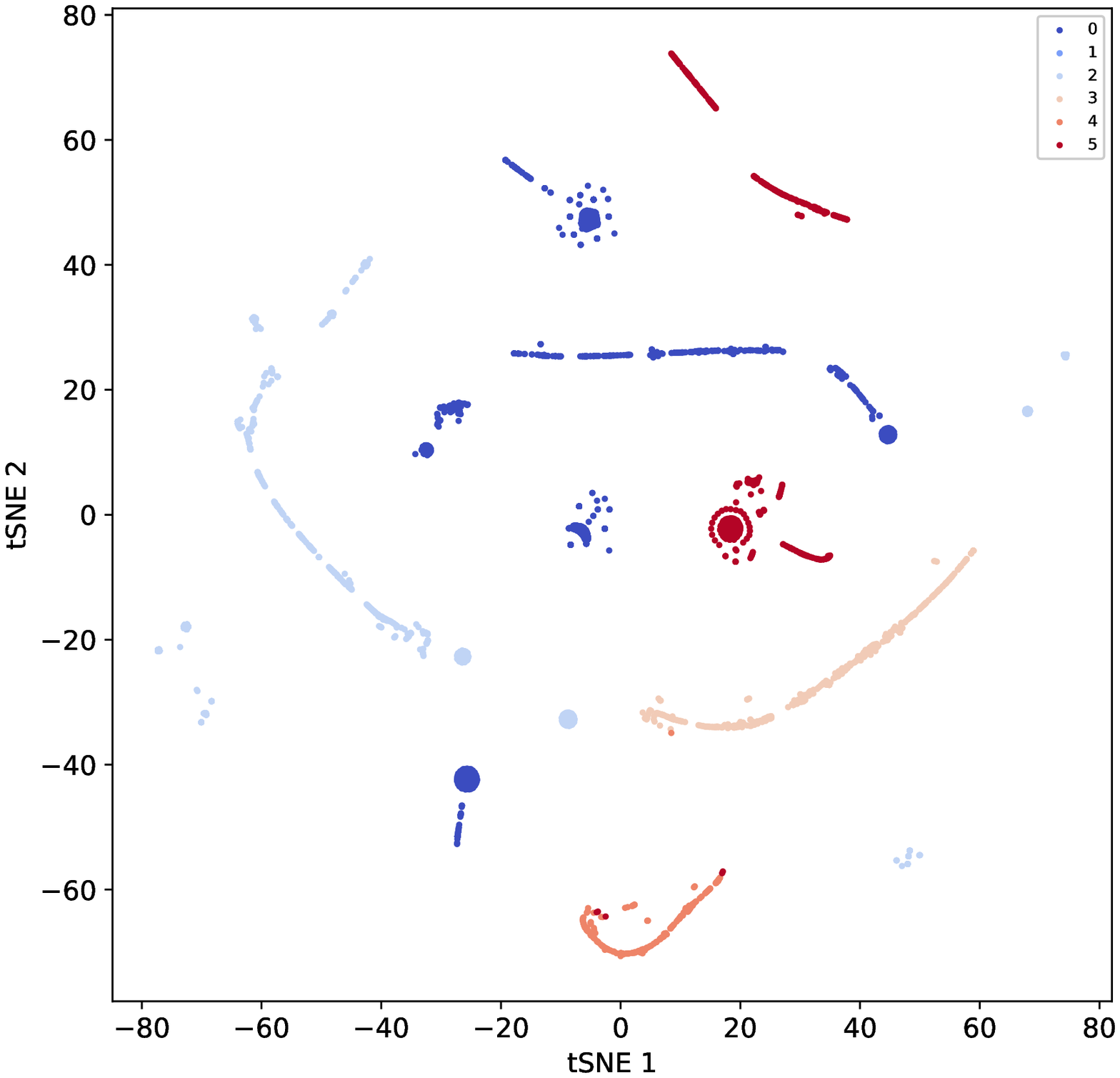}
        \hspace*{-1in}
  
        \hspace*{-1in}
        \includegraphics[width=0.33\linewidth]{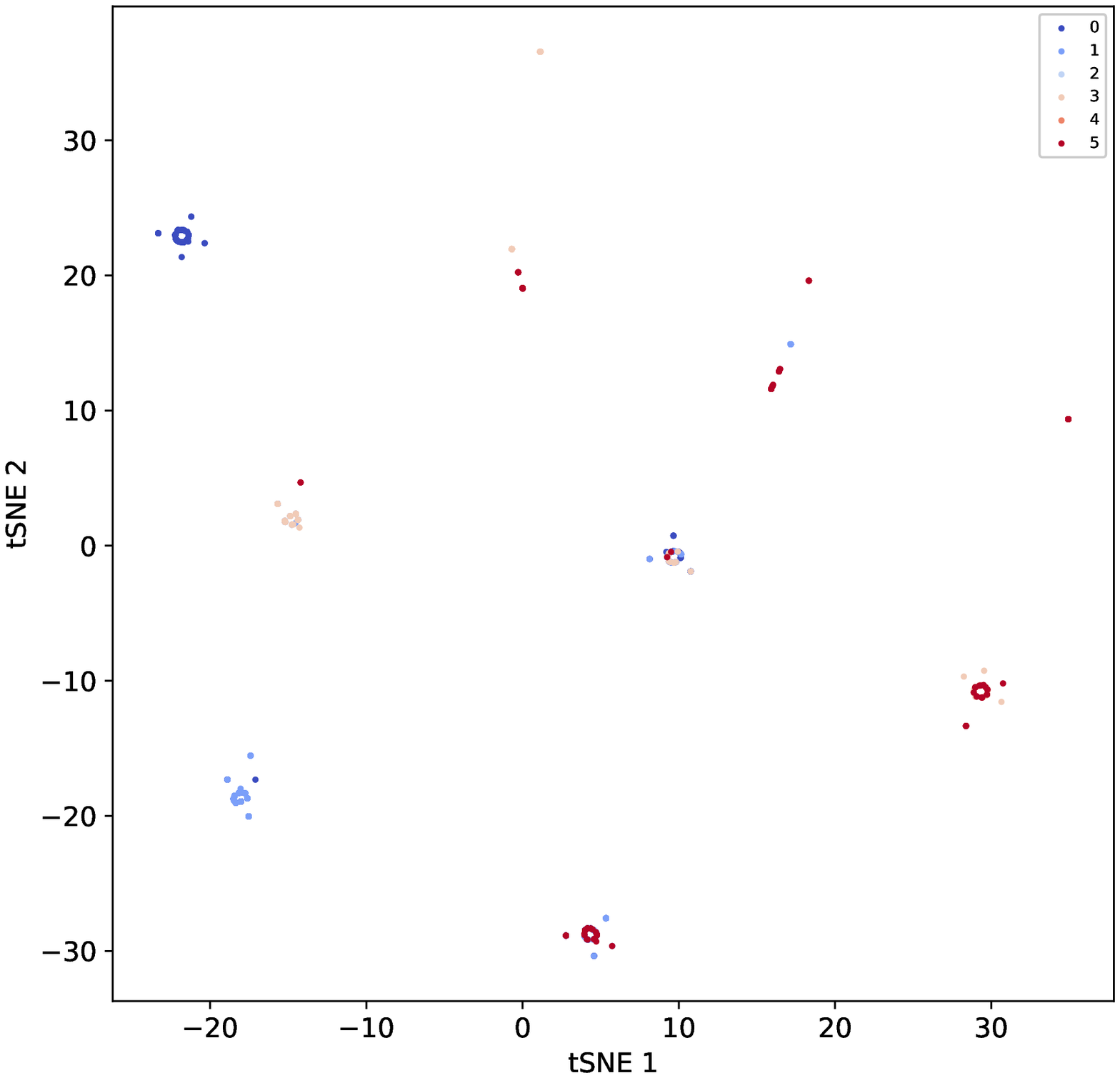}
        \includegraphics[width=0.33\linewidth]{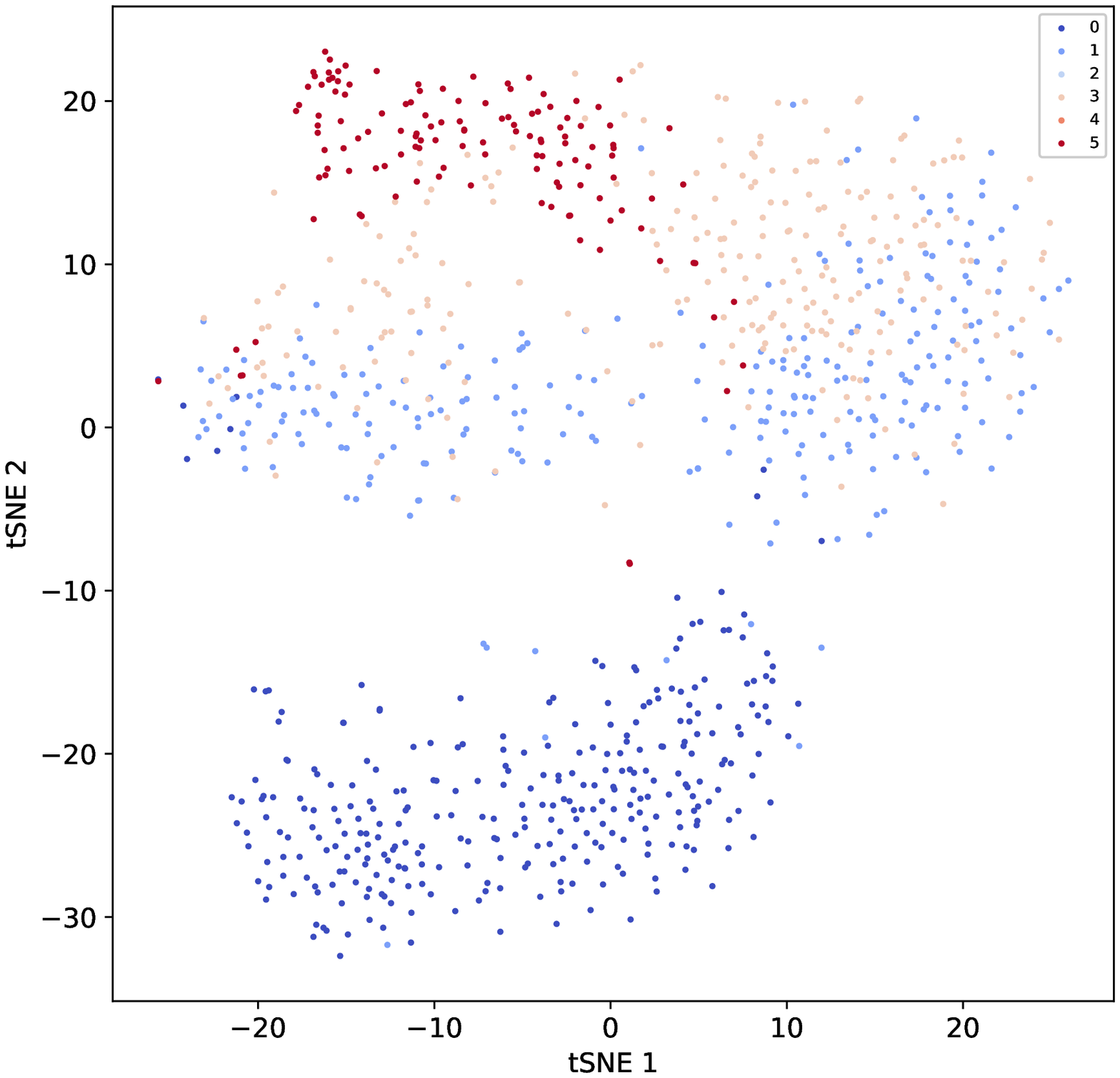}
        \includegraphics[width=0.33\linewidth]{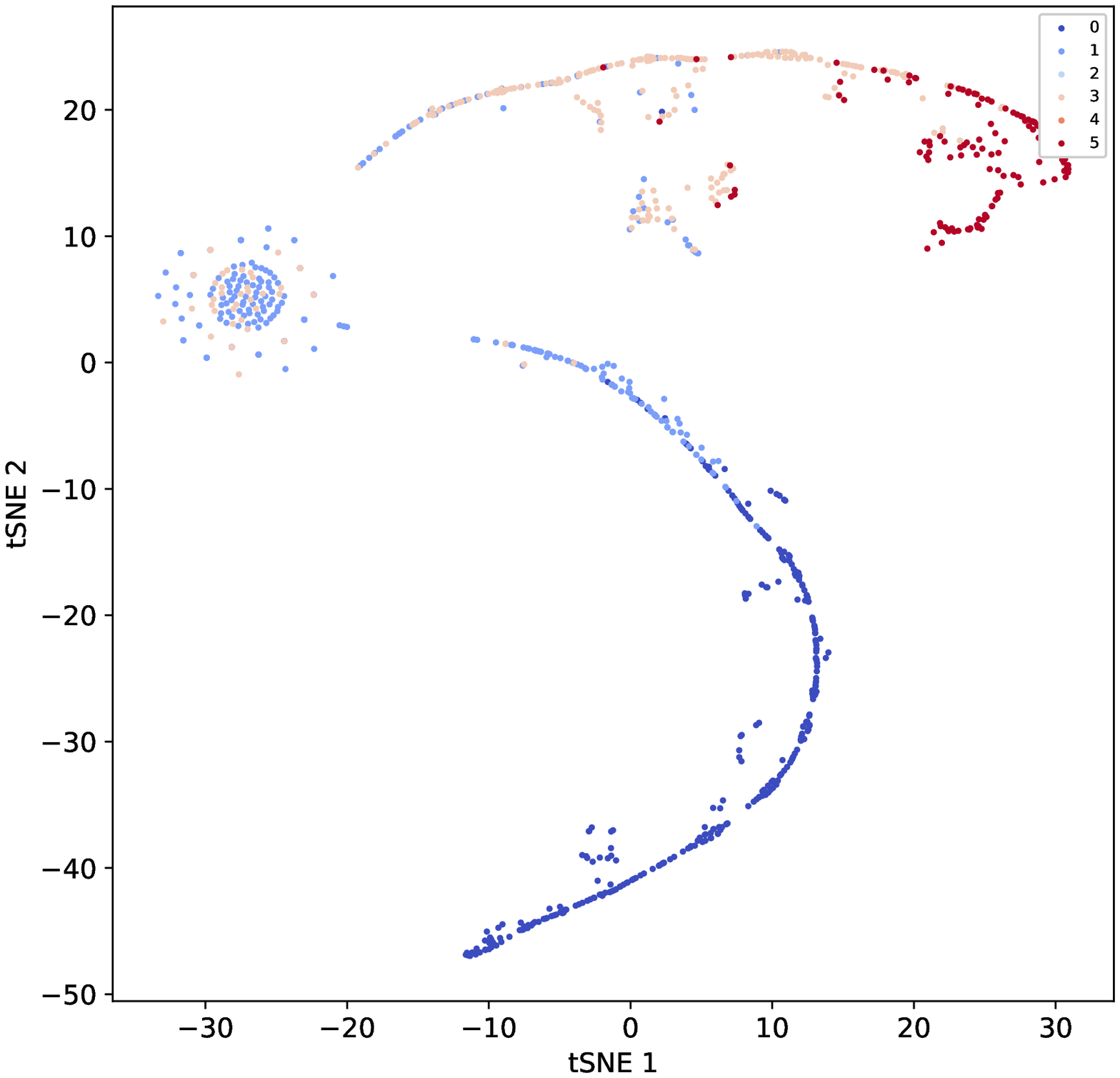}
        \hspace*{-1in}
        \caption{tSNE on initial, graph and MLP embeddings with Constant feature together with Clustering Coefficient predicting Degree, test on
CiteSeer and Proteins datasets(from left to right: initial embedding, graph embedding, MLP embedding, from top to bottom: Citeseer dataset, Proteins dataset)}
        
    \end{center}
    \end{figure}

\section{generalization}

We extend the results that we reach from benchmark datasets on synthetic dataset. We set up five new graph with number of nodes n $\in$ \{50, 200, 400, 800, 1000\}. We use the same model with the same parameter settings in Fea2Fea-single. We found that the feature clustering coefficient is still the most difficult to predict and degree is the easiest to predict, which is the same as what we've previously found in Planetoid and TUdataset. As the number of nodes or features in the network increases, feature mutual prediction becomes more accurate.
  \begin{figure}[!htp]
    \centering
    \begin{center}
    \hspace*{-1in}
        \includegraphics[width=0.33\linewidth]{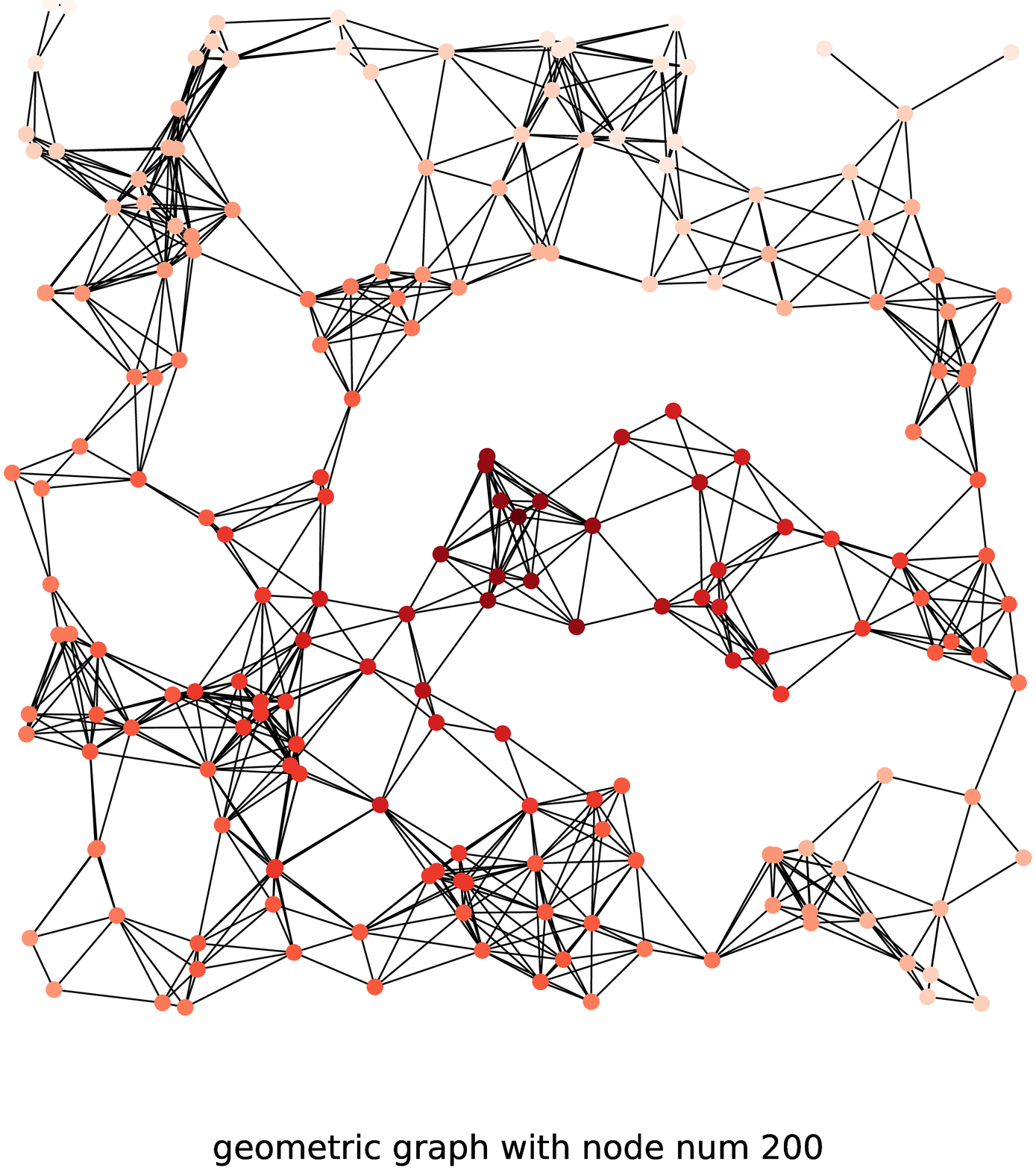}
        \includegraphics[width=0.33\linewidth]{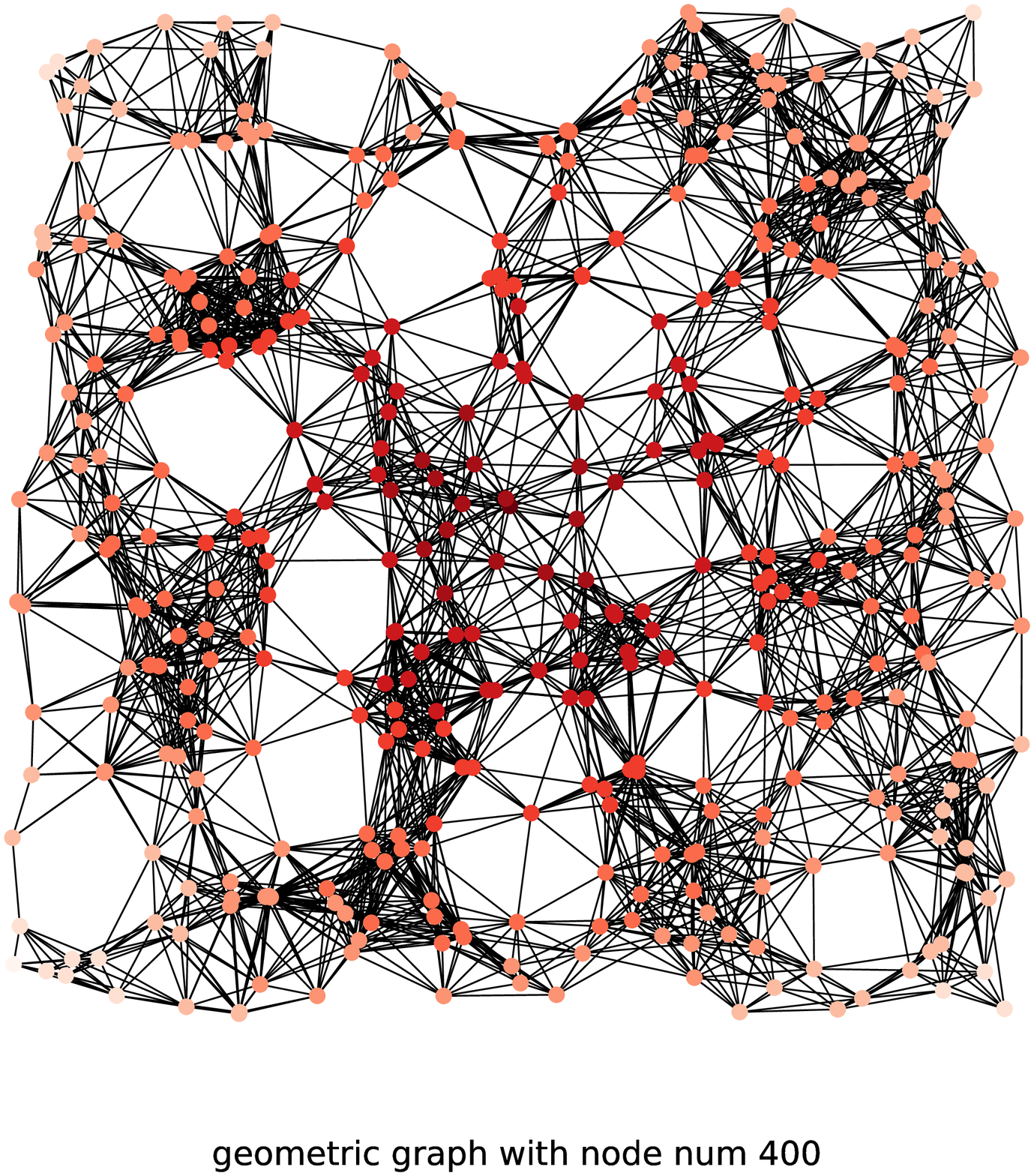}\includegraphics[width=0.33\linewidth]{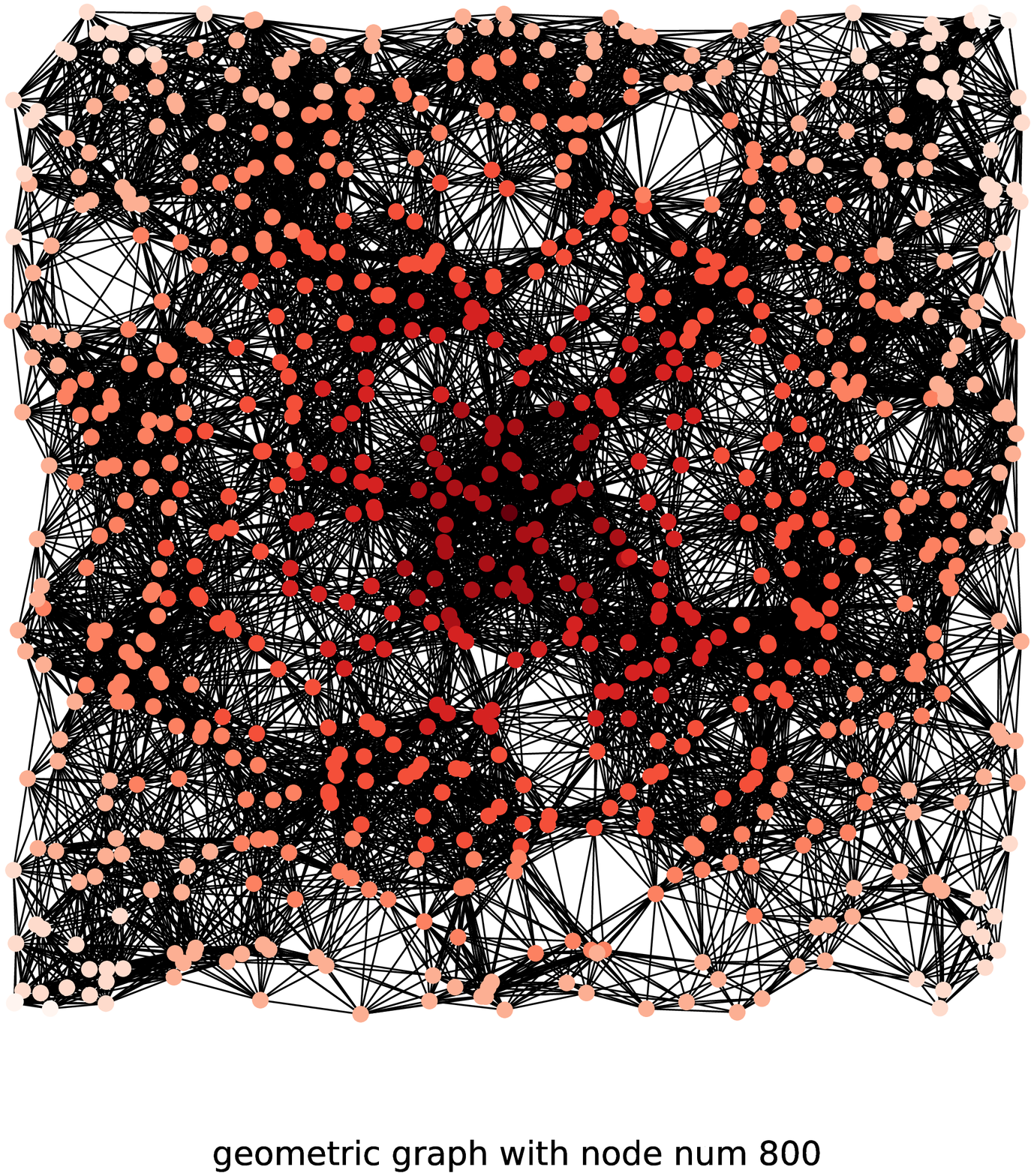}
        \hspace*{-1in}
        \caption{examples of generated geometric dataset from networkx with nodes 200, 400 and 800}
      \end{center}
    \end{figure}
    \begin{table*}[!htb]
      \centering
      \caption{Feature to Feature Prediction on Planetoid Datasets (bins = 6)}
      \begin{tabular}{*{6}{c}} \toprule
    {Aim}  & \multicolumn{5}{c}{ Number of nodes}
    \\
    \cmidrule(lr){2-6}
     & 50 & 200 & 400 & 800 & 1000 \\ \hline
    {\textit{Cons} $\rightarrow$ \textit{Deg}}   & 0.800 & 0.950 & \B1.000 & 0.975 & \B1.000  \\    
    {\textit{Deg} $\rightarrow$ \textit{Deg}}    & 0.800 & 0.950 &\B1.000 & 0.975 &\B 1.000  \\
    {\textit{Clu} $\rightarrow$ \textit{Deg}}    &\B1.000 &\B 1.000 & \B1.000 & 0.975 & \B1.000  \\
    {\textit{PR} $\rightarrow$\textit{Deg}}      & 0.800 & 0.950 & \B1.000 & \B1.000 & 0.990  \\
    {\textit{Avglen} $\rightarrow$ \textit{Deg}} & \B1.000 &\B 1.000 & \B1.000 &\B 1.000 & \B1.000  \\
    {\textit{Cons} $\rightarrow$ \textit{Clu}}   & \B0.800 & 0.750 & 0.550 & 0.650 & \B0.800  \\
    {\textit{Deg} $\rightarrow$ \textit{Clu}}    & 0.800 & \B0.900 & 0.525 & 0.775 & 0.780  \\
    {\textit{Clu}  $\rightarrow$ \textit{Clu}}   & \B1.000 & 0.850 & 0.675 & 0.700 & 0.850  \\   
    {\textit{PR} $\rightarrow$ \textit{Clu}}     & 0.800 & 0.800 & 0.500 & 0.762 & \B0.810 \\      
    {\textit{Avglen} $\rightarrow$ \textit{Clu}} & \B0.800 & 0.700 & 0.625 & 0.688 & 0.720 \\    
    {\textit{Cons} $\rightarrow$ \textit{PR}}    & \B1.000 & 0.850 & 0.850 & 0.912 & 0.910  \\
    {\textit{Deg} $\rightarrow$ \textit{PR}}     & 0.800 & 0.900 & 0.850 & 0.912 & \B0.950  \\
    {\textit{Clu} $\rightarrow$ \textit{PR}}     & 0.600 & 0.850 & 0.825 & 0.825 & \B0.930  \\
    {\textit{PR} $\rightarrow$ \textit{PR}}      & \B1.000 & 0.850 & 0.875 & 0.850 & 0.930  \\
    {\textit{Avglen} $\rightarrow$ \textit{PR}}  & 0.600 & 0.850 & 0.825 & 0.863 & \B0.890  \\\hline
    \bottomrule
      \end{tabular}
    \end{table*}
\end{document}